\documentclass[10pt,journal,cspaper,compsoc]{IEEEtran}
\ifCLASSOPTIONcompsoc
\else
\fi
\ifCLASSINFOpdf
\else
\fi
\usepackage[dvips]{graphicx}
\usepackage{times}
\usepackage{epsfig}
\usepackage{amsmath}
\usepackage{amssymb}
\usepackage{color}
\usepackage{tabularx}
\usepackage{setspace}
\usepackage{subfig}
\usepackage{multirow}
\usepackage{url}
\usepackage{epstopdf}

\newcommand{\calV}{{\cal V}}

\makeatletter
\def\blfootnote{\xdef\@thefnmark{}\@footnotetext}
\makeatother

\begin{document}
%

\title{Lifting  Object Detection Datasets into 3D}

%
%
%
%

\author{Jo{\~a}o~Carreira*, Sara~Vicente*, Lourdes~Agapito and~Jorge~Batista 
\IEEEcompsocitemizethanks{
\IEEEcompsocthanksitem J. Carreira is with the Department of Electrical Engineering and Computer Science, University of California at Berkeley, and the Institute of Systems and Robotics, University of Coimbra. \protect \\
E-mail: carreira@eecs.berkeley.edu
\IEEEcompsocthanksitem S. Vicente is with Anthropics. \protect \\
E-mail: sara@anthropics.co.uk
\IEEEcompsocthanksitem L. Agapito is with University College London. \protect\\
E-mail: l.agapito@cs.ucl.ac.uk
\IEEEcompsocthanksitem J. Batista is with the Institute of Systems and Robotics,
University of Coimbra. \protect\\
E-mail: batista@isr.uc.pt}
\thanks{*First two authors contributed equally.}}

%
%

\markboth{IEEE Transactions on Pattern Analysis and Machine Intelligence (PAMI), Submitted October 2014}{}

%


\IEEEcompsoctitleabstractindextext{%
\begin{abstract}
While data has certainly taken the center stage in computer vision in recent years, it can still be difficult to obtain in certain scenarios. In particular, acquiring ground truth 3D shapes of objects pictured in 2D images remains a challenging feat and this has hampered progress in recognition-based object
reconstruction from a single image. Here we propose to bypass previous solutions such as 3D scanning or manual
design, that scale poorly, and instead populate object category
detection datasets semi-automatically with dense, per-object 3D
reconstructions, bootstrapped from:\emph{(i)} class labels,
\emph{(ii)} ground truth figure-ground segmentations and \emph{(iii)}
a small set of keypoint annotations. Our proposed algorithm first
estimates camera viewpoint using rigid structure-from-motion and then
reconstructs object shapes by optimizing over visual hull proposals
guided by loose within-class shape similarity assumptions. The visual
hull sampling process attempts to intersect an object's projection
cone with the cones of minimal subsets of other similar objects among
those pictured from certain vantage points. We show that our method is
able to produce convincing per-object 3D reconstructions and to
accurately estimate cameras viewpoints on one of the most challenging
existing object-category detection datasets, PASCAL VOC. We hope that
our results will re-stimulate interest on joint object recognition and
3D reconstruction from a single image.
\end{abstract}

}

\maketitle

\IEEEdisplaynotcompsoctitleabstractindextext

%
\IEEEpeerreviewmaketitle

\section{Introduction}

\IEEEPARstart{R}{ecognizing} an object's category and 
the region it occupies in an image, purely from pixel-level information,
is now closer to becoming a reality. Next on the list is to infer the
object's 3D surfaces. The availability of large datasets such as
PASCAL VOC \cite{Everingham10} and Imagenet \cite{imagenet_cvpr09} has
paved the way to important advances in object segmentation and
recognition over the last few years. Progress has been comparatively
slower in the area of object reconstruction from a single image,
hindered by the challenge in acquiring the necessary training data ---
ideally hundreds of thousands of images in uninstrumented settings
aligned with their ground truth 3D shapes. One possible way
forward, as computer graphics evolves, could be to render the data and
learn to reconstruct in an environment that resembles a 3D computer
game setting. Alternatively, depth sensors such as Kinect could be
employed, but these are not yet fully practical in general settings ---
e.g. objects far from the camera, or outdoors. A third option would be
to build large datasets with objects in videos and use
structure-from-motion techniques
\cite{paladini2009factorization,videopopup,katerina14} to recover their shape.

\begin{figure*}
  \centering
  \includegraphics[width=0.80\linewidth]{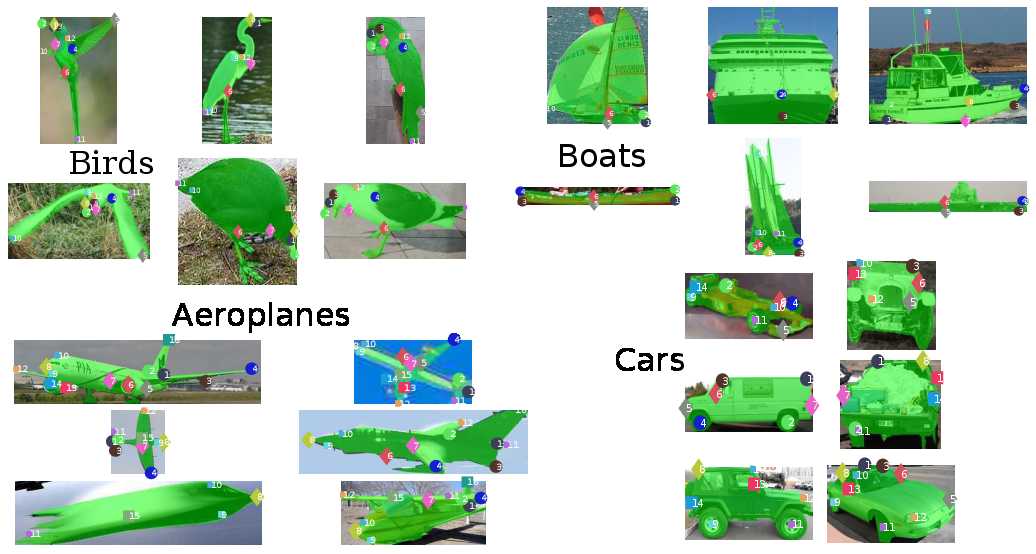} 
  \caption{Example input images to our algorithm for four different classes -- -aeroplanes, birds, boats and cars --- of the PASCAL VOC dataset and their associated figure-ground segmentations and keypoints. Modern detection datasets such as VOC exhibit significant intra-class variability, making it challenging for traditional approaches to class-specific object reconstruction based on linear non-rigid shape models. \label{fig:annotations}}
\end{figure*}

Here we propose instead to build upon existing, and extremely popular, recognition datasets and to directly reconstruct them aided by available annotations. While our experimental focus will be on PASCAL VOC, our proposed techniques are general and could be applied to any other object detection dataset (e.g.~\cite{imagenet_cvpr09}), as long as ground truth class labels, figure-ground segmentations and a small number of per-class keypoints are available, as is the case for VOC \cite{BroxSegmentationCVPR11} and is illustrated in fig.~\ref{fig:annotations}. These types of annotations can nowadays be easily crowdsourced over Mechanical Turk, as they require only a few clicks per image --- e.g. a recently unveiled object detection dataset comes with around 2 million object segmentations \cite{lin2014microsoft}.

Fig.~\ref{fig:annotations} illustrates what may be the major difficulty in our stated intentions: typically there is drastic intra-class shape variation, which makes previous class-specific reconstruction approaches based on linear shape models impractical. Instead we propose a multiview reconstruction strategy. Unlike settings where multiple calibrated images of the same object are available \cite{Hartley2004}, detection datasets are composed of uncalibrated images of different instances of the same class of objects (most often assembled from images available on the web). We bypass the  problem of establishing point correspondences between different objects, which is still unmanageable with current technology, by relying on a very small set of consistent per-class ground truth keypoint matches, from which scaled orthographic camera viewpoints are bootstrapped. We also bypass segmentation, another yet incompletely solved vision problem despite much recent progress \cite{ijcv_12,eccv_12}, and rely on ground truth silhouettes as input to our dense reconstruction engine which is based on a novel visual hull based algorithm.

Visual hull computation has been shown  to be a simple but powerful reconstruction technique when many diverse views of the same object are available. We adapt it to operate on category detection imagery proposing a novel formulation that we denote \textit{imprinted visual hull reconstruction}. The basis of our algorithm is to embed visual hull reconstruction within  a sampling-based approach. We propose a set of candidate reconstructions for each input image by running the visual hull algorithm multiple times using the current image and two additional images sampled from the dataset. We prioritize viewpoints that are known to best expose the 3D shape of most objects. Finally, we select the most consistent reconstruction amongst the proposed candidates by maximizing intra-category similarity.

Our contributions span different areas of computer vision:
\begin{itemize} \itemsep1pt \parskip0pt \topsep0pt
\item \textbf{The recognition problem:} a first attempt to semi-automatically augment object detection datasets, here instantiated on PASCAL VOC, with dense per-object 3D geometry without requiring annotations beyond those readily available online.
\item \textbf{The reconstruction problem:} we propose a new data-driven method for class-based 3D reconstruction that relies only on 2D information, such as figure-ground segmentations and a few keypoint annotations.
\end{itemize}

This paper extends the original conference work \cite{vicente14} with additional visualizations and many new experiments: we present a direct evaluation of viewpoint estimation, a new analysis of the reconstructed shapes on PASCAL VOC and studies on the influence of the main parameters of our algorithm on the results. The full source code of our algorithm, which we call \textit{carvi} as in "carving", and our synthetic dataset are freely available online\footnote{\url{http://www.isr.uc.pt/~joaoluis/carvi}}.

\section{Related Work}

Formerly a dominant paradigm, model-based recognition, which reasoned jointly about object identity and 3D geometry \cite{grimson1990object,huttenlocher1987object,lowe1987three}, was permanently upstaged in the 1990's by a flurry of view-based approaches \cite{turk1991eigenfaces,murase1995visual}. The main appeal of view-based approaches was their flexibility: collecting a few example images of the target objects and annotating their bounding boxes or 2D keypoint locations became all the manual labor required to build a recognition system, averting the need for cumbersome manual 3D design or special instrumentation (3D scanners). This 2D data-driven approach made it possible to attack harder problems such as category-level object recognition. Model-based recognition held important advantages, nevertheless \cite{Mundy}: modeling the 3D geometry of an object enabled arbitrary viewpoints and occlusion patterns to be rendered and recognized, and it also facilitated higher-level reasoning about interactions between objects and a scene. 

While the popularity of model-based recognition was falling, interest in multiview 3D reconstruction was rising, powered by breakthroughs in affine structure-from-motion \cite{ullman1979interpretation,tomasi1992shape,koenderink1991affine} and the adoption of projective geometry \cite{faugeras1993three}. Multiview reconstruction has since been largely solved in the rigid case \cite{Hartley2004}, as calibration parameters and correspondences can be reliably estimated and the problem reduces to a well-understood geometric optimization problem. In this paper we are interested in the harder problem of class-based reconstruction, where the goal is to reconstruct different objects from the same category, each pictured in a single image. This problem will be the focus of the rest of our literature  review in this section.


\subsection{Class-based reconstruction with 3D data}
Most class-based reconstruction methods make use of 3D data in their pipeline which provides prior 3D information about the shape of the objects in the class.

\subsubsection{Reconstruction as prototype selection and alignment}
In certain cases a precise 3D model of the target object is known \cite{roberts1963machine,lowe87} and the problem can be seen as a special case of class-based reconstruction that focuses on reconstructing a specific instance of the class. The goal reduces then to locate and estimate the viewpoint of the object instance in the image. This problem has traditionally attracted much attention in computer vision and is currently going through a revival due to the increased availability of accurate 3D models for many objects and better feature extraction technology \cite{lim2013parsing}.

Other methods rely on a dataset of 3D shapes and automatically choose and align the one that best fits the object in the image \cite{Aubry14}.  To account for differences between the 3D exemplars and the object depicted, Su and Guibas \cite{su2014estimating} choose a few exemplars that appear to be similar to the depicted shape and combine them to produce a single depth map.

\subsubsection{Reconstruction as patch classification}

Hassner and Basri \cite{hassner2006example} performed class-based single-view reconstruction using also 3D training data but without building a parametric model for the class. Instead, their model searches the training set for patches similar to those in the test image, then transfers the associated depth information. Related methods, but employing parametric classifiers, have also been successfully used for scene reconstruction from a single image \cite{saxena2008_depth,Hoiem_2005,ladicky2014pulling}.

\subsubsection{Reconstruction using morphable models}
When multiple 3D shapes corresponding to different instances of the same class are available, they can be used to build a morphable model for the 3D class, that can then generalize and be fit to unseen instances of the class. Morphable models are low-dimensional parametric models that have been used to represent the shape of many object classes. They can be built from 3D scans of different instances of the class, e.g. the face model in \cite{blanz1999morphable} and the human body model in \cite{Anguelov:SCAPE2005}, or using 3D meshes obtained from shape repositories, such as Google Sketchup as in \cite{zia2013detailed}. 

The trained morphable model can then be used in a variety of tasks: (1) to reconstruct from a single image, usually with some user interaction to initialize the viewpoint \cite{blanz1999morphable}, (2) as a prior for reconstruction from multiple images \cite{bao2013dense} or from a depth map \cite{dame2013dense}, or (3) for performing object detection and pose estimation \cite{zia2013detailed} in a single image. One factor that limits the applicability of these models is the need for 3D training data. In order to partially overcome this issue, \cite{Cashman:dolphins:2013} proposed a hybrid method that uses a single 3D shape together with 2D information in order to build a morphable model. The system was demonstrated on classes with limited intra-class shape variability such as dolphins or pigeons.

\subsection{Data-driven class-based reconstruction}
In this paper we focus on a data-driven method for class-based reconstruction that operates directly on an unordered dataset of images and some associated 2D annotations, without using any 3D data. To the best of our knowledge, there have only been two previous attempts \cite{Zhu_ModelEvolution:2010,prasad2010finding} at tackling the problem in a purely data-driven fashion. These two approaches build upon traditional non-rigid structure-from-motion methods \cite{Bregler2000NRSFM}, originally developed for reconstruction from video, and either produce sparse reconstructions \cite{Zhu_ModelEvolution:2010} or have only been demonstrated on simple classes such as flower petals and clown-fish, while requiring complex manual annotations \cite{prasad2010finding}. 

Our method differs from the above in two important aspects: (1) we require only a small set of keypoint correspondences across images and these are not the only points we reconstruct; instead we reconstruct dense 3D models of the objects, and (2) we do not build a morphable model for the class. Instead, our aim is to reconstruct every object instance, using ``borrowed'' shape information from a small number of similar instances seen from different viewpoints. This makes our method applicable to classes with large intra-class variation as those in the VOC dataset.

\subsection{Bottom-up reconstruction}
More general object reconstruction approaches have been devised, that do not require any class information, tracing back to classic works by Binford, Marr and others \cite{agin1976computer,marr1978representation,mohan1989using}. Methods such as shape from shading (SfS) \cite{HORNThesis1970} hold great promise but have so far been applied only in very restricted settings as they make strong assumptions about global illumination conditions and the reflective properties of the object. Recently SIRFS \cite{barron2012shape} went one step beyond traditional SfS and aimed to recover not only the shape and shading of an object but also reflectance and incident illumination, all from a single image.

Another family of approaches attempts to compute shape from a single silhouette \cite{Prasad06,twarog2012playing,toppe2013relative,vicente2013balloon}. For example \cite{Prasad06} employed the representation of geometric images to successfully reconstruct simple shapes symmetric with respect to the image plane, but required large amounts of user interaction for more complex objects. A similar principle of symmetry with respect to the image plane is the basis of \cite{toppe2013relative}, that focused on including shading information and improving the user experience. The same symmetry principle was also used, albeit more lightly, in \cite{vicente2013balloon} where the focus was on coping with deformations.

\subsection{Dataset augmentation into 3D}
The goal of populating detection datasets with 3D annotations has been previously considered for the class \emph{person} \cite{bourdev2009poselets}, using an interactive method to reconstruct a set of body joints. In contrast, we obtain full dense reconstructions for a variety of classes. In a related approach, \cite{russell2009building} targeted the problem of automatically bootstrapping 3D scene geometry from 2D annotations on the LabelMe dataset --- instead, we focus on objects. 

Recently and perhaps closest to our approach, Karsch \emph{et al.} \cite{Karsch2013} experimented with reconstructing VOC objects, using manual curvature annotations on boundaries but computed 2.5D reconstructions while we focus on the full 3D problem. Even more recently --- and concurrently with our original paper --- new 3D annotations were added to 12 rigid classes of the PASCAL dataset \cite{xiang2014beyond} in a largely manual effort. All instances of these 12 classes were manually associated with one out of a small set of 3D CAD models posed in the correct viewpoint. The goal of the dataset is to provide ground truth viewpoint information, not shape, and the CAD models provide only a very coarse approximation to the rich set of shapes in PASCAL (e.g. about 50\% overlap, or roughly just as much as top automatic semantic segmentation systems \cite{eccv_12}).

\section{Problem formulation}

We assume we are given a set of images depicting different instances of the same object class, which may be very diverse in terms of object scale, location, pose, shape and articulation. We make the small simplification in this paper of not addressing the problem of reconstructing occluded objects, that are marked as such in PASCAL. 
Each object instance $n$ has a corresponding binary mask $B_n$ --- a figure-ground segmentation locating the object boundaries in the image --- and $K_i$ specific keypoints for each class $i$, which are on easily identifiable parts of the object, such as \emph{``left mirror''} for cars or \emph{``nose tip''} for aeroplanes. Each object instance $n$ is annotated with its visible keypoints, i.e.~the set $(x^n_{k},y^n_{k})$ of 2D image coordinates\footnote{These annotations are publicly available for all the 20 classes in the VOC dataset \cite{BroxSegmentationCVPR11}.}.

Our goal in this paper is to 
obtain a dense 3D reconstruction of each of the object instances. It is easy to see that this is a severely underconstrained problem since each image corresponds to a different object instance. Without additional prior knowledge, and if each instance is to be reconstructed independently, an infinite number of reconstructions would be available that could exactly generate the silhouette $B_n$. 



\subsection{Our data-driven approach}

Instead of relying on bottom-up reconstruction methods and performing reconstruction completely independently for each instance, we leverage the information contained in images of other objects from the same category, by building upon the assumption that at least some instances of the same class will have a similar 3D shape. We propose a feedforward strategy with two phases: first, camera viewpoints  are estimated for all objects using both keypoint and silhouette information; secondly, a sampling-based approach that employs a novel variant of \emph{visual hull reconstruction} is used to produce dense per-object 3D reconstructions. The details of these two steps will be further explained in the following two sections. 

\section{Camera viewpoint estimation and refinement\label{sec:SFM}}

\begin{figure}[t]
  \centering
  \textbf{Estimated azimuth for cars.}\\
  \vspace{10pt}
  \includegraphics[width=0.48\textwidth]{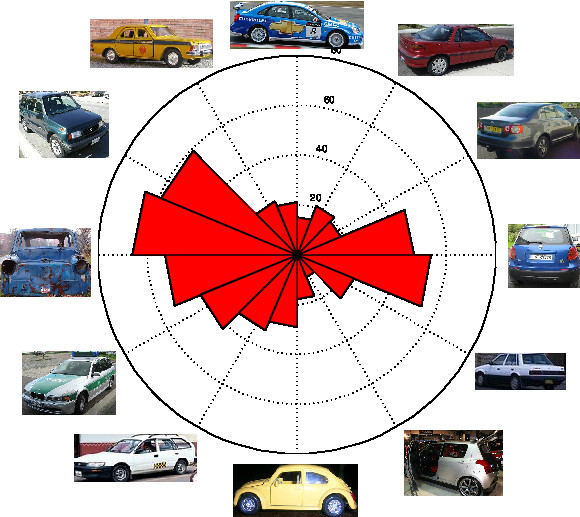}\\
\vspace{20pt}  
  \textbf{Estimated elevation for aeroplanes.} 
  \vspace{5pt}	
  \includegraphics[width=0.48\textwidth]{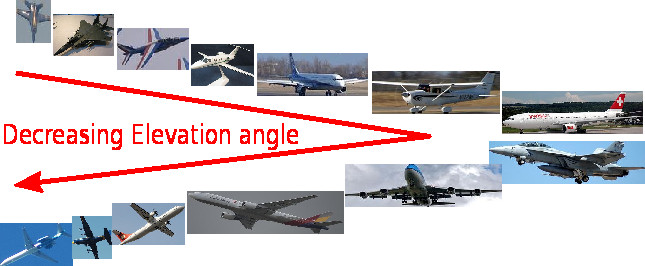}
  \caption{Results of the camera viewpoint estimation. Our method provides useful insight about viewpoint distribution for the different classes in VOC. Here, we show the histogram of different azimuths for ``car'' and a few samples of estimated elevation angle for ``aeroplane''. Note the significant intra-class variation.
   \label{fig:cameras}}
\end{figure}

The first step of our algorithm is to estimate the camera viewpoint for each of the instances using the factorization based rigid structure-from-motion algorithm of Marques and Costeira \cite{Marques:Costeira:2008}. Although rigid modeling may appear to be a suboptimal choice at first sight, several non-rigid structure-from-motion algorithms make use of a similar strategy in viewpoint estimation due to the lack of robustness to noise of specialized non-rigid SfM viewpoint estimates. Simply put, the hope is that the -- admittedly flawed -- assumption of rigidity acts as a regularizer. The algorithm we adopted models projection using scaled orthographic cameras and requires global point correspondences across the different object instances. In comparison with full perspective cameras, scaled orthographic cameras are considerably easier to model, do not require calibration parameters and are a reasonable approximation for the problem considered.

Using the annotated keypoints we form an observation matrix for each instance:
\begin{equation}
W_n = 
\begin{bmatrix}
x^1_n& ... & x^K_n\\
y^1_n& ... & y^K_n\\
\end{bmatrix}
\end{equation}
where $n$ is the object instance and $K$ is the number of annotated keypoints.
Some of the entries in this matrix may be unknown if the keypoint is not visible for this instance.  

The SFM algorithm finds the 3D shape $S$, a $3\times K$ matrix that can be seen as a rough ``mean shape'' for the $N$ object instances in the class, the motion matrices $M_n$ and the translation vectors $T_n$, by minimizing the image reprojection error:
\begin{equation}
\sum_{n = 1}^{N}{\left\lVert W_n - \begin{bmatrix}M_n&T_n\end{bmatrix} 
\begin{bmatrix}S \\ \mathbf{1}_{1\times K} \end{bmatrix}
\right\rVert^2_{F}}
\label{eq:reprojection_error}
\end{equation}
under the constraint that  $M_n M_n^{T} = (\alpha_n)^2 I_{2\times 2} \ \forall n$. This constraint guarantees that matrices $M_n$ correspond to the first two rows of a scaled rotation matrix which can be easily converted into a full rotation matrix $R_n$ and scale parameter $\alpha_n$. The SfM algorithm used does not require that all keypoints are visible in all the instances, i.e.~it can deal with missing data. We follow \cite{Marques:Costeira:2008} and use an iterative method with power factorization to minimize the reprojection error.

For classes with large intra-class variation or articulation, we manually select a subset of the keypoints to perform rigid SfM. There are two types of classes that follow this behavior:  the class \emph{boat} and animal classes \footnote{Note however that most object classes are non-rigid in practice, for example cars can have their doors open, wheels rotate, etc.}. The class \emph{boat} includes both sailing boats and motor boats and since the sails are not present in the motor boats, we estimate the camera by only considering the keypoints on the hull. Excluding the keypoints corresponding to the sails is crucial for the refinement step detailed in the next section.
For animals,  which undergo articulation, different instances may have very different poses. For these classes, we assume that the camera viewpoint is defined with respect to the head and torso and exclude the keypoints corresponding to the limbs or wings when performing rigid SFM. For all classes, for robustness, we double the number of instances by adding left-right flipped versions of each image.

\subsection{Silhouette-based camera refinement\label{sec:SFM_refinement}} 
\begin{figure*}[t]
  \centering
  \begin{tabular}{@{}c @{} c@{}}
  \includegraphics[width=0.5\textwidth]{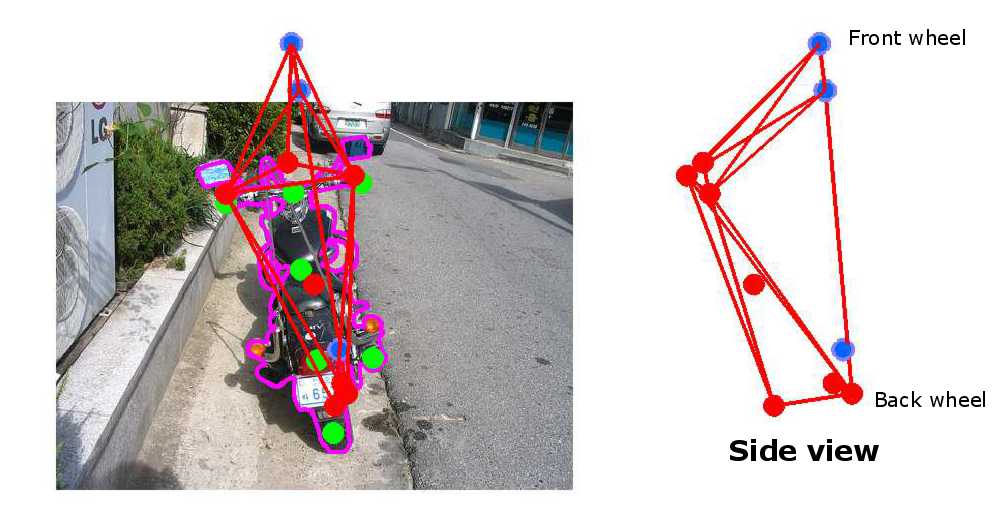}&
  \includegraphics[width=0.5\textwidth]{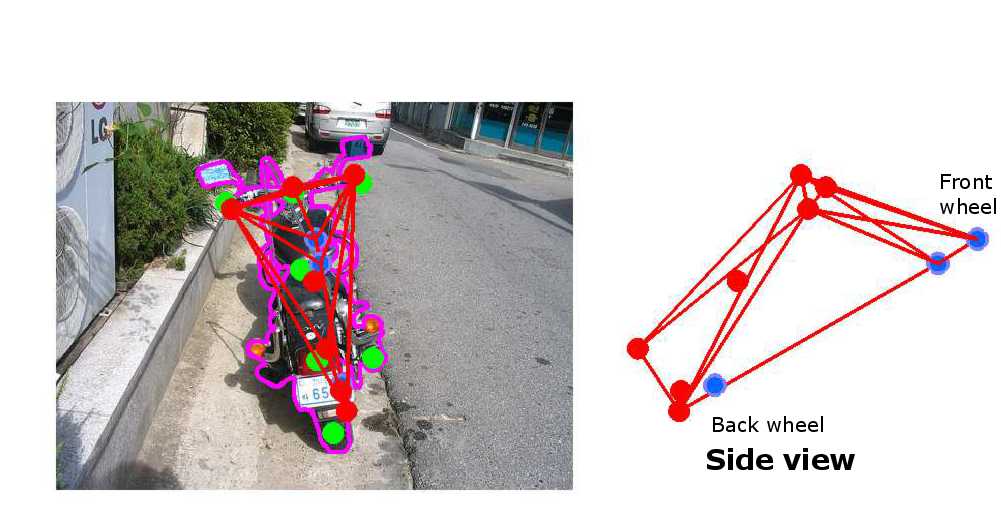}\\
  (a) &(b)
  \end{tabular}
  \caption{\label{fig:refinement} Viewpoint estimate before (a) and after (b) silhouette-based refinement. 
  For both (a) and (b) we show the SfM rigid shape $S$ for the motorbike class from the front and side views. The visible ground truth keypoints are shown in green, the corresponding SfM points are shown in red and the SfM points corresponding to occluded ground truth keypoints are shown in blue. The silhouette is shown in pink. Our viewpoint refinement optimizes for viewpoint estimates where all shape points project inside the object silhouette.}
\end{figure*}

To obtain the camera pose estimate for a particular instance, the SFM algorithm only uses the keypoints visible in that instance. If some keypoints are self-occluded, and since the shape $S$ is an average shape of all the objects in the class, this may lead to an inaccurate estimate of the camera viewpoint (see fig.~\ref{fig:refinement} (a)). However, the silhouette provides extra constraints that can be used to refine this initial estimate of the camera viewpoint. In particular, if the estimated shape was the correct one, all the keypoints, even the ones which are not visible, should reproject inside the silhouette. This constraint is not satisfied by the initial result of fig.~\ref{fig:refinement} (a), but by including a soft-constraint that encourages all points to reproject inside the silhouette we obtain a better viewpoint estimate as can be seen in fig.~\ref{fig:refinement} (b). 
We include this constraint as a soft-constraint to account for imprecisions in the shape estimation and keypoint and silhouette annotations.

More formally, we refine the camera estimate $M_n$ and $T_n$ by fixing the shape $S$ and minimizing an energy function of the form:
\begin{multline}
E(M_n,T_n)  = \\ \left\lVert W_n - \left[M_n\  T_n\right]
\begin{bmatrix}S \\ \mathbf{1}_{1\times K} \end{bmatrix}
\right\rVert^2_{F} + \text{D}\left(\left[M_n\  T_n\right] 
\begin{bmatrix}S \\ \mathbf{1}_{1\times K} \end{bmatrix}\right)
\label{eq:refinement}
\end{multline}
under the constraint $M_n M_n^{T} = (\alpha_n)^2 I_{2\times 2}$. The first term of this energy is the reprojection error as in \eqref{eq:reprojection_error}  and the second term is defined as:
\begin{align}
\text{D}\left(\left[M_n\  T_n\right] 
\begin{bmatrix}S \\ \mathbf{1}_{1\times K} \end{bmatrix}\right) =
\text{D}\left(\begin{bmatrix}u^1\ ...\ u^K\\ v^1\ ...\ v^K\end{bmatrix}\right) = \sum_{k = 1}^{K}{\text{C}\left(u^k,v^k\right)}
\end{align}
where $\text{C}(.,.)$ is the distance transform map from the figure-ground segmentation $B_n$. A point on the mean shape $S$ incurs a penalty if its reprojection, given by $(u^k,v^k)$, is outside the silhouette and this penalty is proportional to the distance to the silhouette. To minimize this function, we use gradient descent with a projection step into the space of scaled rotation matrices. A similar projection step is used in \cite{Marques:Costeira:2008}. Qualitative results of our camera viewpoint estimation can be seen in fig.~\ref{fig:cameras}.

This camera refinement step can also be used to estimate the camera viewpoint of a new test image, by initializing $M_n$ to the identity matrix and $T_n$ to the center of the mask. This allows our method to reconstruct a previously unseen image, the only requirement being that the keypoints are marked or have been detected and that the object is segmented.

\section{Object reconstruction}

After jointly estimating the camera viewpoints for all the instances in each class, we reconstruct the 3D shape of all objects using shape information borrowed from other exemplars in the same class \footnote{An idea similar in spirit was proposed for segmentation \cite{kim2012shape}}.

\subsection{Sampling shape surrogates}

\begin{figure}
  \centering
  \includegraphics[scale=0.2]{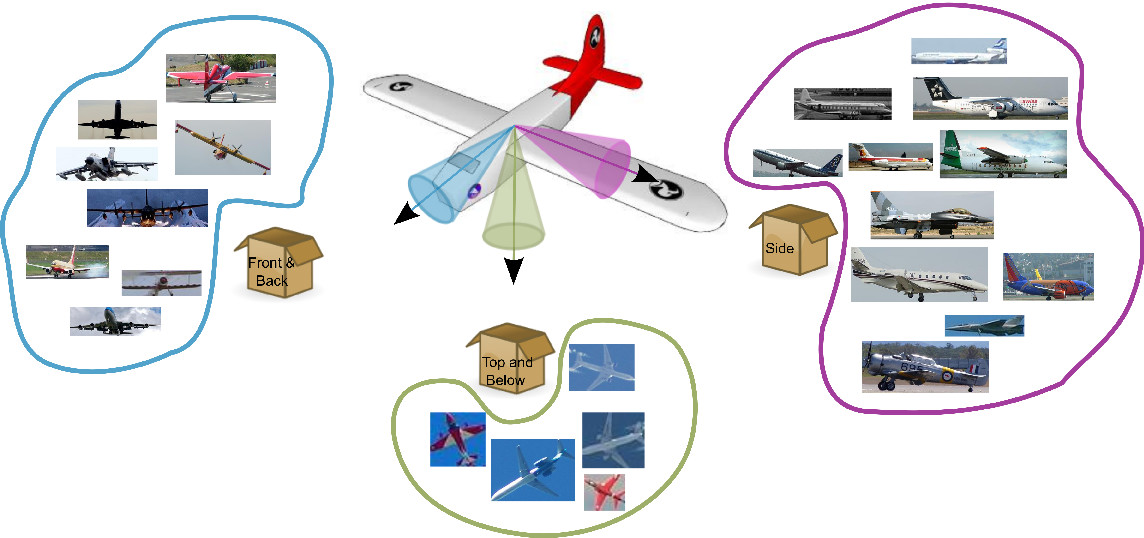}
  \caption{\label{fig:sampling}Illustration of our clustering step.
    Instances which have a viewpoint similar to the principal
    directions are clustered together.  We sample from these clusters
    to generate informative triplets of exemplars for visual hull
    computation. The process is repeated multiple times for each
    target object.}
\end{figure}

In datasets as diverse as VOC, it is reasonable to assume that for every instance $n$ there are at least a few shape surrogates, i.e.~other instances of the same class that, despite not corresponding to the same physical object, have a similar 3D shape. 
Finding shape surrogates is not straightforward, however. When the surrogates have very different viewpoint it is difficult to establish that their 3D shape is similar to the shape of the reference object (e.g. that they are true surrogates) because their appearance changes vastly. In visual hull approaches, such as the one we propose, a tension also exists between reconstructing from fewer silhouettes, which may result in a solution with many uncarved voxels, or from a large number of silhouettes which may instead lead to an over-carved or even empty solution, because  calibration is not exact and ``surrogateness'' is only approximate. Here we strike a compromise: we sample groups of three views, where the two surrogates of the reference instance $n$ are selected among those pictured from far apart viewpoints, so as to maximize the number of background voxels carved away (see fig. ~\ref{fig:sampling}). 

Furthermore, when selecting  \emph{far apart} viewpoints we took inspiration from technical illustration practices, where the goal is to communicate 3D shape as concisely as possible, and it is common to represent the shape by drawing 3D orthographic projections on three orthogonal planes.  
In a similar vein, we restrict surrogate sampling to be over objects pictured from three orthogonal viewpoints, which we will call principal directions.

Our sampling process has three steps:

\vspace{1pt}
\noindent\textbf{(1) Principal direction identification} We found empirically that a good set of principal directions can be obtained by computing the three PCA components of the set of 3D coordinate vectors of the mean shape $S$ (estimated in the rigid SfM step). The results typically correspond to the top/bottom, left/right and front/back directions.

\vspace{1pt}
\noindent\textbf{(2) Clustering instances around the principal
  directions} Instances where the viewpoint difference with respect to
a principal direction is smaller than some threshold ($15^{\circ}$ in
our implementation) are clustered together \footnote{The amount of camera roll is typically low in detection datasets and we did not compensate for it but it may be a good idea in future work.}. All other instances are never chosen as surrogate views.  An illustration of this clustering
step for the ``aeroplanes'' class is shown in fig.~\ref{fig:sampling}.

\vspace{1pt}
\noindent\textbf{(3) Sampling} We start by selecting two of the three
principal directions, with a probability proportional to the number of
associated instances.  Then, from each of the selected
principal directions, we sample one surrogate instance, which together
with the reference instance forms a triplet of views.

Three of the classes in the VOC dataset (\emph{bottle}, \emph{dining 
table} and \emph{potted plant}) have view-dependent keypoints since
it is difficult to define a reference frame for the object \cite{BroxSegmentationCVPR11}.
 This makes 3D registration ambiguous for all the instances of the class.
 Instead of
sampling surrogate instances, we observed that some of the instances of these classes are approximately rotational symmetric
and synthesize the surrogates from the reference instance by rotating
it around the axis of symmetry, every 45 degrees. 
This is obviously a rough approximation for instances that considerably depart from rotational symmetry.

\begin{figure}[t]
  \centering
  \includegraphics[width=0.35\textwidth]{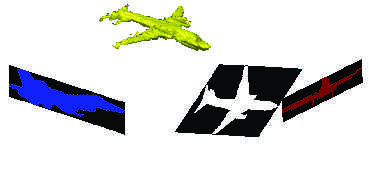}
  \includegraphics[width=0.35\textwidth]{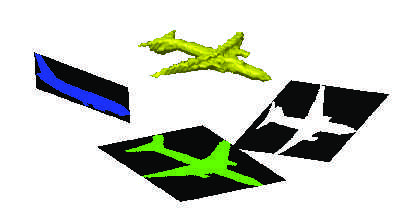}
  \caption{Illustration of the imprinted visual hull reconstruction method, when sampling two different triplets corresponding to the same reference instance (in white). The reconstructions are obtained by intersecting the three instances shown and their left-right flipped versions.\label{fig:visual_hull}  
}
\end{figure}

\subsection{Imprinted visual hull reconstruction} 

\begin{figure*}[!t]
  \centering
  \includegraphics[width=0.27\textwidth]{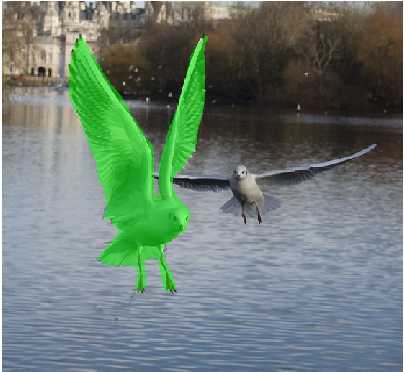}
  \includegraphics[width=0.23\textwidth]{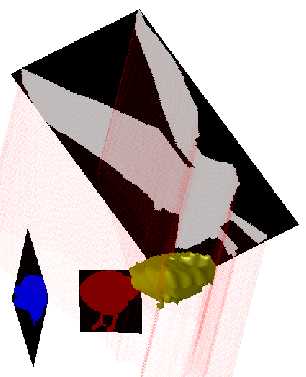}
  \includegraphics[width=0.27\textwidth]{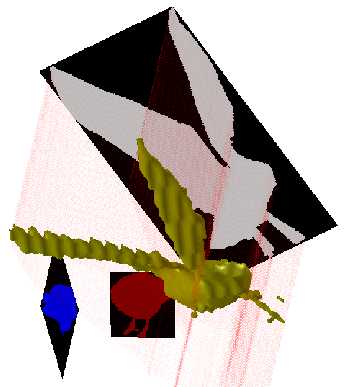}
  \caption{An example image where imprinting improves reconstruction significantly. In the middle, without imprinting, the wings of the reference bird are not reconstructed because it is paired with surrogate shapes having their wings closed. In the reconstruction shown on the right, imprinting fills in wings so as to satisfy silhouette consistency with the reference bird - shown overlaid in green on the left.\label{fig:bird_imprinting}}
\end{figure*}

Recovering the approximate shape of an object from silhouettes seen from different camera viewpoints can be done by finding the visual hull of the shape \cite{laurentini1994visual}, the reconstruction with maximum volume among all of those that reproject inside all the different silhouettes. Visual hull reconstruction is a frequent first step in multi-view stereo \cite{seitz2006comparison}, providing an initial shape that is then refined using photo-consistency. Existing visual hull methods assume that the different silhouettes project from the same physical 3D object \cite{grauman2003inferring}. This is in contrast with our scenario where images of different objects are considered. Visual hull reconstruction is known to be sensitive to errors in the segmentation and in the viewpoint estimate and it is clear that such sources of noise are very present in our framework, and can lead to overcarving if handled naively. 

A clear inefficiency of using the standard visual hull algorithm in our setting is that there is no guarantee that the visual hull is silhouette-consistent with the reference instance $n$, i.e.~that for all the foreground pixels in the mask $B_n$ there will be an active voxel reprojecting on them. This happens because the algorithm trusts equally all silhouettes. Here we propose a variation of the original formulation that does not have this problem, which we denote \textit{imprinted visual hull reconstruction}.
We will use a volumetric representation of shape and formulate imprinted visual hull reconstruction as a binary labelling problem. Let $T$ be the set of instances corresponding to a sampled triplet
and $\calV$ be a set of voxels. The goal is to find a binary labelling $L = \left\{l_v: v \in \calV, l_v \in \{0,1\} \right\}$ such that $l_v = 1$ if voxel $v$ is inside the shape, and $l_v = 0$ otherwise.
Let $C_m(.)$ be a signed distance function such that $C_m(v)<0$ if voxel $v$ is inside the camera cone of instance $m$, and let $\bar{C}(v) = \max_{m \in T} C_m(v)$ be the largest signed distance value over all the cameras, for each voxel $v$.
Visual hull reconstruction can be formulated as the minimization of the energy:
\begin{equation}
E(L)  = \sum_{v \in \calV}{l_v \bar{C}(v)}\label{eq:vh}
\end{equation}

To enforce silhouette consistency with the reference mask $B_n$ (imprinting), we need to guarantee that all the rays cast from the foreground pixels of $B_n$ intersect with an interior voxel. 
Let $R_p$ be the set of voxels that intersect with the ray corresponding to pixel $p$. Imprinting is then enforced by minimizing energy \eqref{eq:vh} under the following constraints:
\begin{equation}
\sum_{v \in R_p}{l_v} \geq 1 \quad \forall \ p \in \text{Foreground}(B_n).\label{eq:silh_consistency}
\end{equation}
Similar constraints have been previously used for multi-view stereo \cite{Cremers-Kolev-pami11}, where they were enforced equally for all the images.
Energy \eqref{eq:vh} can be minimized exactly under constraint \eqref{eq:silh_consistency}, by simply setting
$l^{*}_v = 1$ if and only if $\bar{C}(v)<0$ or if $\exists p, v = \arg \min_{u\in R_p}\bar{C}(u)$. Basically, this energy has a prior for thin structures, in depth. It promotes the construction of a thin layer in depth that fills in the reference mask and is positioned so as to minimize the distance to all considered masks. This is a sensible prior in many cases, because thin surfaces are likely to be carved away using visual hulls, for example sails of boats, bird wings, chair legs, etc. In a few cases, however, it is not ideal, most notoriously when masks are mismatched due to perspective effects in buses and trains (e.g. see fig. ~\ref{fig:reconstructions}).

We chose to formulate our reconstruction algorithm as a labelling problem, to motivate future extensions such as adding pairwise constraints between voxels or connectivity priors \cite{Vicente_Kolmogorov_Rother_2008}. An example case where imprinting is particularly useful is the bird in fig.~\ref{fig:bird_imprinting}.

\subsection{Reconstruction selection}

\begin{figure}[t]
\centering
  \includegraphics[trim=0cm 1cm 0cm 0cm,clip=true,scale = 0.14]{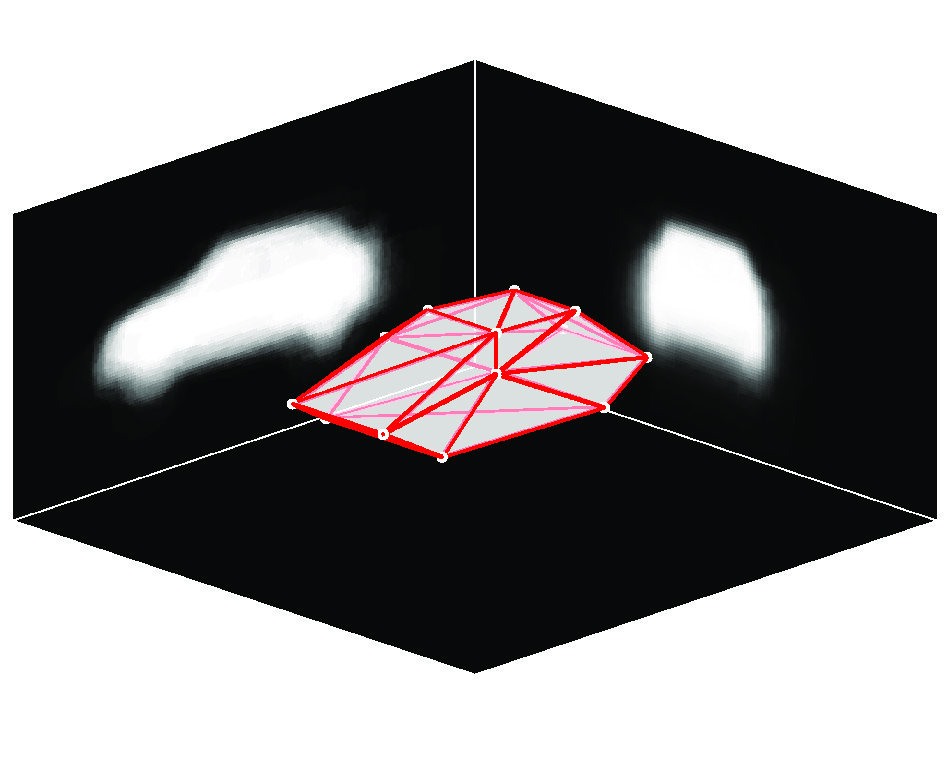}
  \includegraphics[trim=0cm 1cm 0cm 0cm,clip=true,scale = 0.15]{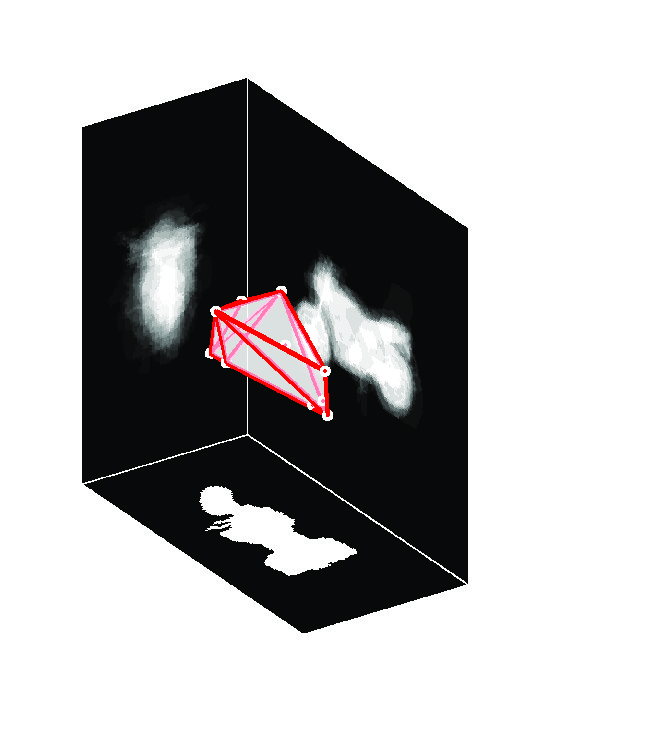}
  \vskip -0.09in
  \caption{Average mask for each of the principal directions for the car and motorbike classes, as well as the convex hull of the 3D keypoints obtained with SFM. These average masks are used when ranking the reconstructions for a single instance. Note that for the class car, there is no instance associated with the top-bottom axis and for motorbike there is only one instance. 
 \label{fig:reconstruction_selection}}
\end{figure}

Once all reconstruction proposals have been computed based on different sampled triplets, the final step is to choose the best reconstruction for the reference instance. Here we propose a selection criterion that follows a simple observation: reconstructions should be similar to the average shape of their object class. Our selection procedure first computes an average mask for each of the principal directions. This is done by aligning the masks of all the instances in each principal direction cluster and averaging them. Afterwards, each reconstruction proposal is projected onto a plane perpendicular to each principal direction and the difference between this projection and the average mask associated with that direction is measured. The final score is the sum of the three differences, one for each direction. The average masks for each principal direction for two classes are shown in fig.~\ref{fig:reconstruction_selection}.




\section{Experiments}

Our main goals in terms of experiments were to evaluate the accuracy of 1) viewpoint estimation and 2) shape reconstruction. We focused on the PASCAL VOC dataset because it is still the most popular object detection dataset and the annotations that our algorithm uses as input are already publicly available. In work published concurrently with our original publication \cite{vicente14}, human provided viewpoint annotations have been gathered for PASCAL VOC \cite{xiang2014beyond}. This allows us to evaluate the camera viewpoint estimation. Regarding the shape reconstruction, in the absence of accurate ground truth data and considering the simplicity of our inputs (keypoints and figure-ground segmentation) \footnote{Synthetic datasets were also successfully used in Kinect \cite{shotton2013efficient}, where the inputs can also be rendered realistically (e.g. depth maps).} we  rely on a synthetic dataset, which we hold as sufficiently realistic.

\subsection{Reconstructing PASCAL VOC}

We consider the subset of 9,087 fully visible objects in 5,363 images from the 20,775 objects and 10,803 images available in the PASCAL VOC 2012 training data and use the publicly available keypoints and figure-ground segmentations \cite{BharathICCV2011}. VOC has $20$ classes, including highly articulated ones (dogs, cats, people), vehicles (cars, trains, bicycles) and indoor objects (dining tables, potted plants) in realistic images drawn from FLICKR. Amongst these, fewer than 1\% have focal lengths in their EXIF metadata, which we ignored. 

\begin{figure*}
\centering
\renewcommand{\arraystretch}{1}
\begin{tabular}{@{}c@{} c@{} c@{} c@{} c@{} c@{} c@{} c@{} c@{} c@{}}
\includegraphics[height = 0.041\textheight, width = 0.125\textwidth, keepaspectratio = true]{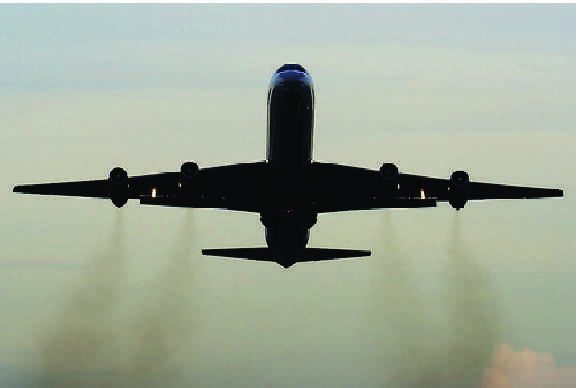} &
\includegraphics[height = 0.041\textheight, width = 0.125\textwidth, keepaspectratio = true]{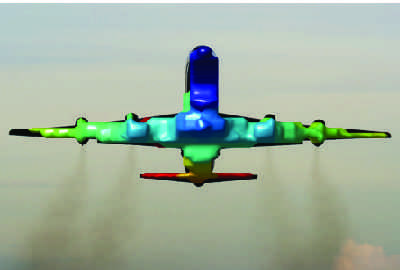} &
\includegraphics[height = 0.041\textheight, width = 0.125\textwidth, keepaspectratio = true]{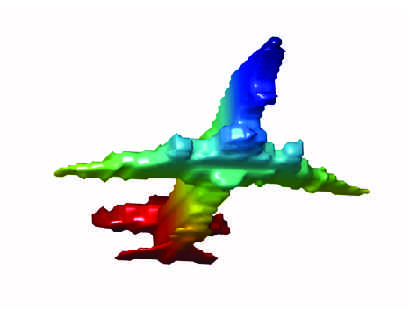} &
\includegraphics[height = 0.041\textheight, width = 0.125\textwidth, keepaspectratio = true]{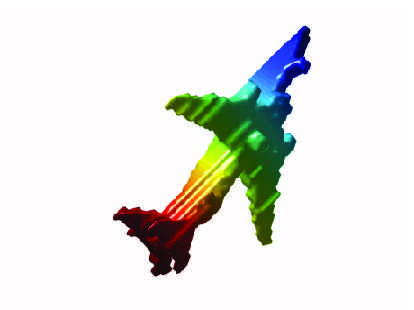}&
\includegraphics[height = 0.041\textheight, width = 0.125\textwidth, keepaspectratio = true]{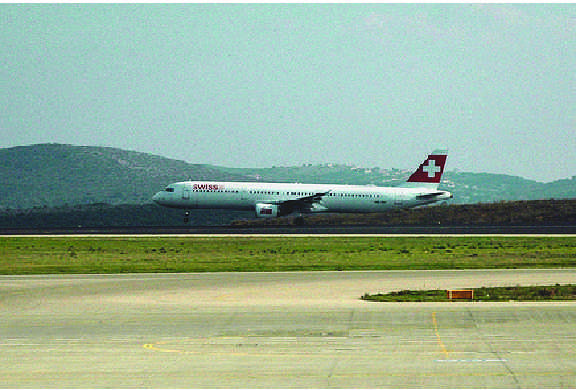} &
\includegraphics[height = 0.041\textheight, width = 0.125\textwidth, keepaspectratio = true]{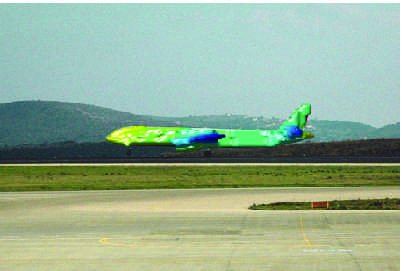} &
\includegraphics[height = 0.041\textheight, width = 0.125\textwidth, keepaspectratio = true]{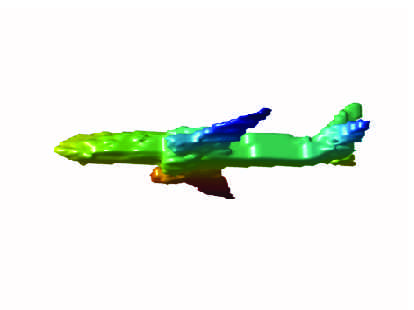} &
\includegraphics[height = 0.041\textheight, width = 0.125\textwidth, keepaspectratio = true]{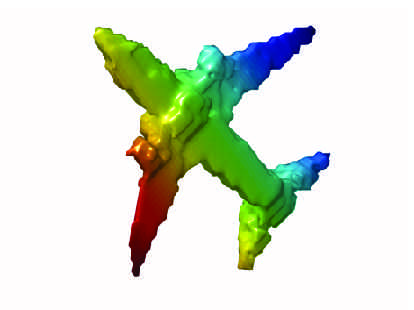}\\
\includegraphics[height = 0.041\textheight, width = 0.125\textwidth, keepaspectratio = true]{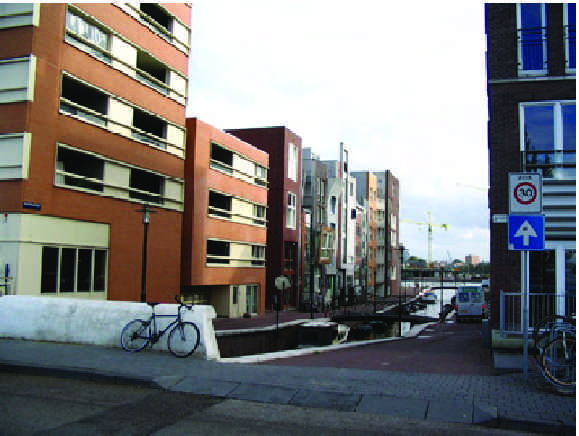} &
\includegraphics[height = 0.041\textheight, width = 0.125\textwidth, keepaspectratio = true]{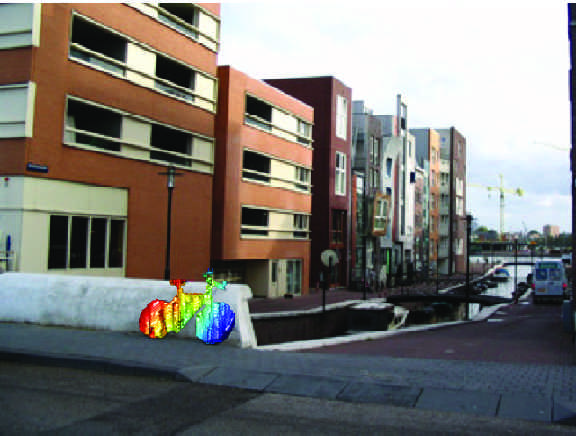} &
\includegraphics[height = 0.041\textheight, width = 0.125\textwidth, keepaspectratio = true]{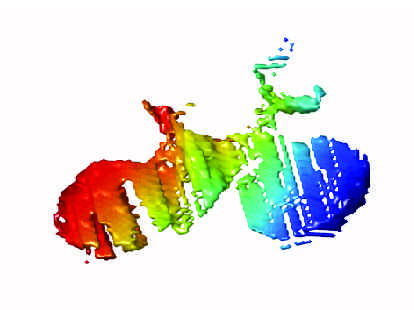} &
\includegraphics[height = 0.041\textheight, width = 0.125\textwidth, keepaspectratio = true]{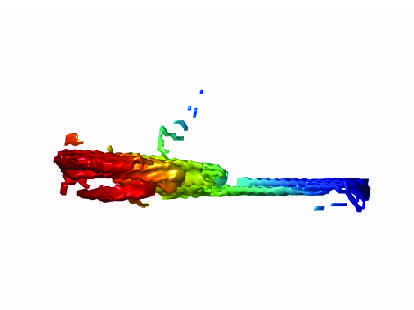}&
\includegraphics[height = 0.041\textheight, width = 0.125\textwidth, keepaspectratio = true]{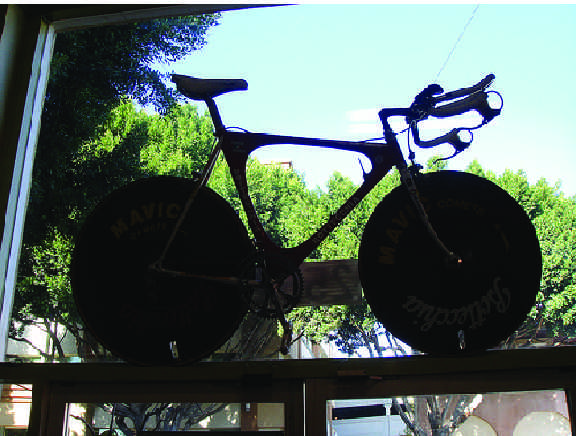} &
\includegraphics[height = 0.041\textheight, width = 0.125\textwidth, keepaspectratio = true]{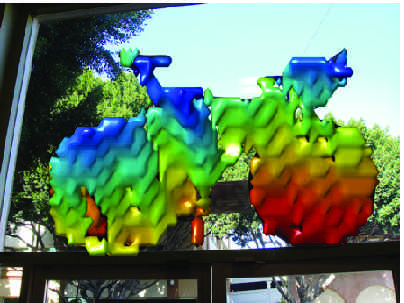} &
\includegraphics[height = 0.041\textheight, width = 0.125\textwidth, keepaspectratio = true]{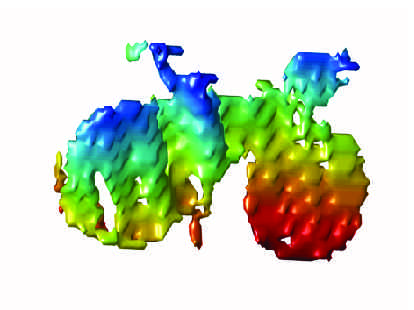} &
\includegraphics[height = 0.041\textheight, width = 0.125\textwidth, keepaspectratio = true]{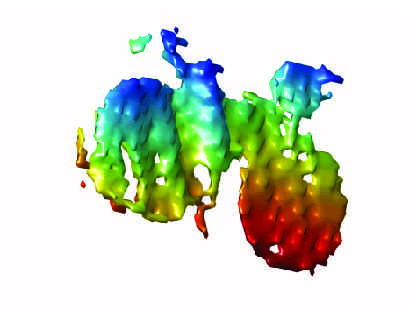}\\
\includegraphics[height = 0.041\textheight, width = 0.125\textwidth, keepaspectratio = true]{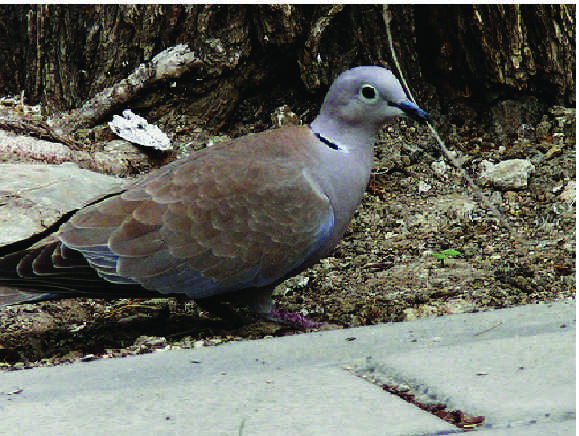} &
\includegraphics[height = 0.041\textheight, width = 0.125\textwidth, keepaspectratio = true]{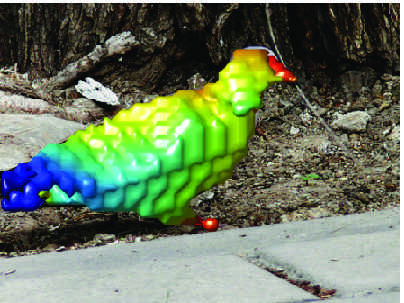} &
\includegraphics[height = 0.041\textheight, width = 0.125\textwidth, keepaspectratio = true]{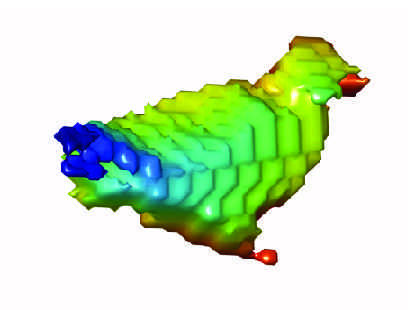} &
\includegraphics[height = 0.041\textheight, width = 0.125\textwidth, keepaspectratio = true]{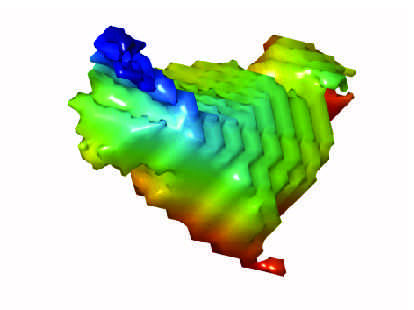}&
\includegraphics[height = 0.041\textheight, width = 0.125\textwidth, keepaspectratio = true]{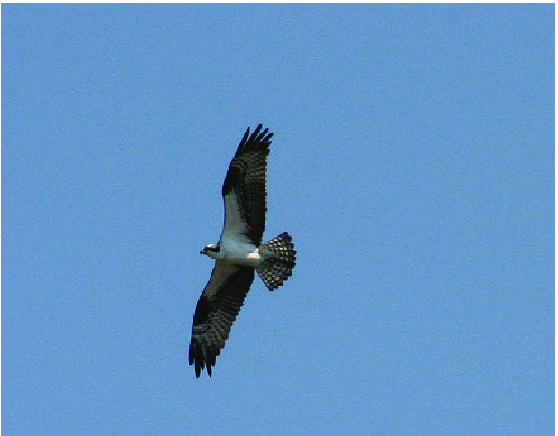} &
\includegraphics[height = 0.041\textheight, width = 0.125\textwidth, keepaspectratio = true]{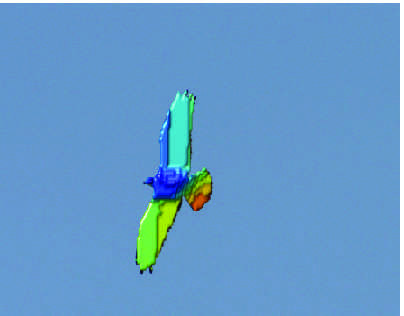} &
\includegraphics[height = 0.041\textheight, width = 0.125\textwidth, keepaspectratio = true]{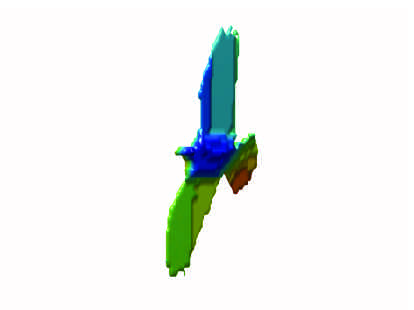} &
\includegraphics[height = 0.041\textheight, width = 0.125\textwidth, keepaspectratio = true]{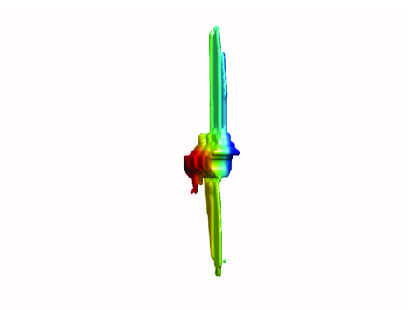}\\
\includegraphics[height = 0.041\textheight, width = 0.125\textwidth, keepaspectratio = true]{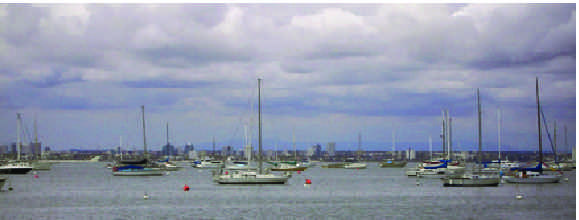} &
\includegraphics[height = 0.041\textheight, width = 0.125\textwidth, keepaspectratio = true]{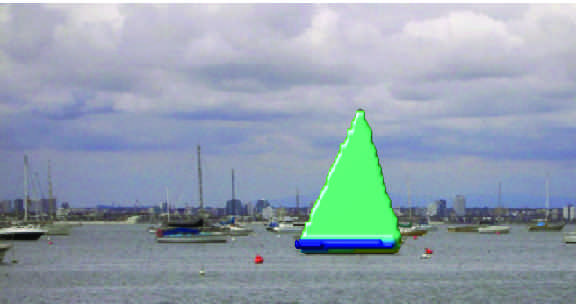} &  
\includegraphics[height = 0.041\textheight, width = 0.125\textwidth, keepaspectratio = true]{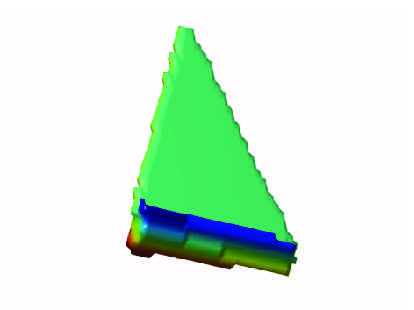} &  
\includegraphics[height = 0.041\textheight, width = 0.125\textwidth, keepaspectratio = true]{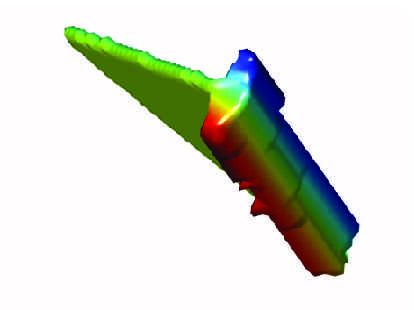}&   
\includegraphics[height = 0.041\textheight, width = 0.125\textwidth, keepaspectratio = true]{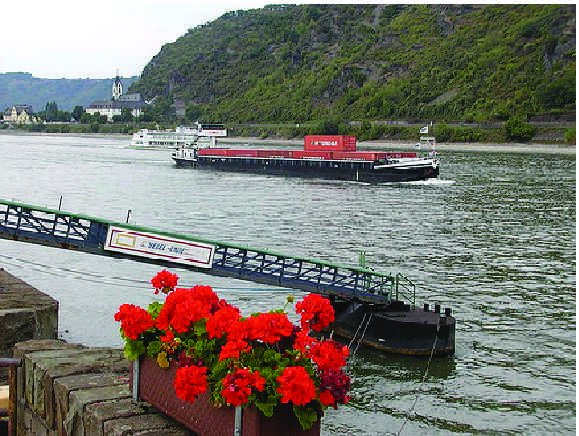} &
\includegraphics[height = 0.041\textheight, width = 0.125\textwidth, keepaspectratio = true]{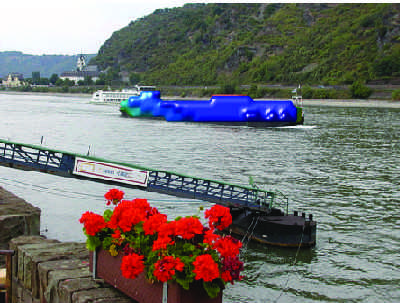} &  
\includegraphics[height = 0.041\textheight, width = 0.125\textwidth, keepaspectratio = true]{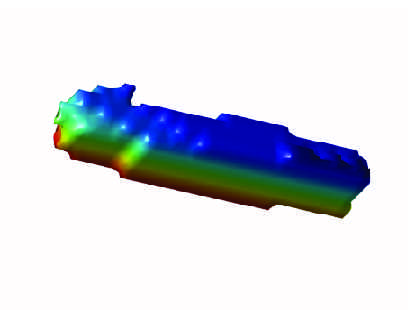} &  
\includegraphics[height = 0.041\textheight, width = 0.125\textwidth, keepaspectratio = true]{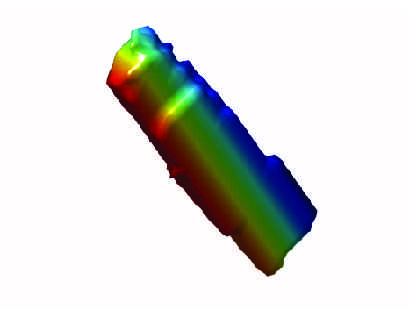}\\  
\includegraphics[height = 0.041\textheight, width = 0.125\textwidth, keepaspectratio = true]{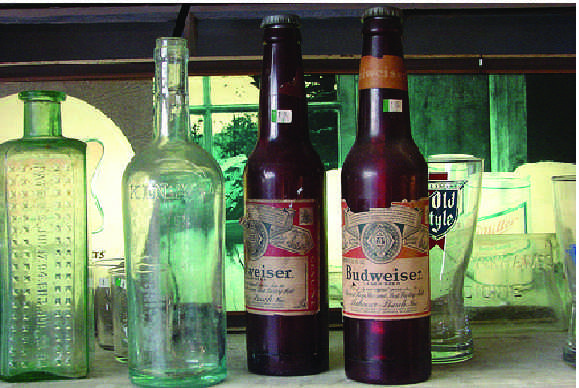} &
\includegraphics[height = 0.041\textheight, width = 0.125\textwidth, keepaspectratio = true]{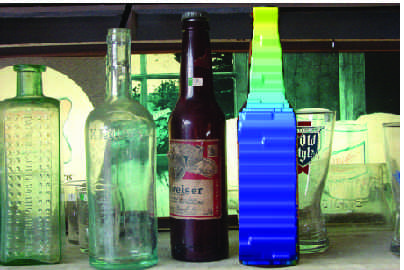} &
\includegraphics[height = 0.041\textheight, width = 0.125\textwidth, keepaspectratio = true]{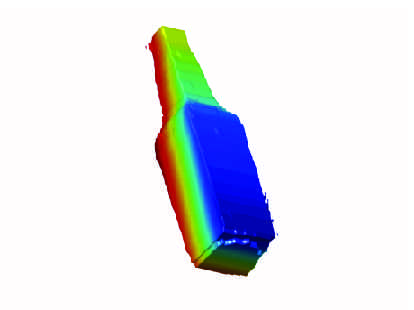} &
\includegraphics[height = 0.041\textheight, width = 0.125\textwidth, keepaspectratio = true]{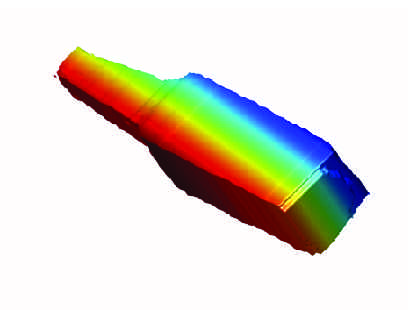}&
\includegraphics[height = 0.041\textheight, width = 0.125\textwidth, keepaspectratio = true]{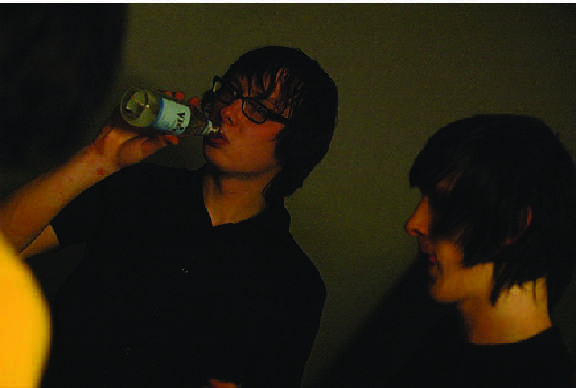} &
\includegraphics[height = 0.041\textheight, width = 0.125\textwidth, keepaspectratio = true]{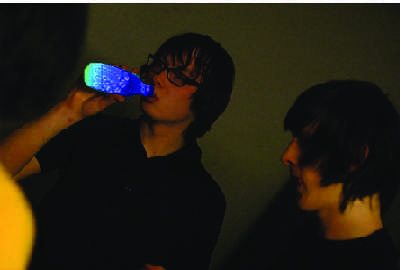} &
\includegraphics[height = 0.041\textheight, width = 0.125\textwidth, keepaspectratio = true]{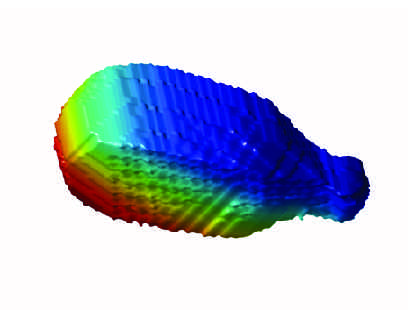} &
\includegraphics[height = 0.041\textheight, width = 0.125\textwidth, keepaspectratio = true]{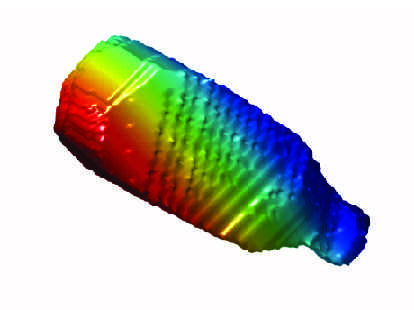}\\
\includegraphics[height = 0.041\textheight, width = 0.125\textwidth, keepaspectratio = true]{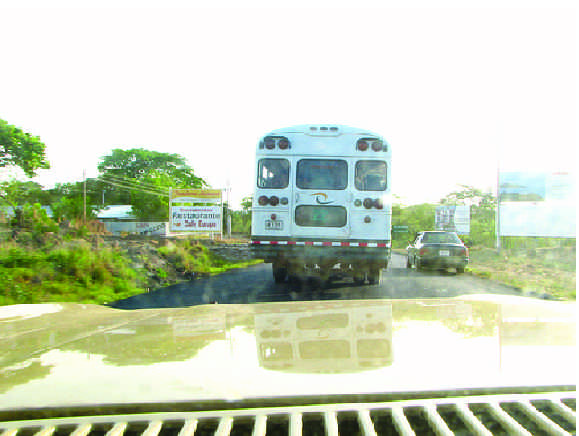} &
\includegraphics[height = 0.041\textheight, width = 0.125\textwidth, keepaspectratio = true]{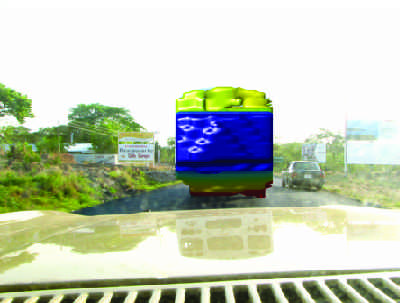} &
\includegraphics[height = 0.041\textheight, width = 0.125\textwidth, keepaspectratio = true]{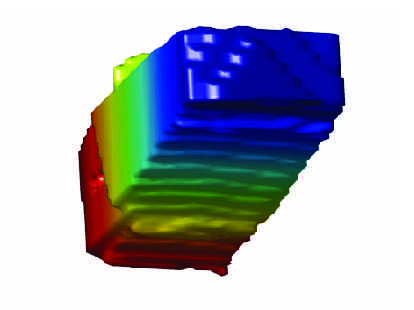} &
\includegraphics[height = 0.041\textheight, width = 0.125\textwidth, keepaspectratio = true]{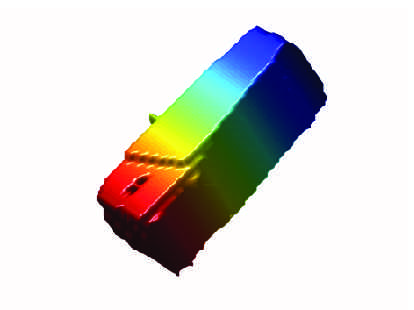}&
\includegraphics[height = 0.041\textheight, width = 0.125\textwidth, keepaspectratio = true]{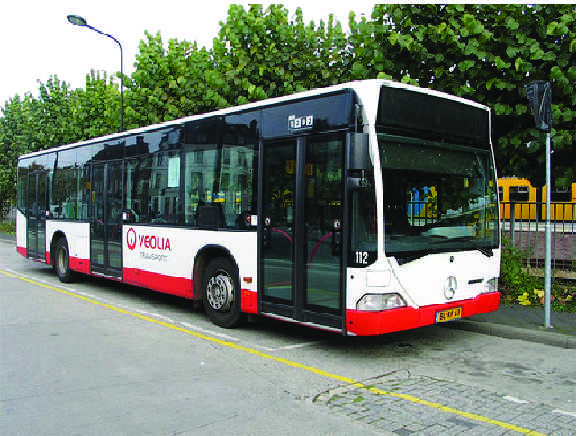} &
\includegraphics[height = 0.041\textheight, width = 0.125\textwidth, keepaspectratio = true]{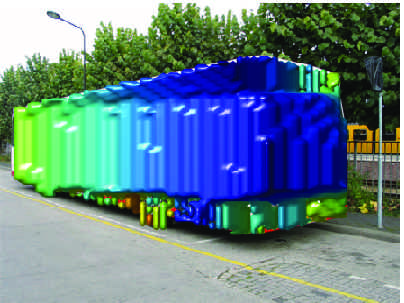} &
\includegraphics[height = 0.041\textheight, width = 0.125\textwidth, keepaspectratio = true]{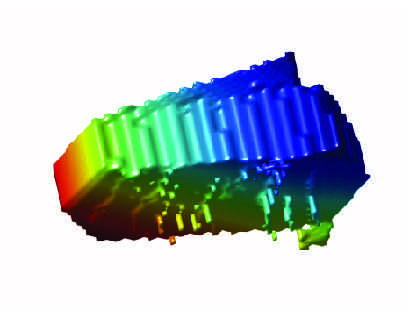} &
\includegraphics[height = 0.041\textheight, width = 0.125\textwidth, keepaspectratio = true]{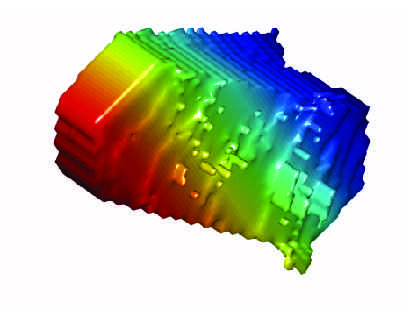}\\
\includegraphics[height = 0.041\textheight, width = 0.125\textwidth, keepaspectratio = true]{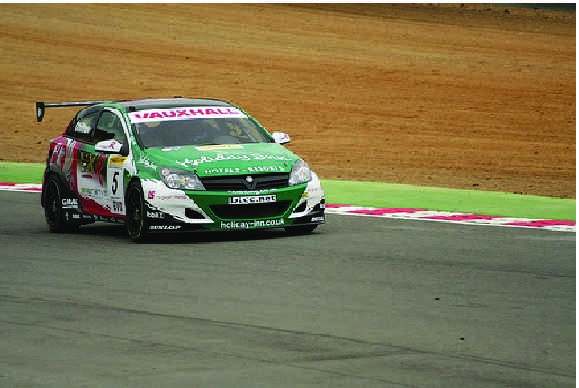} &
\includegraphics[height = 0.041\textheight, width = 0.125\textwidth, keepaspectratio = true]{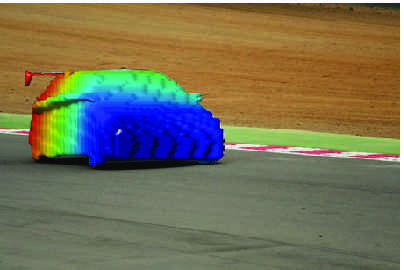} &
\includegraphics[height = 0.041\textheight, width = 0.125\textwidth, keepaspectratio = true]{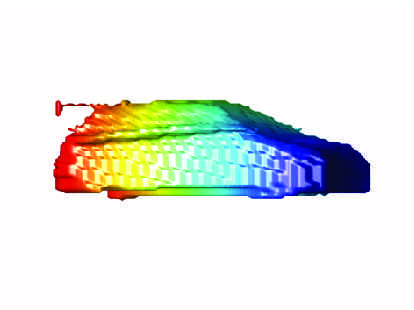} &
\includegraphics[height = 0.041\textheight, width = 0.125\textwidth, keepaspectratio = true]{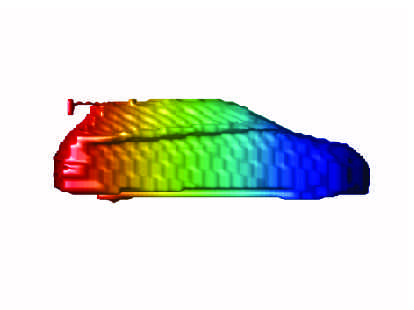}&
\includegraphics[height = 0.041\textheight, width = 0.125\textwidth, keepaspectratio = true]{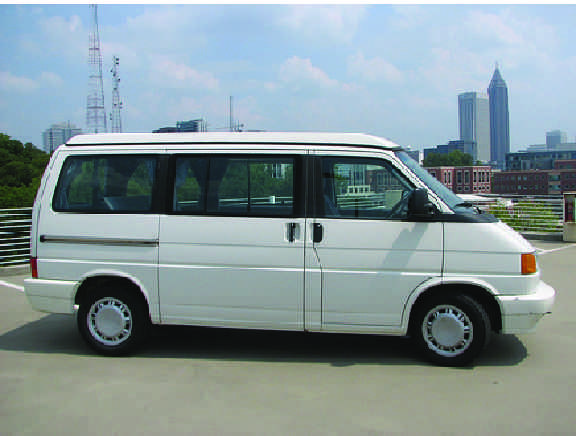} &
\includegraphics[height = 0.041\textheight, width = 0.125\textwidth, keepaspectratio = true]{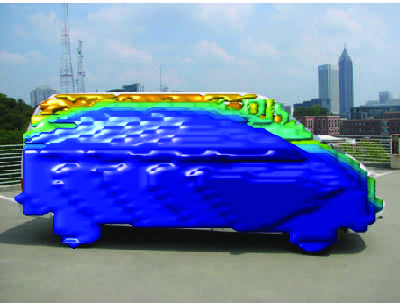} &
\includegraphics[height = 0.041\textheight, width = 0.125\textwidth, keepaspectratio = true]{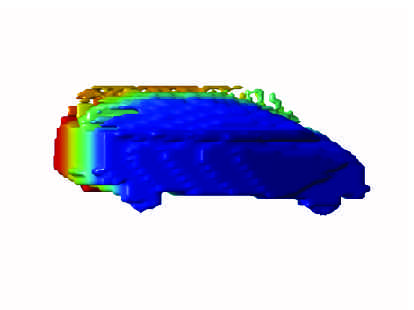} &
\includegraphics[height = 0.041\textheight, width = 0.125\textwidth, keepaspectratio = true]{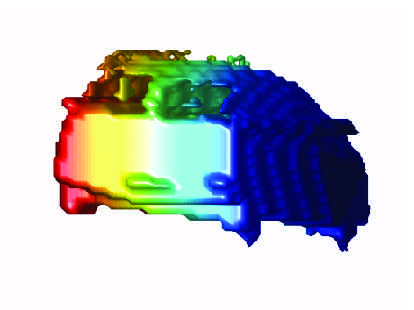}\\
\includegraphics[height = 0.041\textheight, width = 0.125\textwidth, keepaspectratio = true]{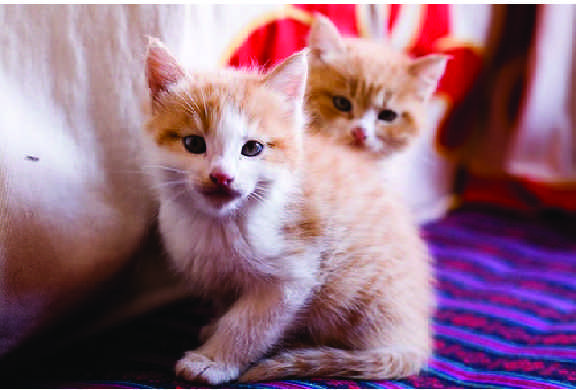} &
\includegraphics[height = 0.041\textheight, width = 0.125\textwidth, keepaspectratio = true]{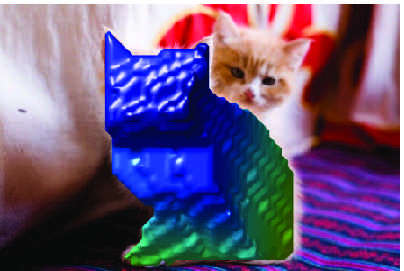} &
\includegraphics[height = 0.041\textheight, width = 0.125\textwidth, keepaspectratio = true]{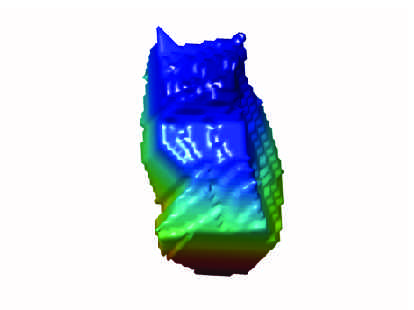} &
\includegraphics[height = 0.041\textheight, width = 0.125\textwidth, keepaspectratio = true]{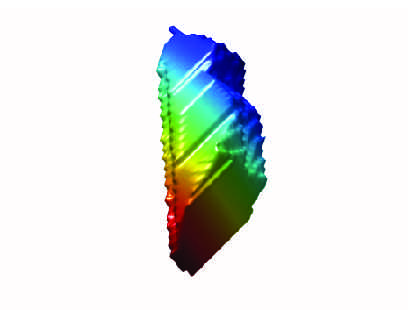}&
\includegraphics[height = 0.041\textheight, width = 0.125\textwidth, keepaspectratio = true]{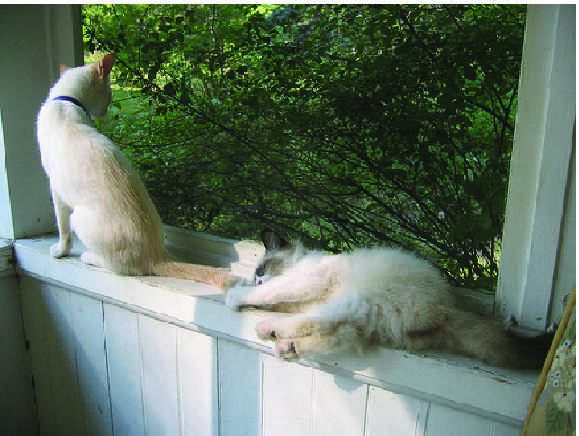} &
\includegraphics[height = 0.041\textheight, width = 0.125\textwidth, keepaspectratio = true]{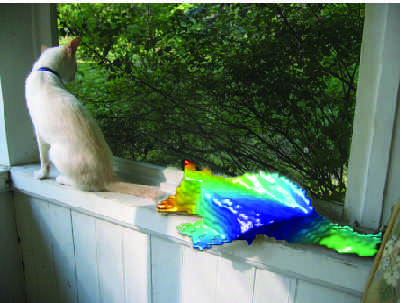} &
\includegraphics[height = 0.041\textheight, width = 0.125\textwidth, keepaspectratio = true]{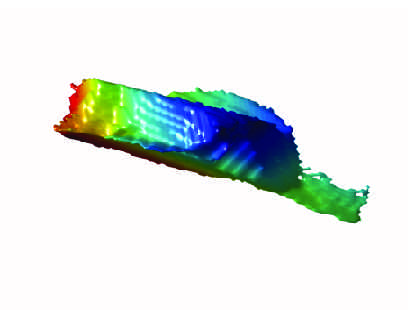} &
\includegraphics[height = 0.041\textheight, width = 0.125\textwidth, keepaspectratio = true]{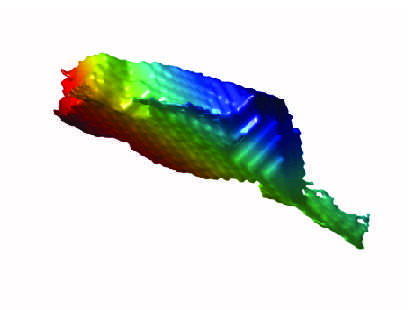}\\
\includegraphics[height = 0.041\textheight, width = 0.125\textwidth, keepaspectratio = true]{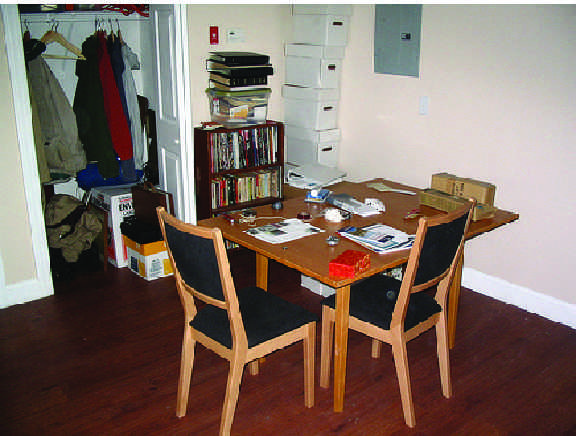} &
\includegraphics[height = 0.041\textheight, width = 0.125\textwidth, keepaspectratio = true]{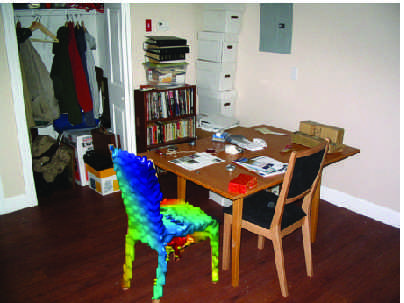} &
\includegraphics[height = 0.041\textheight, width = 0.125\textwidth, keepaspectratio = true]{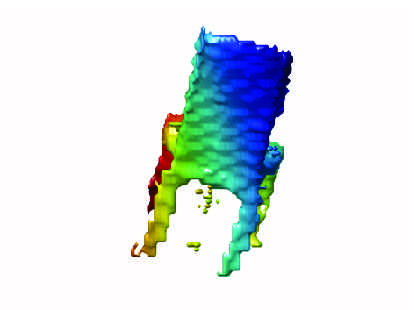} &
\includegraphics[height = 0.041\textheight, width = 0.125\textwidth, keepaspectratio = true]{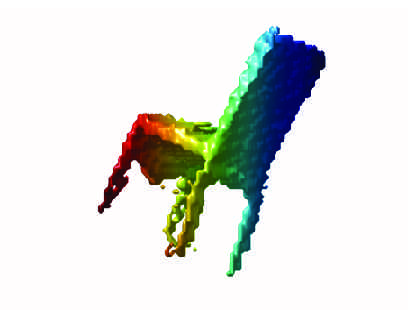}&
\includegraphics[height = 0.041\textheight, width = 0.125\textwidth, keepaspectratio = true]{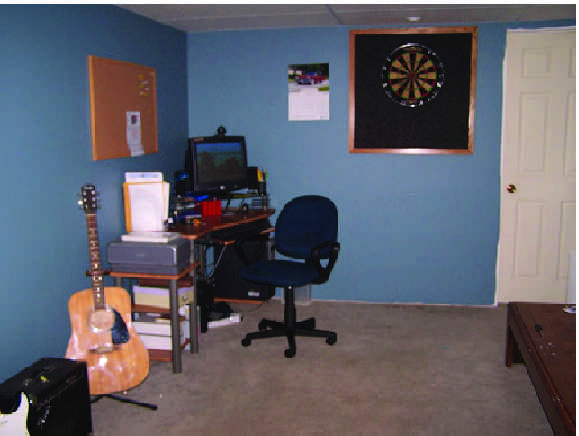} &
\includegraphics[height = 0.041\textheight, width = 0.125\textwidth, keepaspectratio = true]{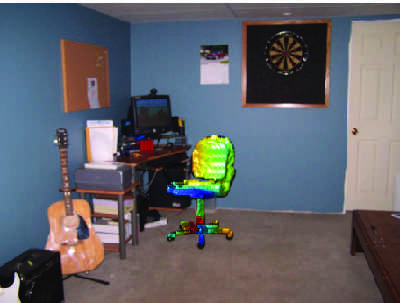} &
\includegraphics[height = 0.041\textheight, width = 0.125\textwidth, keepaspectratio = true]{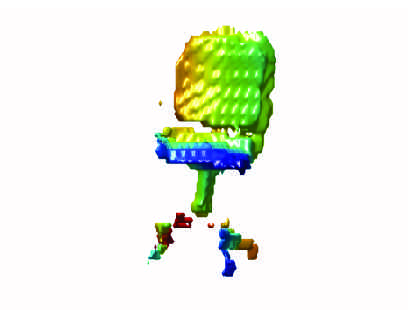} &
\includegraphics[height = 0.041\textheight, width = 0.125\textwidth, keepaspectratio = true]{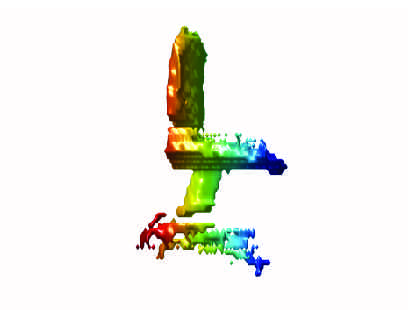}\\
\includegraphics[height = 0.041\textheight, width = 0.125\textwidth, keepaspectratio = true]{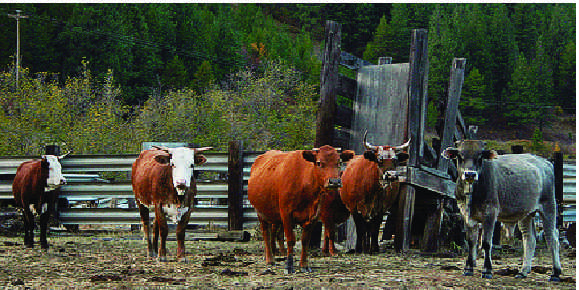} &
\includegraphics[height = 0.041\textheight, width = 0.125\textwidth, keepaspectratio = true]{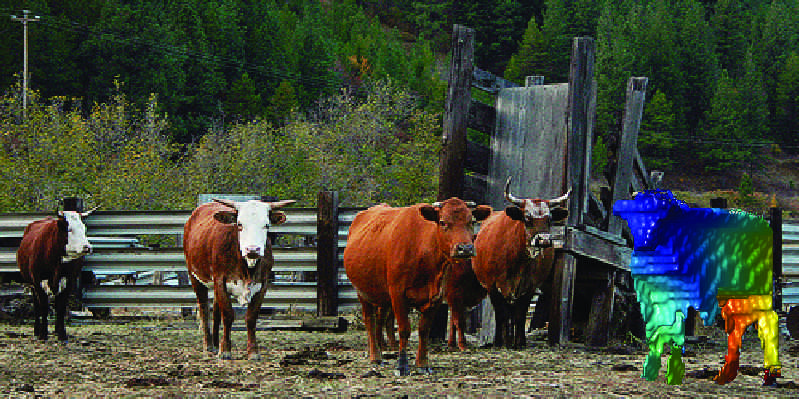} &
\includegraphics[height = 0.041\textheight, width = 0.125\textwidth, keepaspectratio = true]{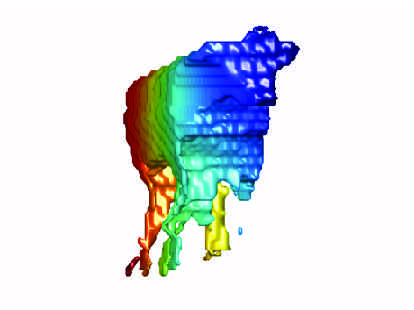} &
\includegraphics[height = 0.041\textheight, width = 0.125\textwidth, keepaspectratio = true]{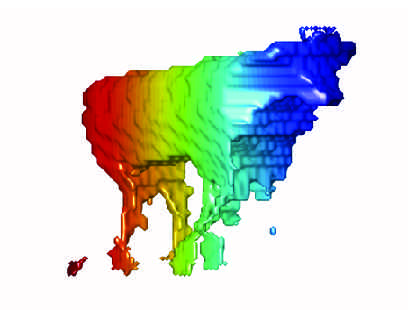}&
\includegraphics[height = 0.041\textheight, width = 0.125\textwidth, keepaspectratio = true]{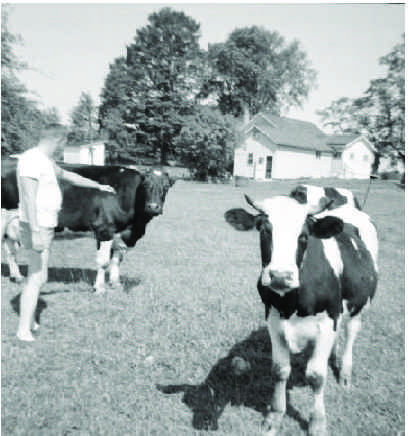} &
\includegraphics[height = 0.041\textheight, width = 0.125\textwidth, keepaspectratio = true]{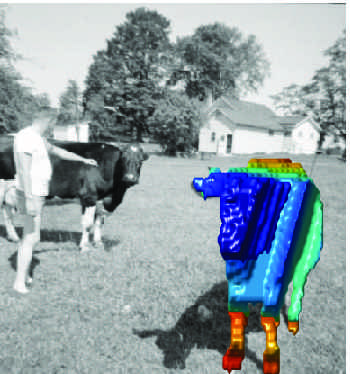} &
\includegraphics[height = 0.041\textheight, width = 0.125\textwidth, keepaspectratio = true]{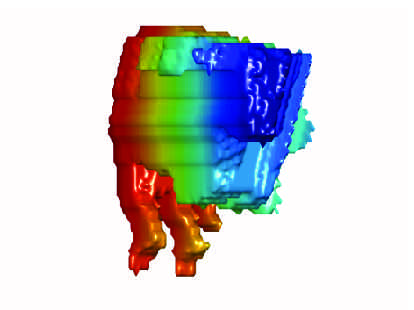} &
\includegraphics[height = 0.041\textheight, width = 0.125\textwidth, keepaspectratio = true]{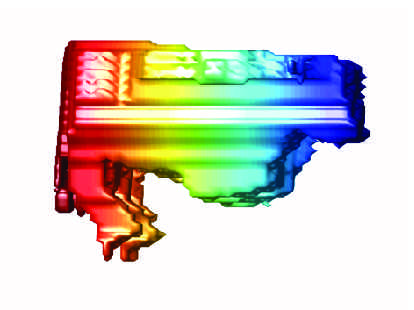}\\
\includegraphics[height = 0.041\textheight, width = 0.125\textwidth, keepaspectratio = true]{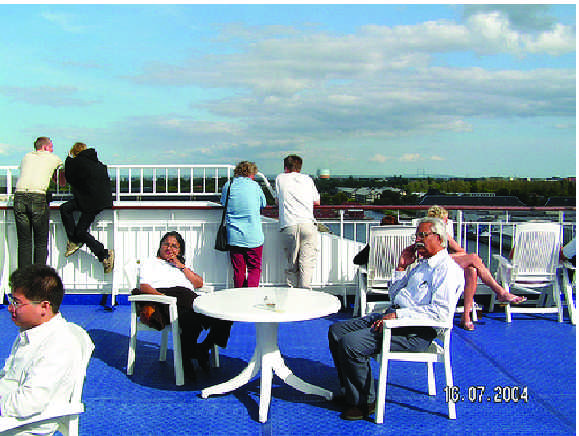} &
\includegraphics[height = 0.041\textheight, width = 0.125\textwidth, keepaspectratio = true]{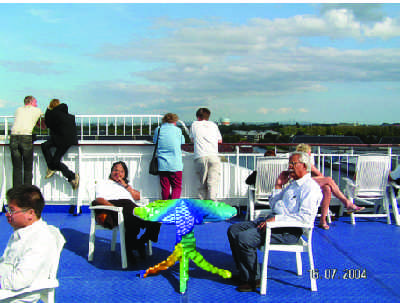} &
\includegraphics[height = 0.041\textheight, width = 0.125\textwidth, keepaspectratio = true]{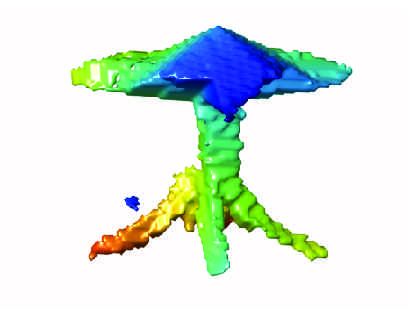} &
\includegraphics[height = 0.041\textheight, width = 0.125\textwidth, keepaspectratio = true]{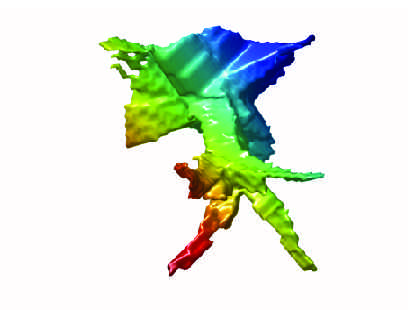}&
\includegraphics[height = 0.041\textheight, width = 0.125\textwidth, keepaspectratio = true]{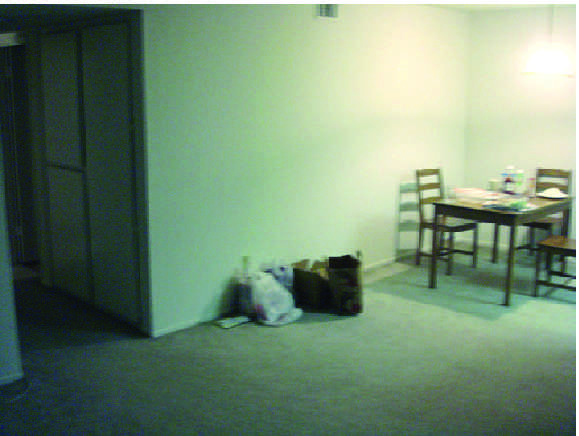} &
\includegraphics[height = 0.041\textheight, width = 0.125\textwidth, keepaspectratio = true]{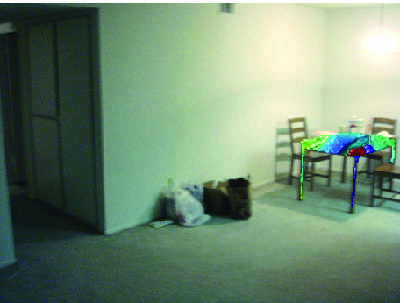} &
\includegraphics[height = 0.041\textheight, width = 0.125\textwidth, keepaspectratio = true]{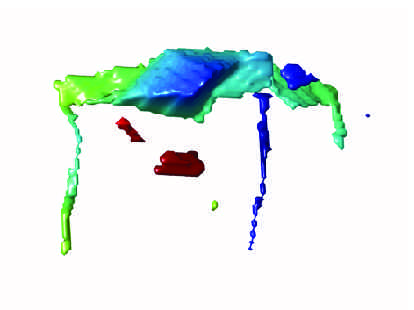} &
\includegraphics[height = 0.041\textheight, width = 0.125\textwidth, keepaspectratio = true]{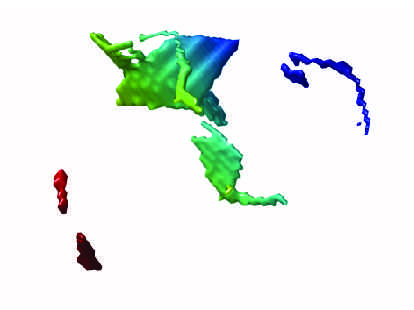}\\
\includegraphics[height = 0.041\textheight, width = 0.125\textwidth, keepaspectratio = true]{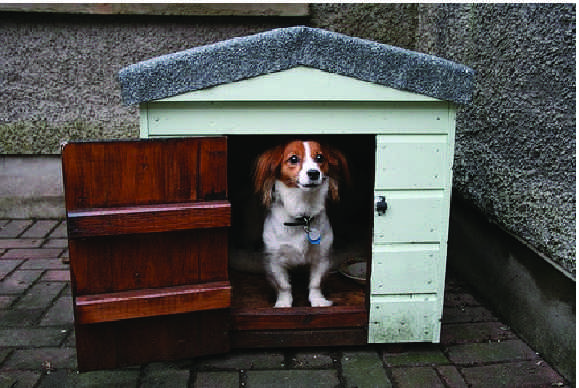} &
\includegraphics[height = 0.041\textheight, width = 0.125\textwidth, keepaspectratio = true]{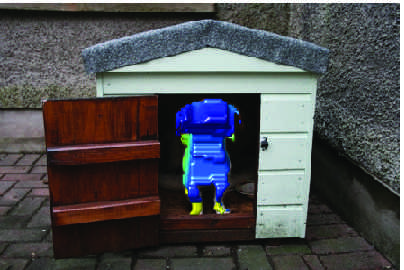} &
\includegraphics[height = 0.041\textheight, width = 0.125\textwidth, keepaspectratio = true]{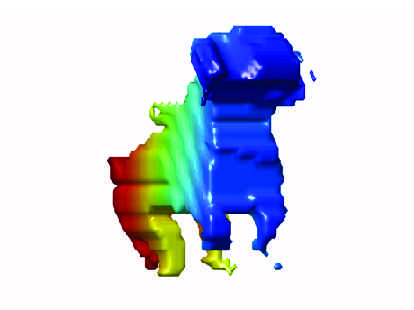} &
\includegraphics[height = 0.041\textheight, width = 0.125\textwidth, keepaspectratio = true]{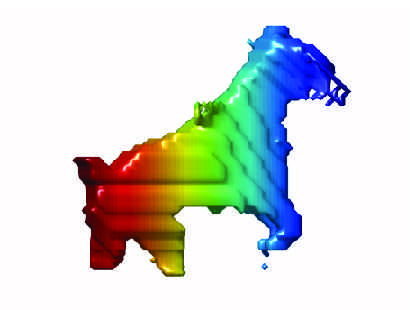}&
\includegraphics[height = 0.041\textheight, width = 0.125\textwidth, keepaspectratio = true]{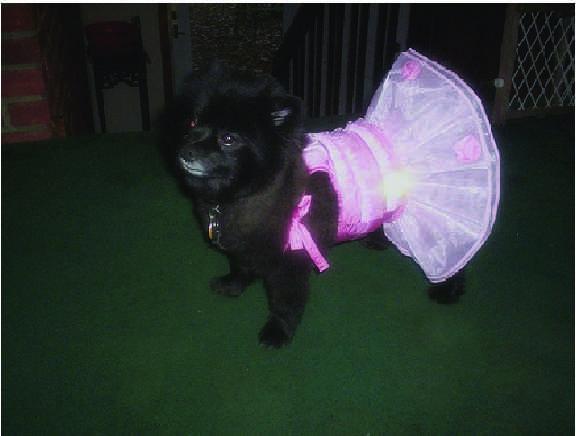} &
\includegraphics[height = 0.041\textheight, width = 0.125\textwidth, keepaspectratio = true]{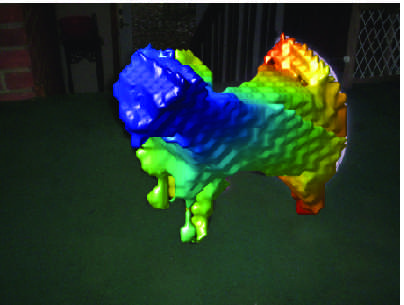} &
\includegraphics[height = 0.041\textheight, width = 0.125\textwidth, keepaspectratio = true]{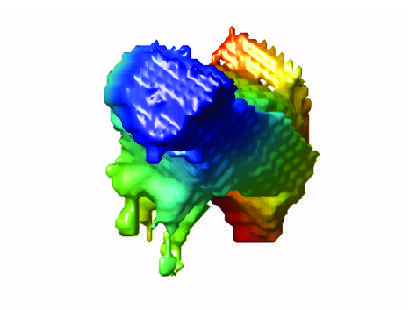} &
\includegraphics[height = 0.041\textheight, width = 0.125\textwidth, keepaspectratio = true]{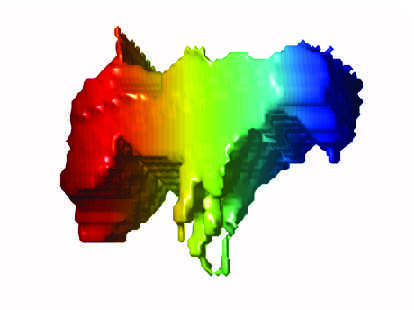}\\
\includegraphics[height = 0.041\textheight, width = 0.125\textwidth, keepaspectratio = true]{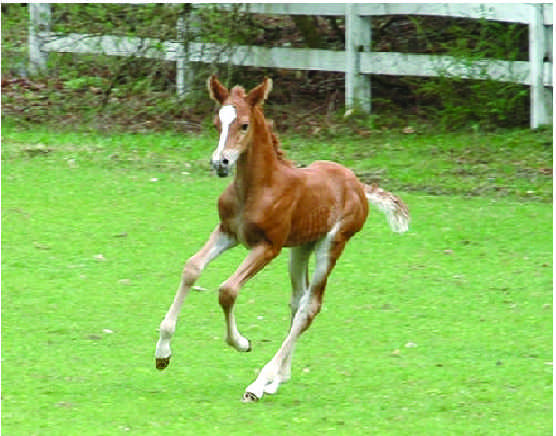} &
\includegraphics[height = 0.041\textheight, width = 0.125\textwidth, keepaspectratio = true]{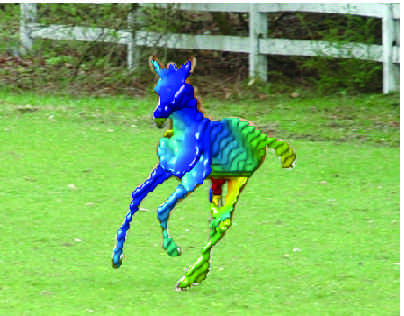} &
\includegraphics[height = 0.041\textheight, width = 0.125\textwidth, keepaspectratio = true]{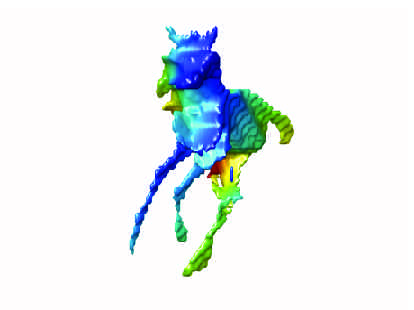} &
\includegraphics[height = 0.041\textheight, width = 0.125\textwidth, keepaspectratio = true]{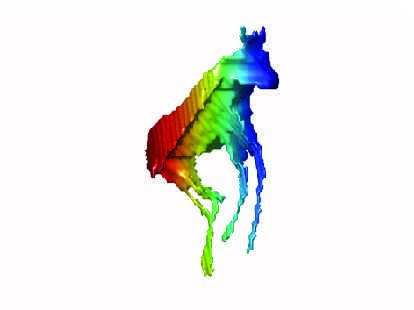}&
\includegraphics[height = 0.041\textheight, width = 0.125\textwidth, keepaspectratio = true]{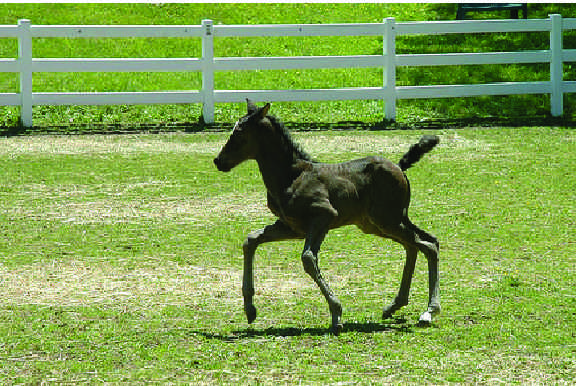} &
\includegraphics[height = 0.041\textheight, width = 0.125\textwidth, keepaspectratio = true]{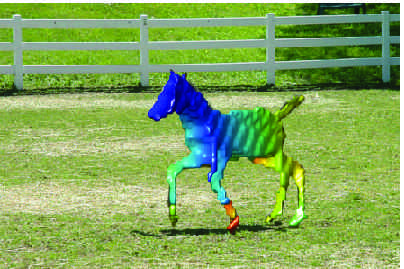} &
\includegraphics[height = 0.041\textheight, width = 0.125\textwidth, keepaspectratio = true]{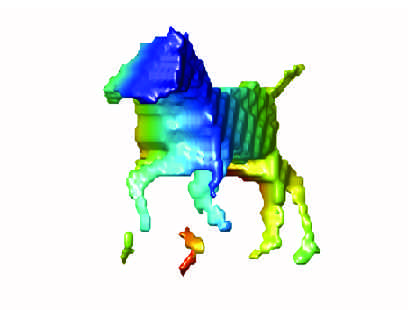} &
\includegraphics[height = 0.041\textheight, width = 0.125\textwidth, keepaspectratio = true]{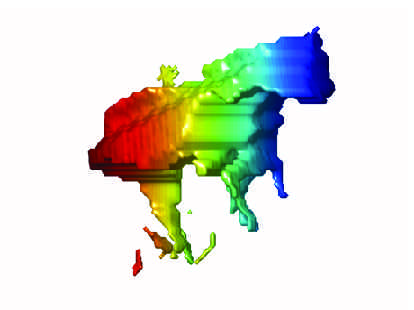}\\
\includegraphics[height = 0.041\textheight, width = 0.125\textwidth, keepaspectratio = true]{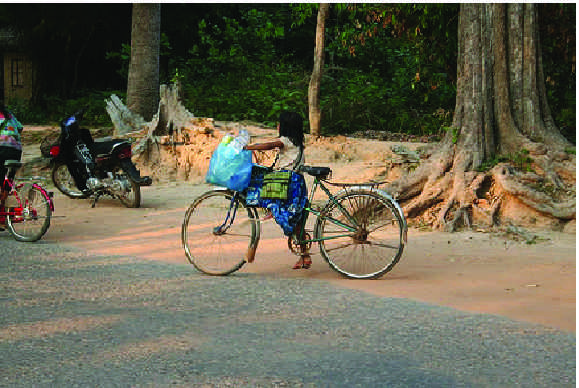} &
\includegraphics[height = 0.041\textheight, width = 0.125\textwidth, keepaspectratio = true]{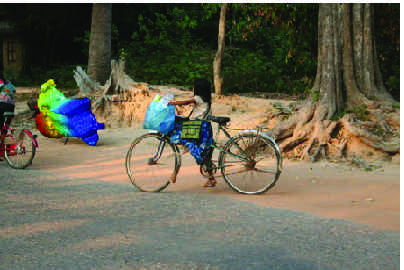} &
\includegraphics[height = 0.041\textheight, width = 0.125\textwidth, keepaspectratio = true]{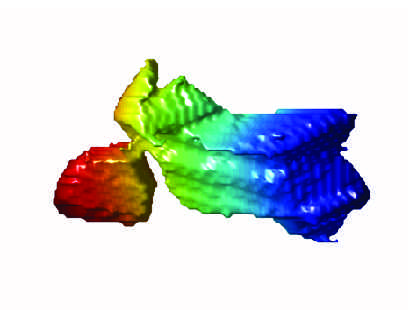} &
\includegraphics[height = 0.041\textheight, width = 0.125\textwidth, keepaspectratio = true]{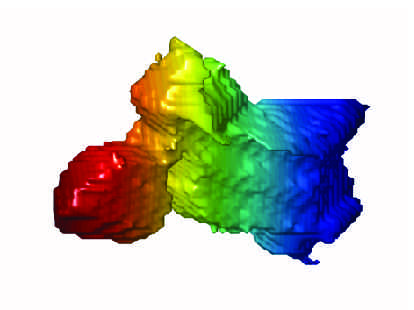}&
\includegraphics[height = 0.041\textheight, width = 0.125\textwidth, keepaspectratio = true]{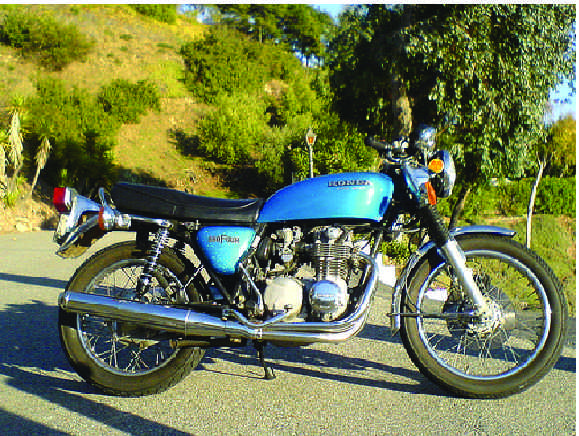} &
\includegraphics[height = 0.041\textheight, width = 0.125\textwidth, keepaspectratio = true]{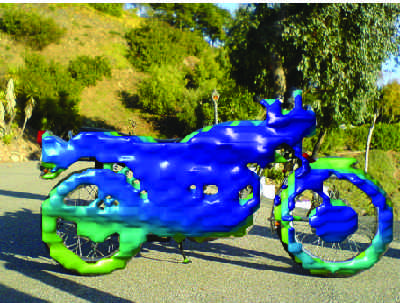} &
\includegraphics[height = 0.041\textheight, width = 0.125\textwidth, keepaspectratio = true]{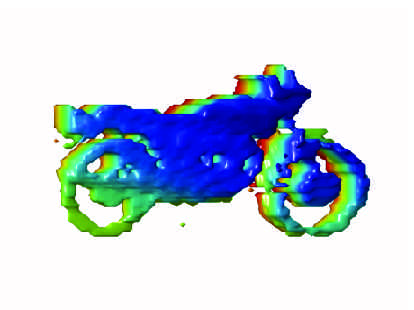} &
\includegraphics[height = 0.041\textheight, width = 0.125\textwidth, keepaspectratio = true]{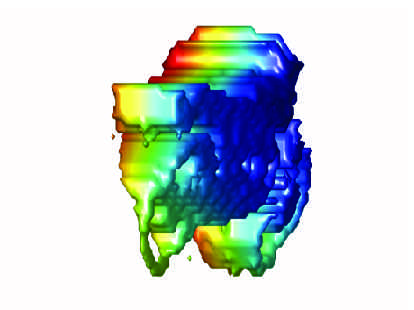}\\
\includegraphics[height = 0.041\textheight, width = 0.125\textwidth, keepaspectratio = true]{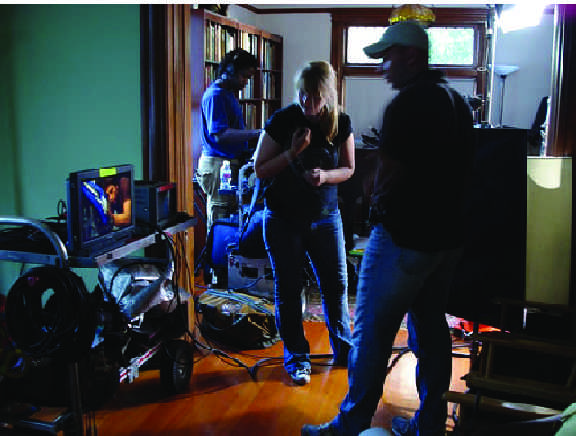} &
\includegraphics[height = 0.041\textheight, width = 0.125\textwidth, keepaspectratio = true]{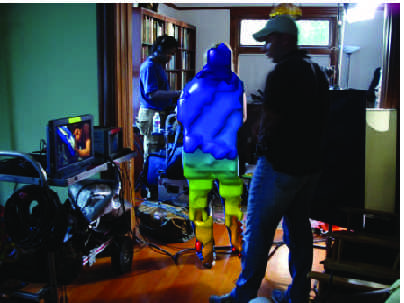} &
\includegraphics[height = 0.041\textheight, width = 0.125\textwidth, keepaspectratio = true]{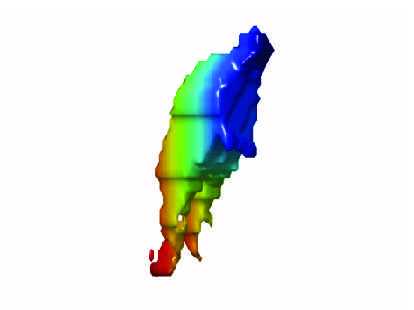} &
\includegraphics[height = 0.041\textheight, width = 0.125\textwidth, keepaspectratio = true]{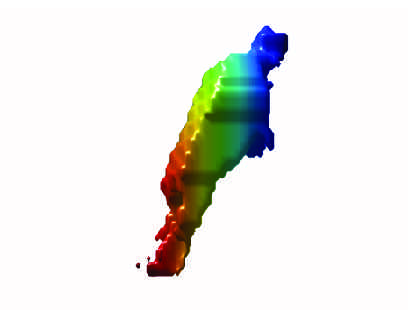}&
\includegraphics[height = 0.041\textheight, width = 0.125\textwidth, keepaspectratio = true]{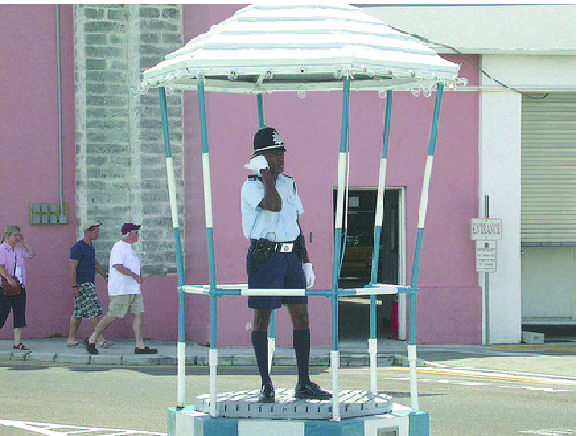} &
\includegraphics[height = 0.041\textheight, width = 0.125\textwidth, keepaspectratio = true]{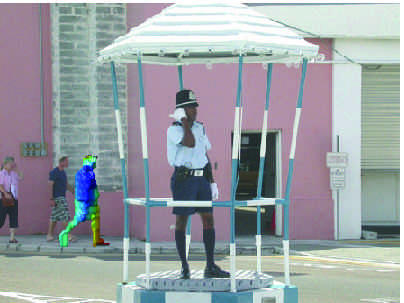} &
\includegraphics[height = 0.041\textheight, width = 0.125\textwidth, keepaspectratio = true]{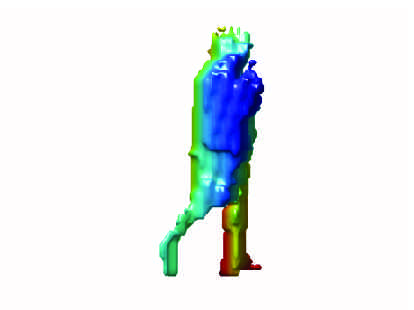} &
\includegraphics[height = 0.041\textheight, width = 0.125\textwidth, keepaspectratio = true]{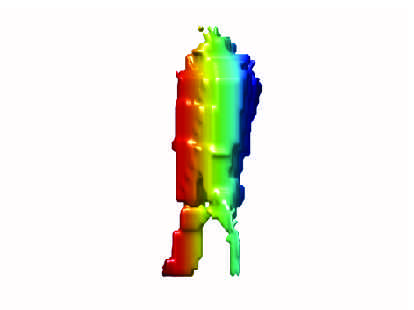}\\
\includegraphics[height = 0.041\textheight, width = 0.125\textwidth, keepaspectratio = true]{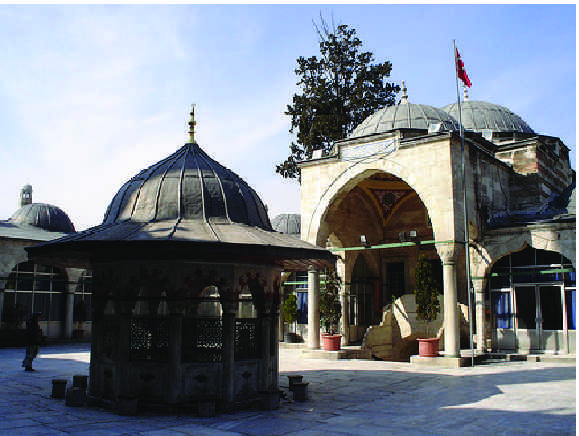} &
\includegraphics[height = 0.041\textheight, width = 0.125\textwidth, keepaspectratio = true]{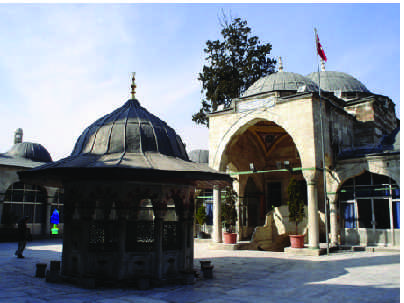} &
\includegraphics[height = 0.041\textheight, width = 0.125\textwidth, keepaspectratio = true]{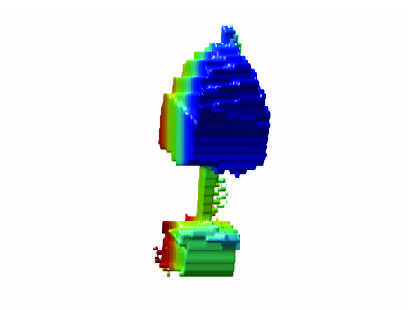} &
\includegraphics[height = 0.041\textheight, width = 0.125\textwidth, keepaspectratio = true]{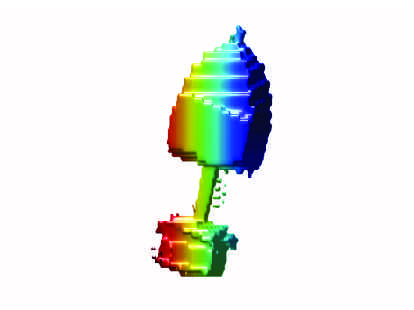}&
\includegraphics[height = 0.041\textheight, width = 0.125\textwidth, keepaspectratio = true]{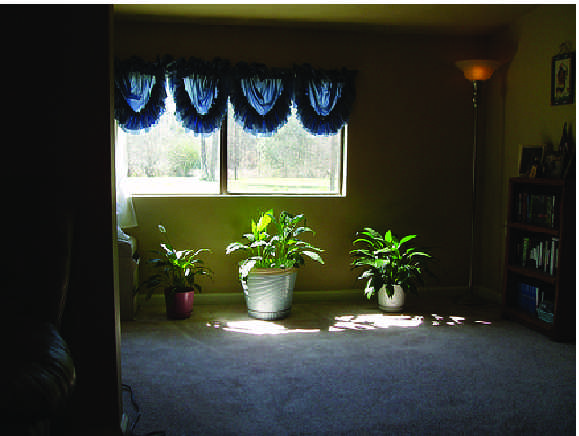} &
\includegraphics[height = 0.041\textheight, width = 0.125\textwidth, keepaspectratio = true]{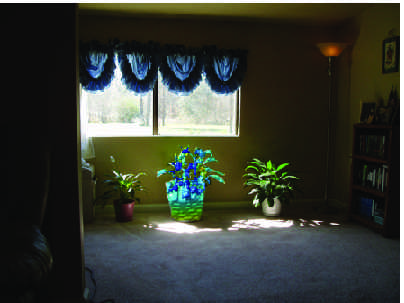} &
\includegraphics[height = 0.041\textheight, width = 0.125\textwidth, keepaspectratio = true]{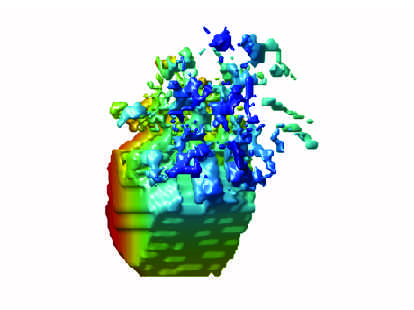} &
\includegraphics[height = 0.041\textheight, width = 0.125\textwidth, keepaspectratio = true]{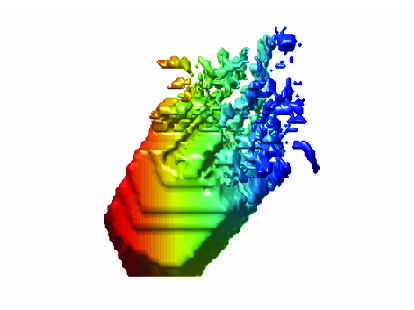}\\
\includegraphics[height = 0.041\textheight, width = 0.125\textwidth, keepaspectratio = true]{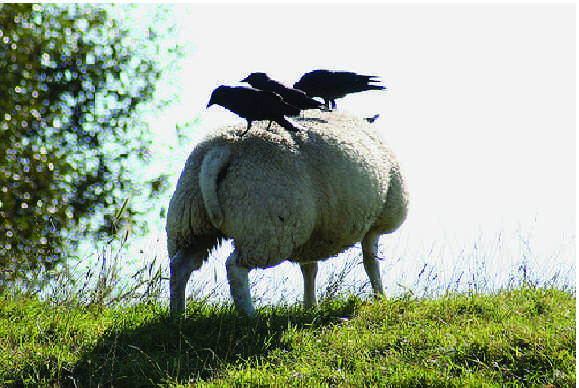} &
\includegraphics[height = 0.041\textheight, width = 0.125\textwidth, keepaspectratio = true]{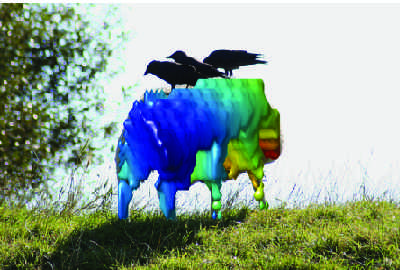} &
\includegraphics[height = 0.041\textheight, width = 0.125\textwidth, keepaspectratio = true]{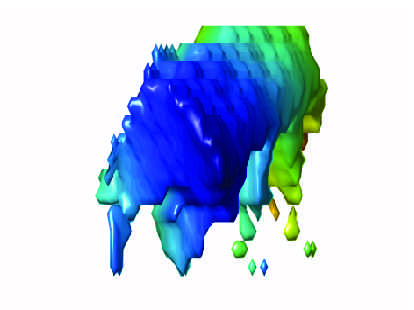} &
\includegraphics[height = 0.041\textheight, width = 0.125\textwidth, keepaspectratio = true]{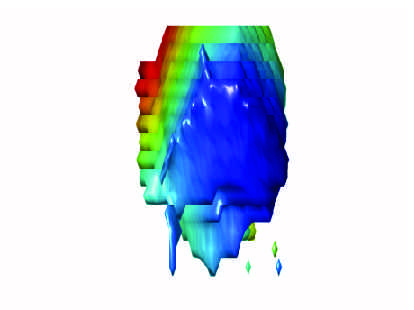}&
\includegraphics[height = 0.041\textheight, width = 0.125\textwidth, keepaspectratio = true]{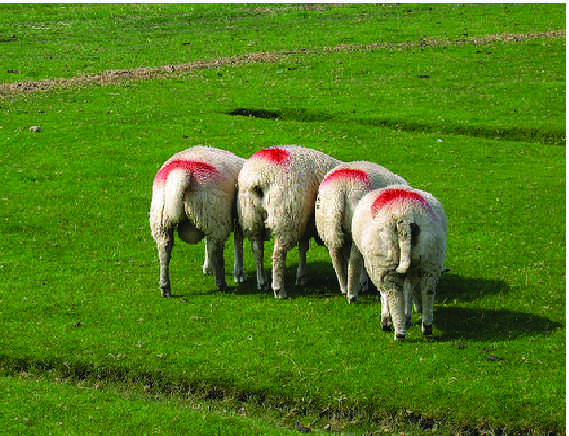} &
\includegraphics[height = 0.041\textheight, width = 0.125\textwidth, keepaspectratio = true]{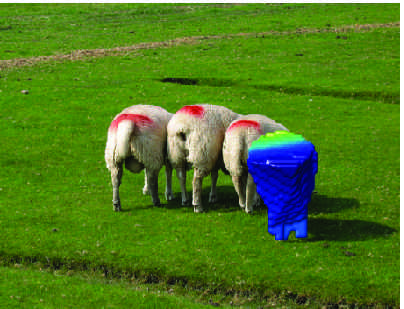} &
\includegraphics[height = 0.041\textheight, width = 0.125\textwidth, keepaspectratio = true]{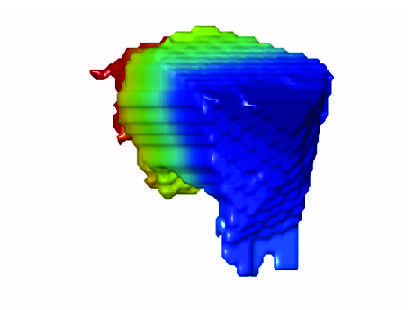} &
\includegraphics[height = 0.041\textheight, width = 0.125\textwidth, keepaspectratio = true]{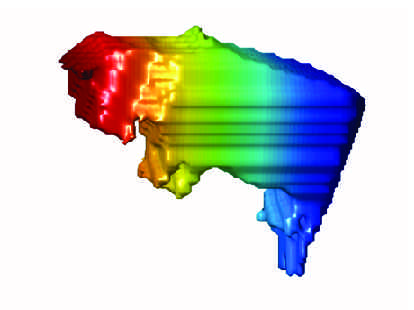}\\
\includegraphics[height = 0.041\textheight, width = 0.125\textwidth, keepaspectratio = true]{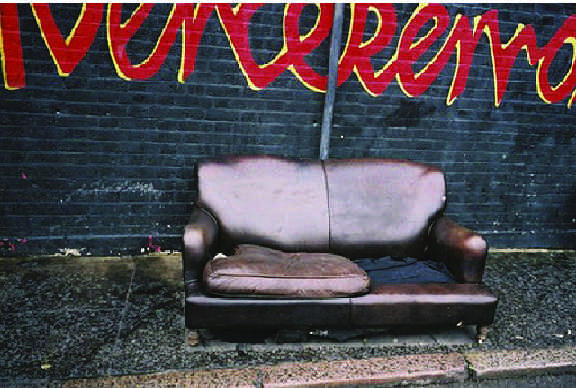} &
\includegraphics[height = 0.041\textheight, width = 0.125\textwidth, keepaspectratio = true]{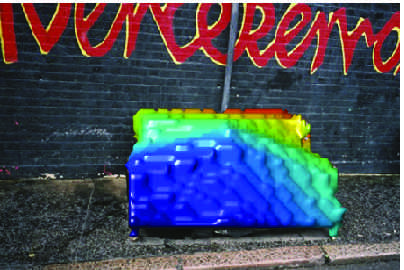} &
\includegraphics[height = 0.041\textheight, width = 0.125\textwidth, keepaspectratio = true]{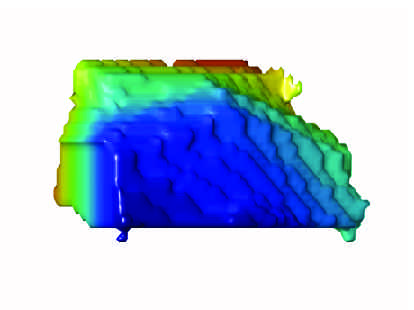} &
\includegraphics[height = 0.041\textheight, width = 0.125\textwidth, keepaspectratio = true]{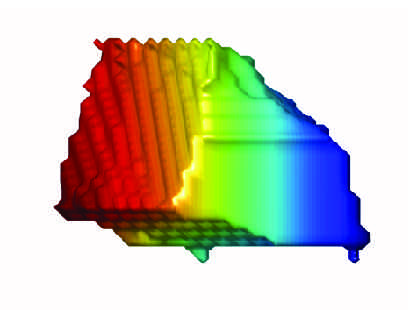}&
\includegraphics[height = 0.041\textheight, width = 0.125\textwidth, keepaspectratio = true]{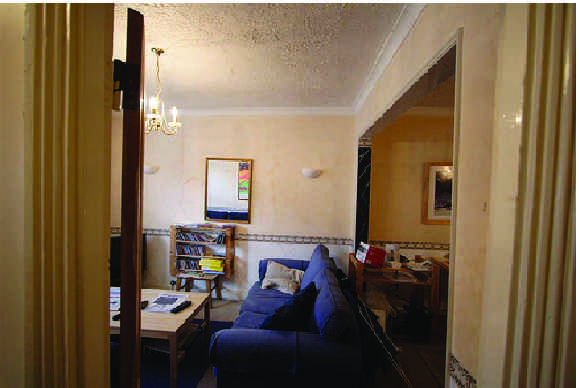} &
\includegraphics[height = 0.041\textheight, width = 0.125\textwidth, keepaspectratio = true]{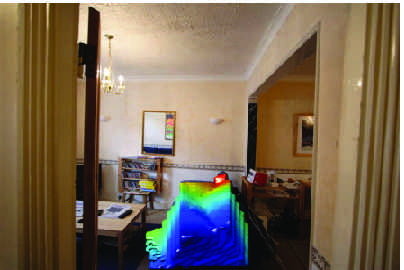} &
\includegraphics[height = 0.041\textheight, width = 0.125\textwidth, keepaspectratio = true]{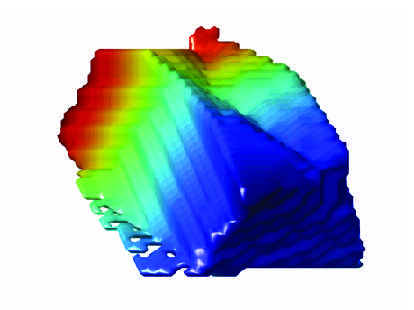} &
\includegraphics[height = 0.041\textheight, width = 0.125\textwidth, keepaspectratio = true]{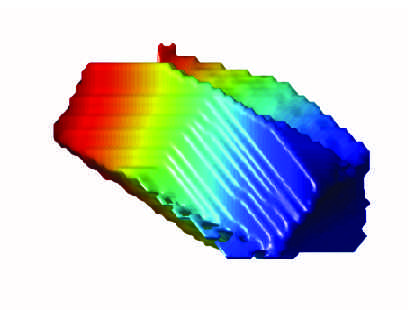}\\
\includegraphics[height = 0.041\textheight, width = 0.125\textwidth, keepaspectratio = true]{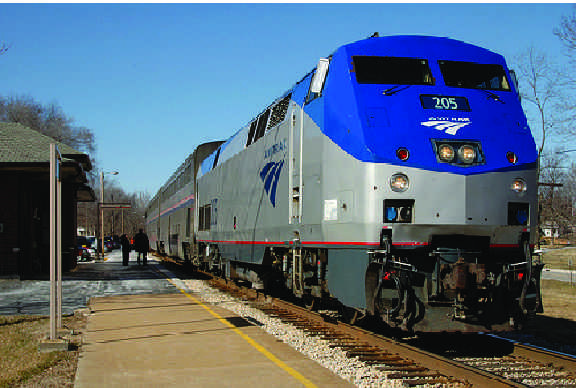} &
\includegraphics[height = 0.041\textheight, width = 0.125\textwidth, keepaspectratio = true]{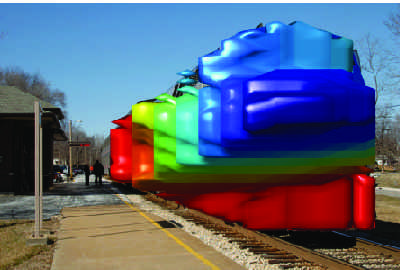} &
\includegraphics[height = 0.041\textheight, width = 0.125\textwidth, keepaspectratio = true]{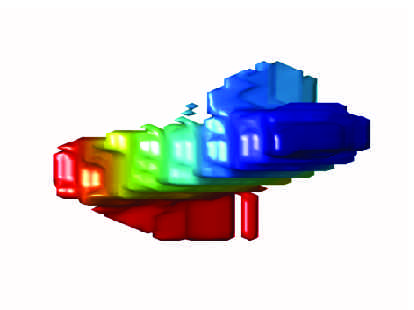} &
\includegraphics[height = 0.041\textheight, width = 0.125\textwidth, keepaspectratio = true]{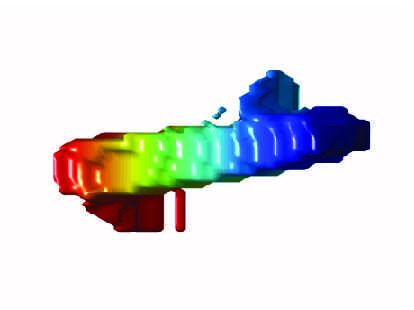}&
\includegraphics[height = 0.041\textheight, width = 0.125\textwidth, keepaspectratio = true]{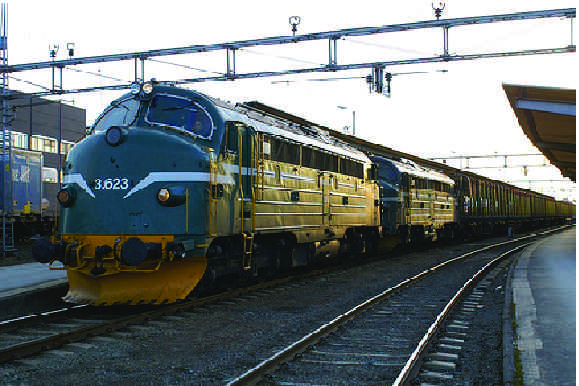} &
\includegraphics[height = 0.041\textheight, width = 0.125\textwidth, keepaspectratio = true]{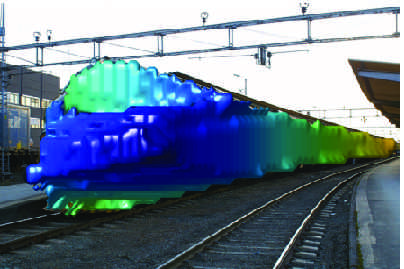} &
\includegraphics[height = 0.041\textheight, width = 0.125\textwidth, keepaspectratio = true]{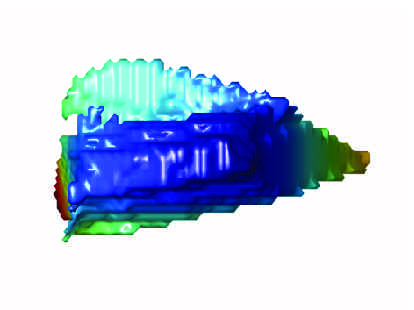} &
\includegraphics[height = 0.041\textheight, width = 0.125\textwidth, keepaspectratio = true]{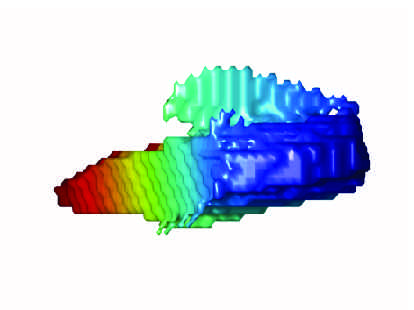}\\
\includegraphics[height = 0.041\textheight, width = 0.125\textwidth, keepaspectratio = true]{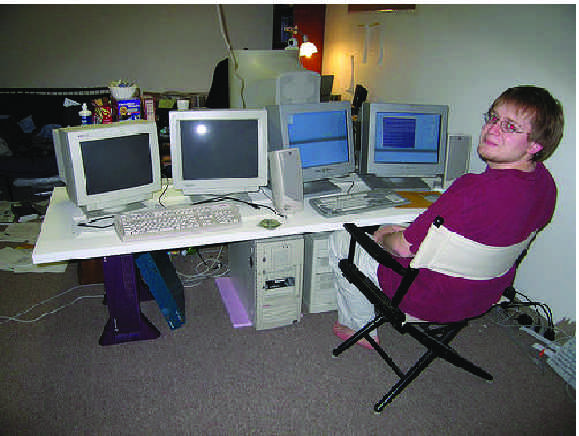} &
\includegraphics[height = 0.041\textheight, width = 0.125\textwidth, keepaspectratio = true]{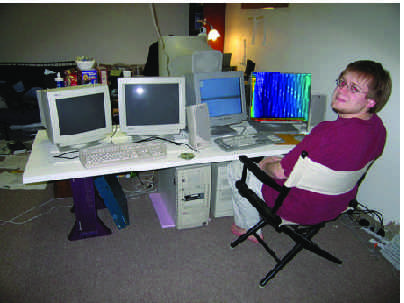} &
\includegraphics[height = 0.041\textheight, width = 0.125\textwidth, keepaspectratio = true]{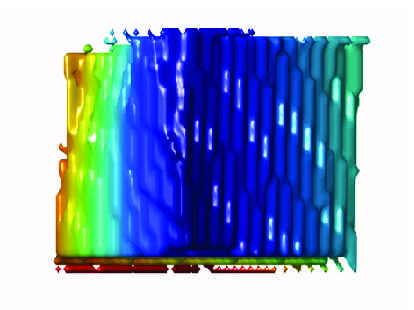} &
\includegraphics[height = 0.041\textheight, width = 0.125\textwidth, keepaspectratio = true]{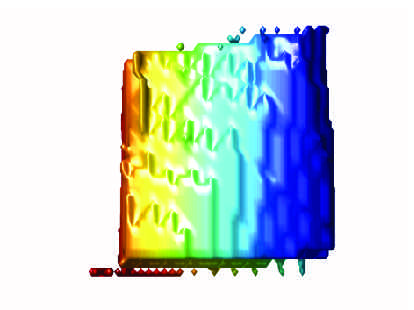}&
\includegraphics[height = 0.041\textheight, width = 0.125\textwidth, keepaspectratio = true]{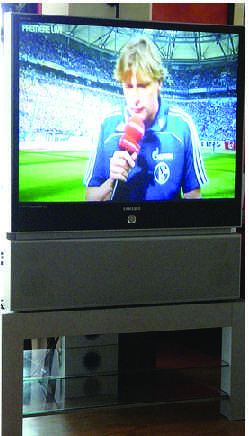} &
\includegraphics[height = 0.041\textheight, width = 0.125\textwidth, keepaspectratio = true]{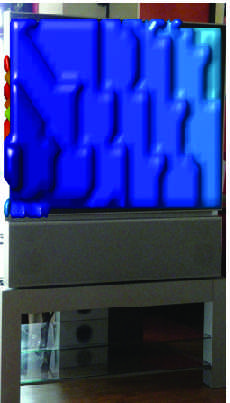} &
\includegraphics[height = 0.041\textheight, width = 0.125\textwidth, keepaspectratio = true]{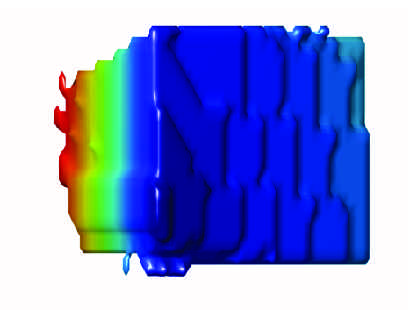} &
\includegraphics[height = 0.041\textheight, width = 0.125\textwidth, keepaspectratio = true]{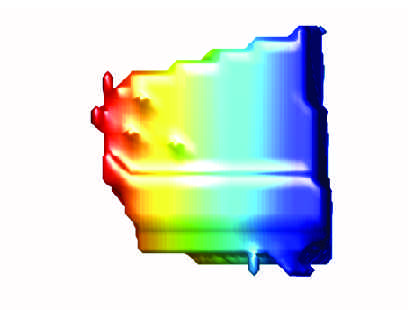}\\
\end{tabular}
\caption{\label{fig:reconstructions} Examples of our reconstructions for all 20 PASCAL VOC categories. For each object we show the original image, the original image with the reconstruction overlaid and two different viewpoint of our reconstruction. Blue is closer to the camera, red is farther (best seen in color). For most classes, our reconstructions convey the overall shape of the object, which is a remarkable achievement given the limited information used as input and the large amount of intra-class variation.
}
\end{figure*}

We reconstructed all the objects and show two example outputs from each class in fig.~\ref{fig:reconstructions}. We observe that surprisingly accurate reconstructions are obtained for most classes, with some apparent difficulties for ``dining table'', ``sofa'' and ``train''. The problems with ``dining table'' can be explained by there being only 13 exemplars marked as unoccluded, which makes camera viewpoint estimation frail. ``Sofa'' has a strong concavity which makes visual-hull reconstruction hard and would benefit from stereo-based post-processing, which we leave for future work. ``Train'' is a very difficult class to reconstruct in general: different trains may have a different number of carriages, there are strong perspective effects and it is articulated. Finally, sometimes our reconstructions of animals have either fewer or more limbs than in the image, and certain reconstructions have disconnected components. 

In all experiments, we sampled 20 reconstructions of each reference object instance and found our algorithm to be very efficient: it took just 7 hours to reconstruct VOC on a 12-core computer, with the camera refinement algorithm taking around 5 hours. 

\begin{figure*}[t]
  \includegraphics[width=0.33\linewidth]{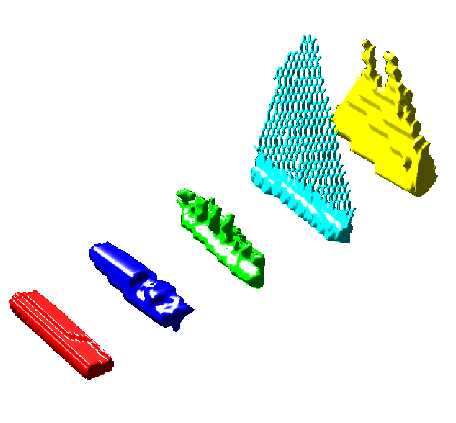}
  \includegraphics[width=0.33\linewidth]{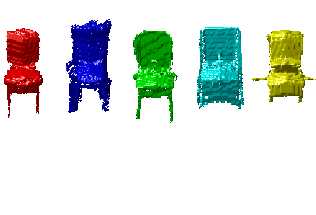}
  \includegraphics[width=0.33\linewidth]{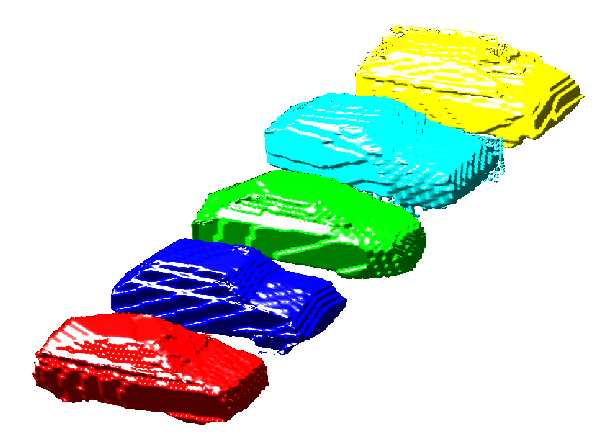}
  \caption{Clusters of boat, chair and car shapes, ordered from most frequent (red) to least frequent (yellow). The results agree with observation, namely SUV-like shapes and large sailboats (in yellow) are more rare than hatchbacks and something resembling a flat fishing boat (in red).\label{fig:shape_analysis}}
\end{figure*}

\vspace{2mm}
\noindent \textbf{Simple shape analysis.} Reconstruction of image collections, as pursued in this paper, holds the potential to greatly extend the domain of powerful shape analysis techniques developed in the graphics community \cite{Kim13}, that have however been mostly applied to collections of CAD models. Here, we made one small step in this direction and experimented with clustering our reconstructions on PASCAL VOC. For each class we first computed a distance matrix between all instances (using the symmetric mesh distance from \cite{aspert2002mesh}), then clustered each class into 5 clusters using the K-medoids algorithm. We show the resulting prototypes (medoids) for each cluster in fig.~\ref{fig:shape_analysis}, ordered by the number of elements assigned to that cluster, which reflects how frequently each type of shape appears in the dataset. 

\vspace{2mm}
\noindent \textbf{Viewpoint evaluation using Pascal3D+ ground truth cameras.} Recently, Xiang et. al. \cite{xiang2014beyond} augmented 12 of the object classes in the PASCAL dataset with human-provided 3D information -- the Pascal3D+ dataset. To construct this dataset, for each object, a human first selected the 3D CAD model most similar to it from a small set of options (a total of 70 CAD models for the 12 classes were used), then manually oriented and aligned it with the image. The viewpoint is then refined by optimizing the projection of ground truth keypoints in the 3D model to ground truth keypoints in the image. 

We use the Pascal3D+ dataset to evaluate our camera viewpoint estimation algorithm detailed in section \ref{sec:SFM}.  In Pascal3D+ the annotations for each object contain the camera's azimuth, elevation and camera roll angles which we compare to our estimates. We report results for 10 of the annotated classes (we exclude classes "bottle" and "dining table" because the keypoints we experimented with \cite{BroxSegmentationCVPR11} for both classes are view-dependent).

\begin{figure*}[t]
  \centering
  \includegraphics[width=0.31\linewidth]{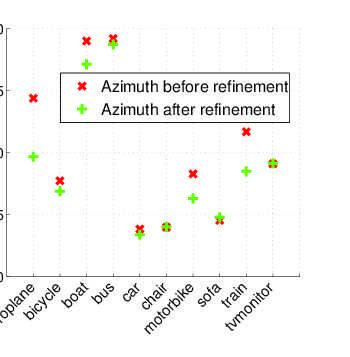}
  \includegraphics[width=0.31\linewidth]{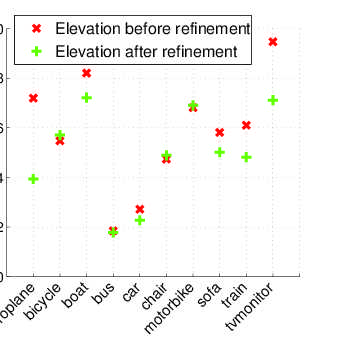}
  \includegraphics[width=0.31\linewidth]{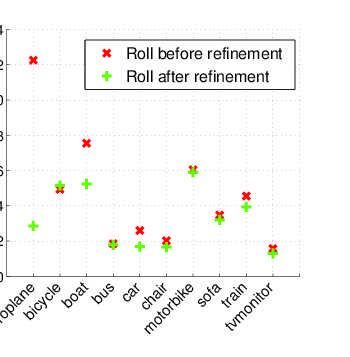}
   \caption{\label{fig:pascal3d_angle_error} Viewpoint estimation errors measured using the ground truth cameras from Pascal3D+. We measure the angle between our estimate and the ground truth and report the median angle in degrees for each class.  Elevation error is below 10 degrees for all classes, perhaps because most photos are taken from typical human viewpoints (PASCAL is assembled from FLICKR). Azimuth error is slightly higher for classes such as boat and bus, with possible causes being the extreme intra-class shape variation among boats (eg. kayaks, sailboats and cruise ships differ significantly) and perspective distortion, frequent in pictures of buses.}
\end{figure*}

\begin{figure}
\centering
\renewcommand{\arraystretch}{1}
\begin{tabular}{@{}c@{\hspace{1pt}}c@{\hspace{1pt}}c@{\hspace{1pt}}c@{}}
\includegraphics[height = 0.065\textheight, width = 0.2\textwidth, keepaspectratio = true]{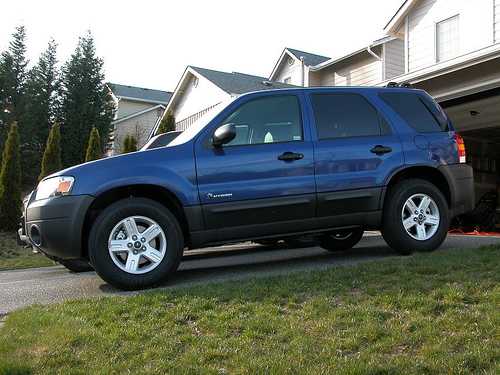} &
\includegraphics[height = 0.065\textheight, width = 0.2\textwidth, keepaspectratio = true]{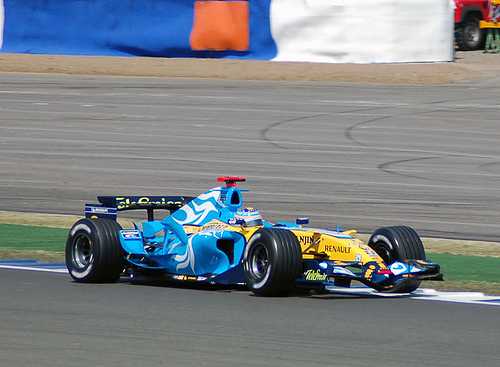} &
\includegraphics[height = 0.065\textheight, width = 0.2\textwidth, keepaspectratio = true]{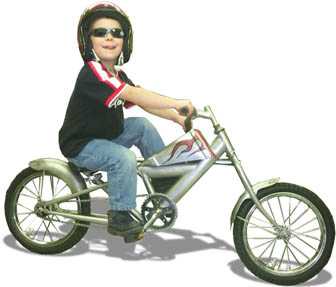} &
\includegraphics[height = 0.065\textheight, width = 0.2\textwidth, keepaspectratio = true]{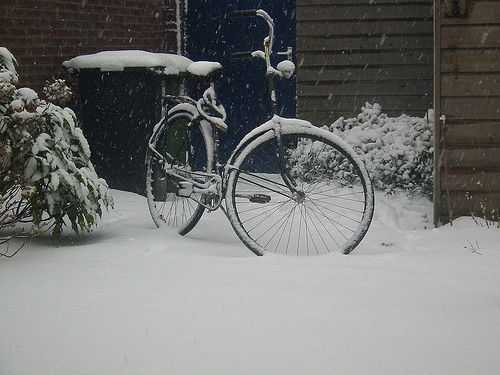} 
\\
\includegraphics[height = 0.065\textheight, width = 0.2\textwidth, keepaspectratio = true]{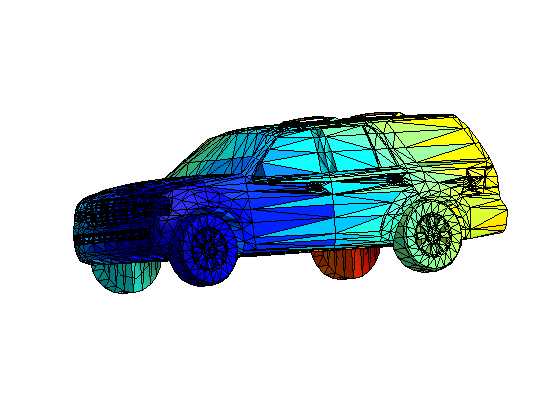} &
\includegraphics[height = 0.065\textheight, width = 0.2\textwidth, keepaspectratio = true]{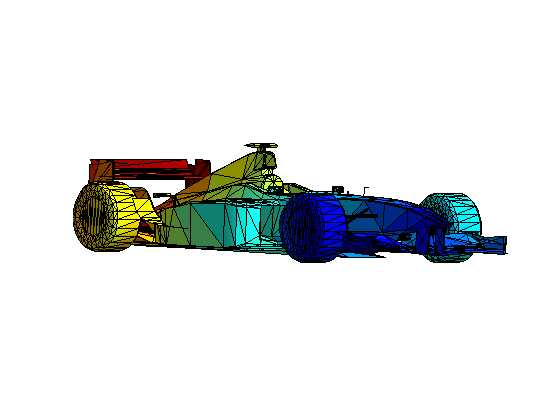} &
\includegraphics[height = 0.065\textheight, width = 0.2\textwidth, keepaspectratio = true]{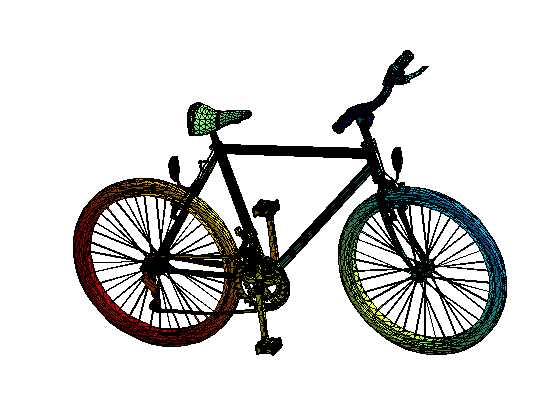} &
\includegraphics[height = 0.065\textheight, width = 0.2\textwidth, keepaspectratio = true]{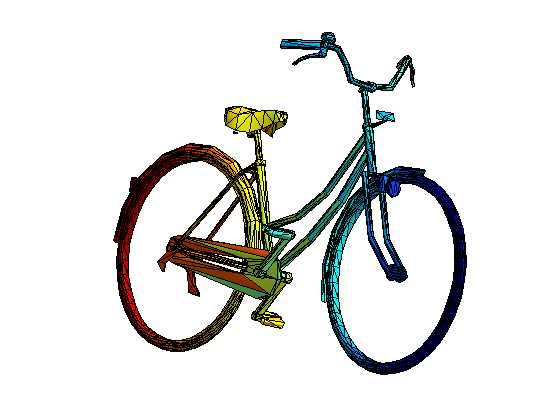} 
\\
\includegraphics[height = 0.065\textheight, width = 0.2\textwidth, keepaspectratio = true]{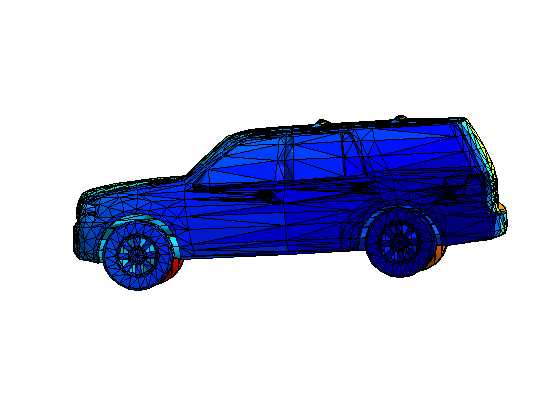} &
\includegraphics[height = 0.065\textheight, width = 0.2\textwidth, keepaspectratio = true]{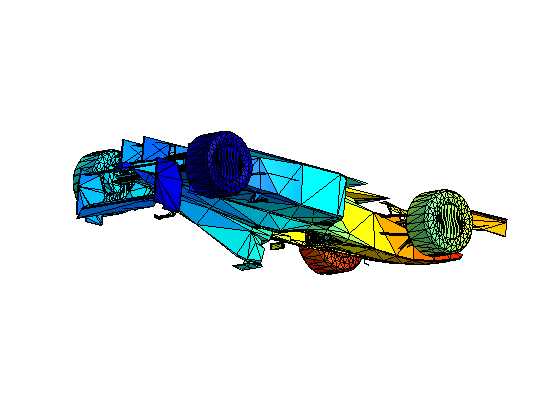} &
\includegraphics[height = 0.065\textheight, width = 0.2\textwidth, keepaspectratio = true]{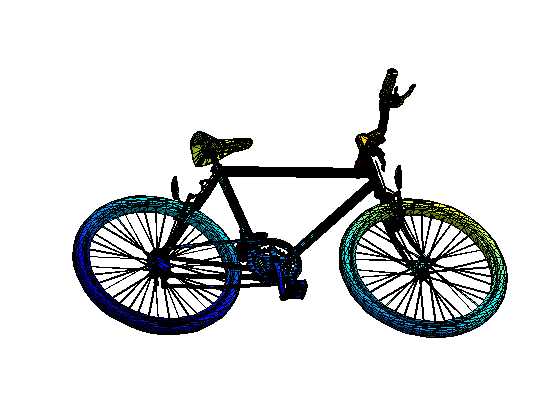} &
\includegraphics[height = 0.065\textheight, width = 0.2\textwidth, keepaspectratio = true]{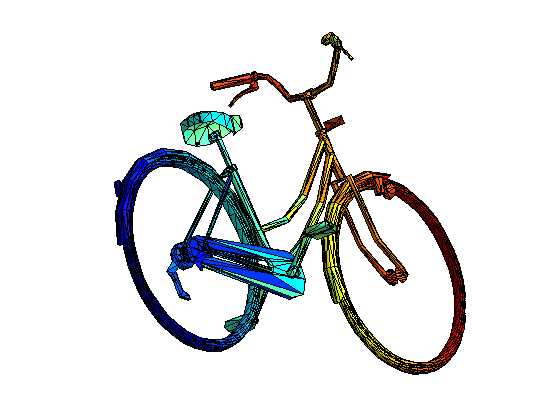} 
\end{tabular}
\caption{Viewpoint estimation failure cases. The first row is the input image, the second row the ground truth viewpoint in the Pascal3D+ dataset, displayed using the 3D CAD model associated with that image and the third row our viewpoint estimate using the same CAD model. Our viewpoint estimation algorithm is sometimes less accurate for images showing large perspective effects (first column), for instances very different from the class average (second and third columns) and in the presence of articulation (last column).  
\label{fig:sfm_failure_cases}}
\end{figure}

The results are shown in fig.~\ref{fig:pascal3d_angle_error} and are in most cases lower than $10^{\circ}$, which is the precision of the ground truth annotations of Pascal3D+ \cite{private_comm_roozbeh}. We measure the angle error in degrees and report the median for each class. The results show that our method is effective in estimating the viewpoint for most objects in all classes. It also shows that 
our refinement step detailed in section \ref{sec:SFM_refinement} consistently outperforms the initial estimate using rigid SFM. 
In fig.~\ref{fig:sfm_failure_cases} we show typical failure cases of viewpoint estimation: large perspective effects, large intra-class variation and articulations.

\subsection{Reconstructing a synthetic PASCAL VOC}

\begin{table}[t]
\begin{tabular}{|c | c c c | c  c |}
\hline
 		& Full	 	& -CRef		 	& -SImp 	& \cite{twarog2012playing} 	& SFMc  	\\
\hline		
aeroplane  	& \textbf{3.58}	& 4.94			& 3.95 		& 9.64		&  		5.79	 	\\
bicycle  	& 4.30		& \textbf{3.26} 	& 4.75		& 10.51		&  		6.56	 	\\
bird  		& 9.98		& 10.92			& 10.34		& \textbf{8.76}	& 		12.01	 	\\
boat  		& \textbf{5.91}	& 6.78			& 6.05 		& 8.81		&  		6.52	 	\\
bottle  	& 8.09		& 10.77			& 8.53		& \textbf{6.25}	&  		12.13	 	\\
bus  		& 6.45		& \textbf{6.10}		& 6.49		& 11.02		&  		7.34	 	\\
car  		& \textbf{3.04}	& 6.33 			& 3.10		& 11.07		&	  	3.22 		\\
cat  		& \textbf{6.98}	& 7.57 			& 7.49		& 11.39		&	 	9.61 		\\
chair  		& \textbf{5.36}	& 5.73			& 6.06		& 8.13		&	  	7.37		\\
cow  		& 5.44		& \textbf{5.24}		& 5.83		& 9.17		&	 	7.50 		\\
diningtable  	& 8.97  	& 12.57			& 14.30		& \textbf{8.67}	&	 	9.52		\\
dog  		& \textbf{7.08}	& 8.38			& 7.19		& 11.61		&	 	9.91  		\\
horse  		& \textbf{6.05}	& 7.05			& 6.38		& 6.90		&	 	7.41 		\\
motorbike  	& \textbf{4.12}	& 4.24			& 4.16  	& 9.24		&		5.32  		\\
person  	& \textbf{7.35}	& 7.95 			& 7.55		& 9.14		&	  	19.46		\\
pottedplant  	& 7.72		& 8.15			& 7.99		& \textbf{7.58}	&	  	17.86		\\
sheep  		& 7.18 		& \textbf{7.15}		& 7.66		& 8.77		&	  	7.16		\\
sofa  		& 6.11		& 6.24			& 6.31		& 8.06		&		\textbf{5.75}		\\
train  		& \textbf{15.73}& 20.55			& 16.19		& 17.01		&	  	17.47		\\
tv/monitor 	& 9.73		& 10.45			& 10.28		& \textbf{9.67}	&		10.08  		\\
\hline
\textbf{Mean}  	& \textbf{6.96}	& 8.01			& 7.53		& 9.57 		&	 	9.40		\\
\hline
\end{tabular}

\vskip +0.1in
\caption{\label{tab:synth_results} Symmetric root mean square error between reconstructed and ground truth 3D models. Lowest errors are displayed in bold. We compare our full model (Full), with severed versions without our proposed camera refinement process (-CRef) and reference silhouette imprinting (-SImp). As baselines we consider a recent single view silhouette-based reconstruction method \cite{twarog2012playing} and the convex hull of the 3D points returned by our rigid structure-from-motion component (SFMc).}
\end{table}

We also performed a quantitative evaluation on synthetic test images with similar segmentations and keypoints as those in VOC. To make results as representative of performance on real data as possible, we reconstruct using only surrogate shapes from VOC. We downloaded 10 meshes for each category from the web, then manually annotated keypoints consistent with those of \cite{BroxSegmentationCVPR11} in 3D and rendered them using 5 different cameras, sampled from the ones estimated on VOC for that class. This resulted in 50 synthetic images per class, each with associated segmentation and visible keypoints, for a total of 1000 test examples. 

We measure the distortion between a reconstruction and a ground truth 3D mesh using, as in the clustering experiment, the symmetric root mean squared error between the two meshes \cite{aspert2002mesh}. Let the root mean squared error be:
\begin{equation}
e_{RMS}(S,S') = \sqrt{\frac{1}{|S|}\iint_{p \in S}{d(p,S')^2 dS}},
\end{equation}
where $S$ and $S'$ are the two meshes we want to compare and $d(p,S')$ is the distance of a point to a mesh, defined by the Hausdorff  distance, i.e.~the minimum euclidean distance between point $p$ and any point on the mesh $S'$. Since this distance is not symmetric we use instead:
\begin{equation}
e(S,S') = max [e_{RMS}(S,S'), e_{RMS}(S',S)]
\end{equation}

\begin{figure*}
\centering
\renewcommand{\arraystretch}{1}
\begin{tabular}{@{}c@{} c@{} c@{} c@{} c@{} | c@{} c@{} | c@{} c@{}}

\multicolumn{2}{c}{GT mesh} & \multicolumn{1}{c}{Input} & \multicolumn{2}{c|}{Our result}
&  \multicolumn{2}{c|}{\cite{twarog2012playing}} & \multicolumn{2}{c}{SFMc} \\
 {\small Image view} & {\small Top view} & ~ & {\small Image view} & {\small Top view} & {\small Image view} & {\small Top view} & {\small Image view} & {\small Top view} \\

\includegraphics[height = 0.055\textheight, width = 0.195\textwidth, keepaspectratio = true]{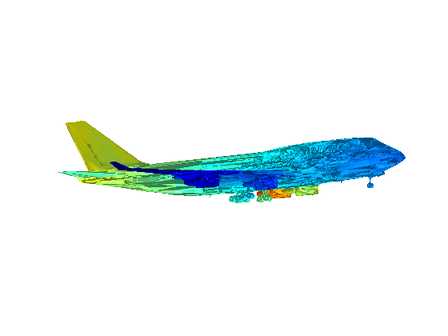} &
\includegraphics[height = 0.055\textheight, width = 0.195\textwidth, keepaspectratio = true]{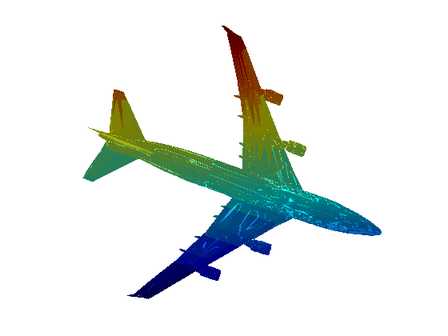} &
\includegraphics[height = 0.055\textheight, width = 0.195\textwidth, keepaspectratio = true]{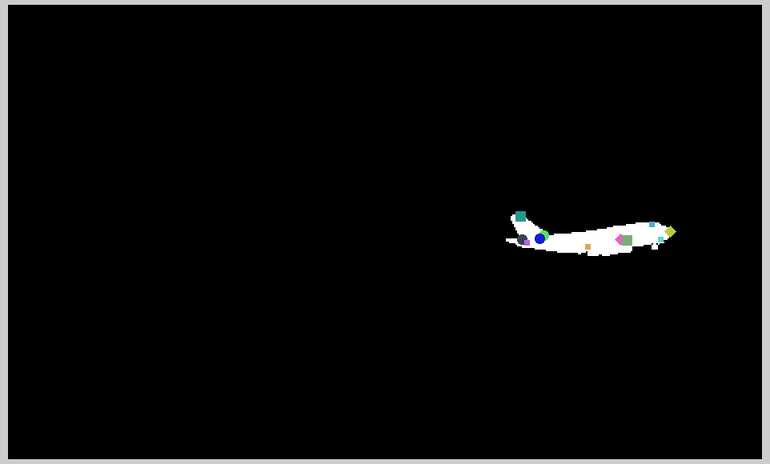} &
\includegraphics[height = 0.055\textheight, width = 0.195\textwidth, keepaspectratio = true]{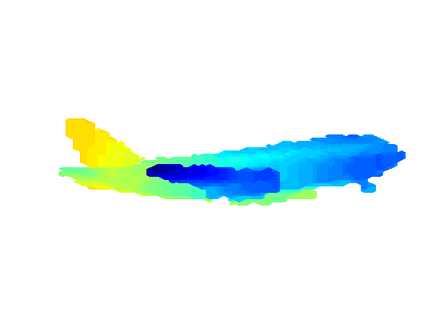} &
\includegraphics[height = 0.055\textheight, width = 0.195\textwidth, keepaspectratio = true]{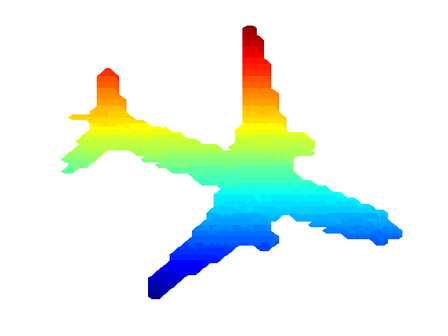} &
\includegraphics[height = 0.055\textheight, width = 0.195\textwidth, keepaspectratio = true]{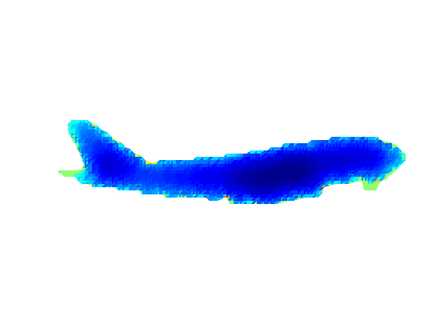} &
\includegraphics[height = 0.055\textheight, width = 0.195\textwidth, keepaspectratio = true]{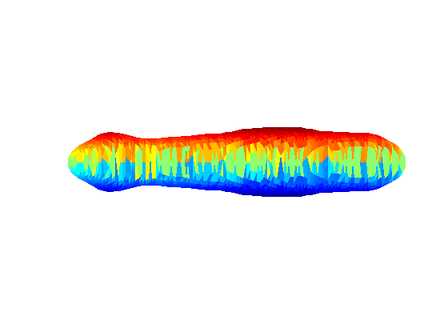} &
\includegraphics[height = 0.055\textheight, width = 0.195\textwidth, keepaspectratio = true]{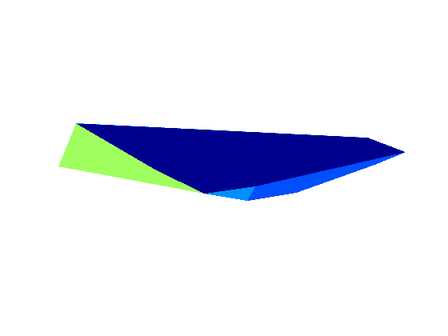} &
\includegraphics[height = 0.055\textheight, width = 0.195\textwidth, keepaspectratio = true]{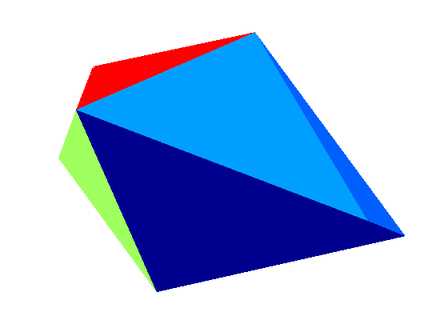}\\

\includegraphics[height = 0.055\textheight, width = 0.195\textwidth, keepaspectratio = true]{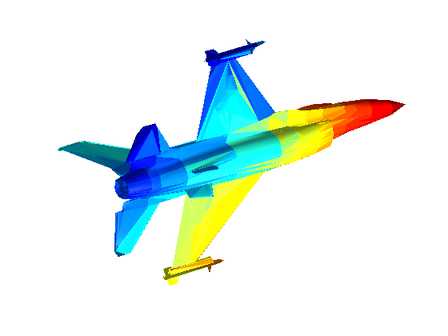} &
\includegraphics[height = 0.055\textheight, width = 0.195\textwidth, keepaspectratio = true]{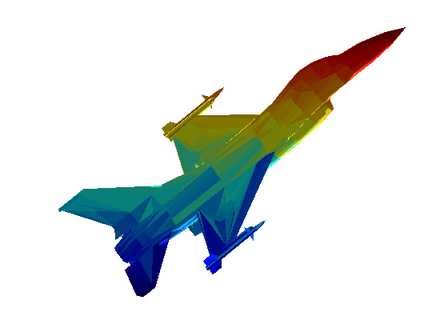} &
\includegraphics[height = 0.055\textheight, width = 0.195\textwidth, keepaspectratio = true]{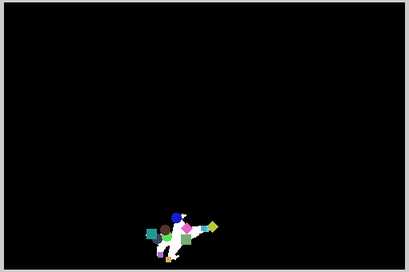} &
\includegraphics[height = 0.055\textheight, width = 0.195\textwidth, keepaspectratio = true]{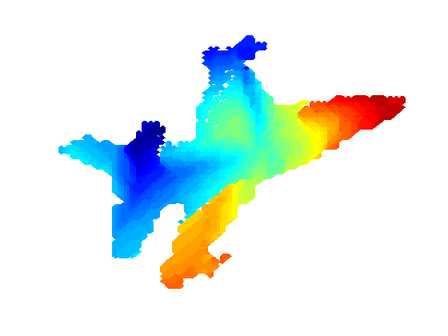} &
\includegraphics[height = 0.055\textheight, width = 0.195\textwidth, keepaspectratio = true]{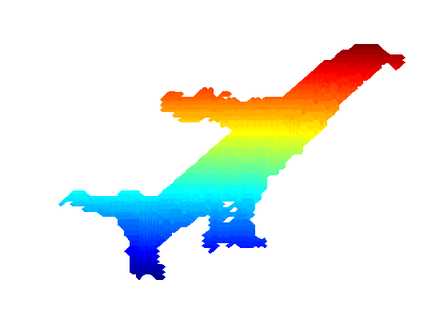} &
\includegraphics[height = 0.055\textheight, width = 0.195\textwidth, keepaspectratio = true]{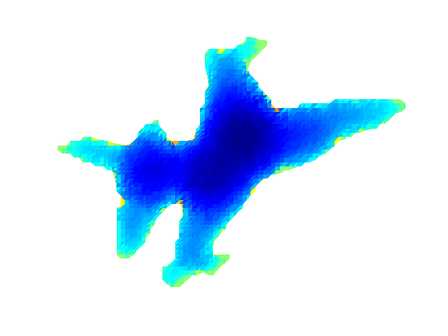} &
\includegraphics[height = 0.055\textheight, width = 0.195\textwidth, keepaspectratio = true]{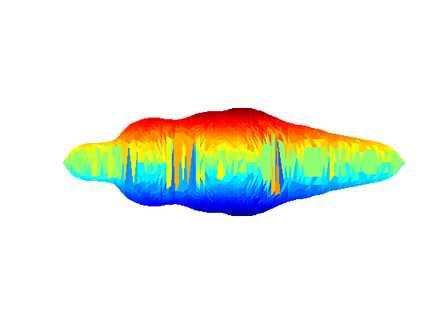} &
\includegraphics[height = 0.055\textheight, width = 0.195\textwidth, keepaspectratio = true]{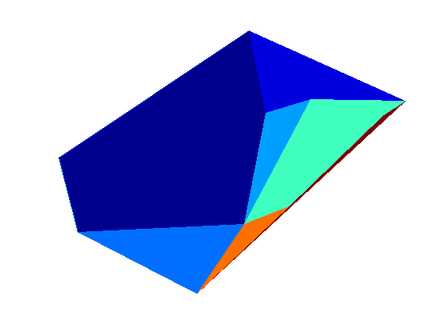} &
\includegraphics[height = 0.055\textheight, width = 0.195\textwidth, keepaspectratio = true]{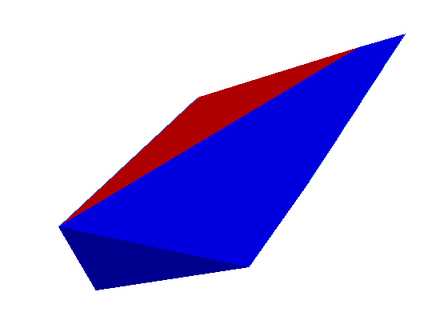}\\

\includegraphics[height = 0.055\textheight, width = 0.195\textwidth, keepaspectratio = true]{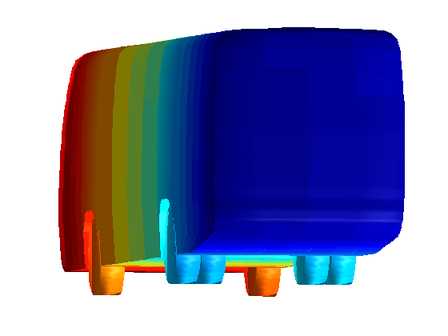} &
\includegraphics[height = 0.055\textheight, width = 0.195\textwidth, keepaspectratio = true]{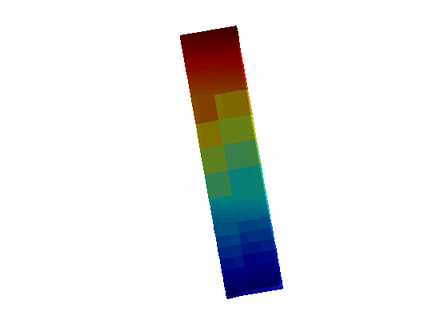} &
\includegraphics[height = 0.055\textheight, width = 0.195\textwidth, keepaspectratio = true]{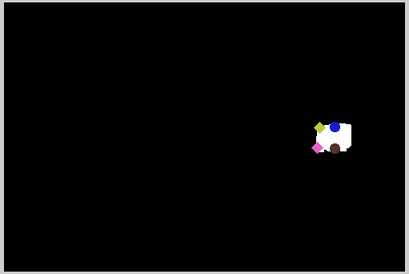} &
\includegraphics[height = 0.055\textheight, width = 0.195\textwidth, keepaspectratio = true]{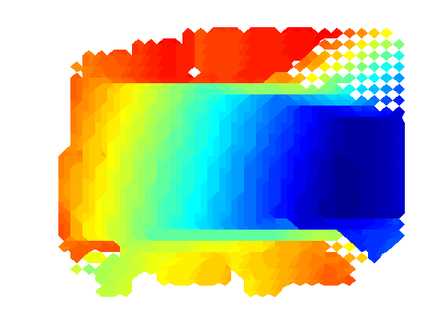} &
\includegraphics[height = 0.055\textheight, width = 0.195\textwidth, keepaspectratio = true]{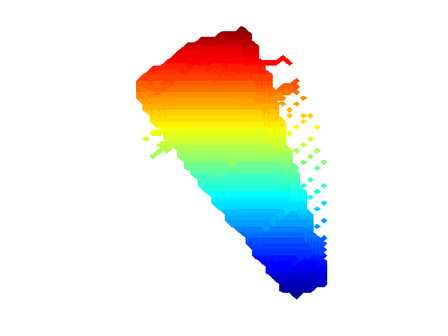} &
\includegraphics[height = 0.055\textheight, width = 0.195\textwidth, keepaspectratio = true]{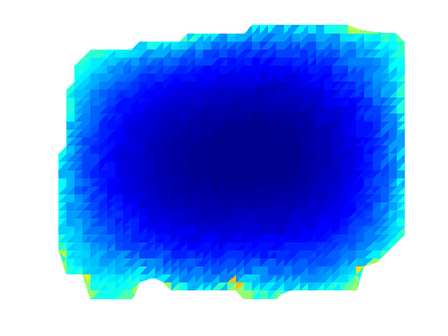} &
\includegraphics[height = 0.055\textheight, width = 0.195\textwidth, keepaspectratio = true]{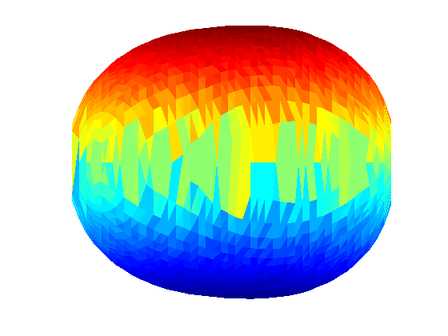} &
\includegraphics[height = 0.055\textheight, width = 0.195\textwidth, keepaspectratio = true]{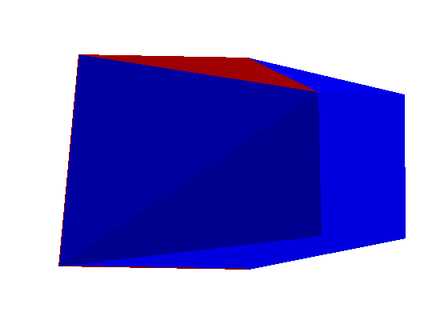} &
\includegraphics[height = 0.055\textheight, width = 0.195\textwidth, keepaspectratio = true]{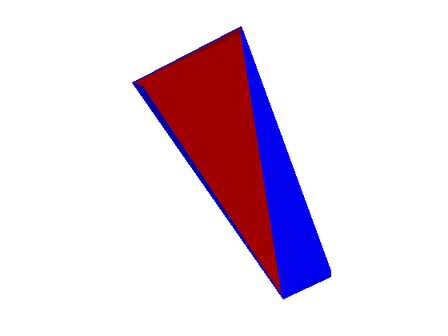}\\

\includegraphics[height = 0.055\textheight, width = 0.195\textwidth, keepaspectratio = true]{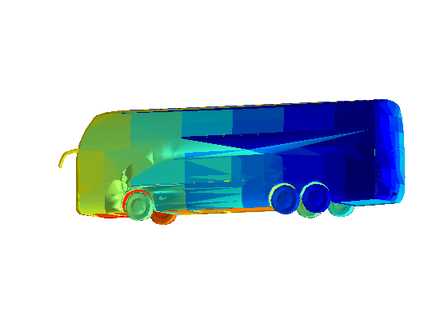} &
\includegraphics[height = 0.055\textheight, width = 0.195\textwidth, keepaspectratio = true]{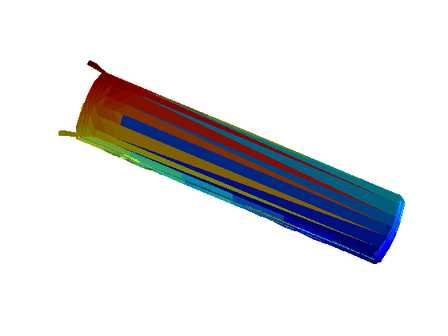} &
\includegraphics[height = 0.055\textheight, width = 0.195\textwidth, keepaspectratio = true]{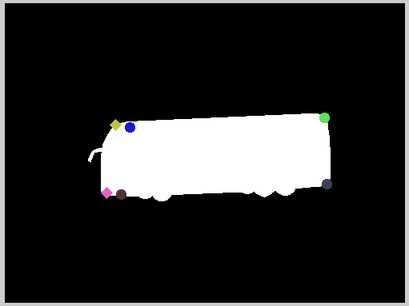} &
\includegraphics[height = 0.055\textheight, width = 0.195\textwidth, keepaspectratio = true]{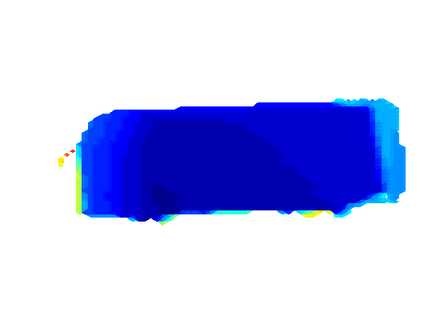} &
\includegraphics[height = 0.055\textheight, width = 0.195\textwidth, keepaspectratio = true]{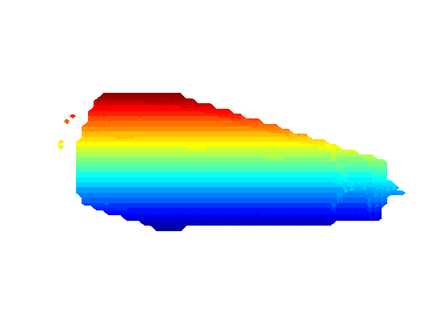} &
\includegraphics[height = 0.055\textheight, width = 0.195\textwidth, keepaspectratio = true]{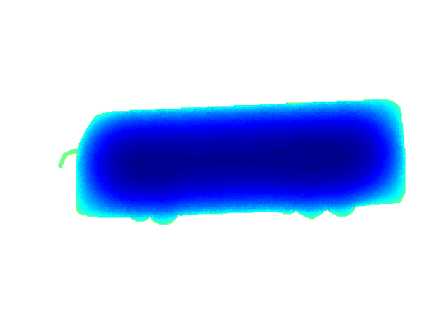} &
\includegraphics[height = 0.055\textheight, width = 0.195\textwidth, keepaspectratio = true]{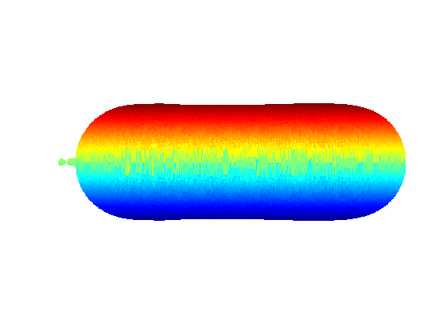} &
\includegraphics[height = 0.055\textheight, width = 0.195\textwidth, keepaspectratio = true]{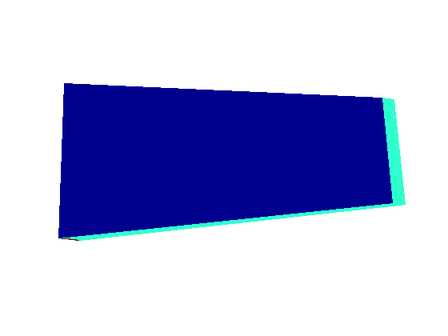} &
\includegraphics[height = 0.055\textheight, width = 0.195\textwidth, keepaspectratio = true]{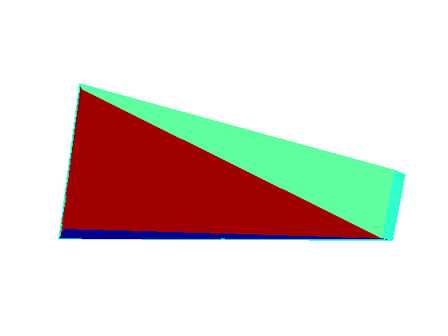}\\

\includegraphics[height = 0.055\textheight, width = 0.195\textwidth, keepaspectratio = true]{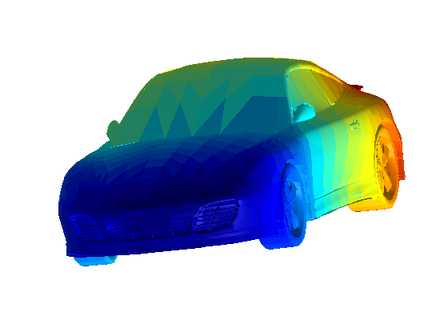} &
\includegraphics[height = 0.055\textheight, width = 0.195\textwidth, keepaspectratio = true]{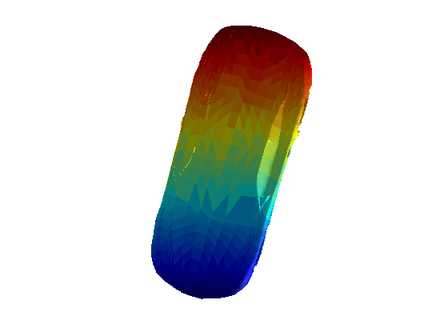} &
\includegraphics[height = 0.055\textheight, width = 0.195\textwidth, keepaspectratio = true]{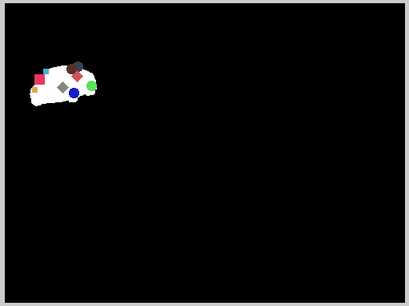} &
\includegraphics[height = 0.055\textheight, width = 0.195\textwidth, keepaspectratio = true]{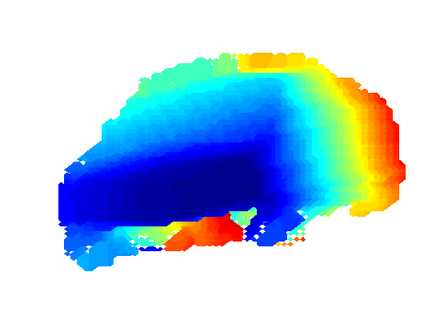} &
\includegraphics[height = 0.055\textheight, width = 0.195\textwidth, keepaspectratio = true]{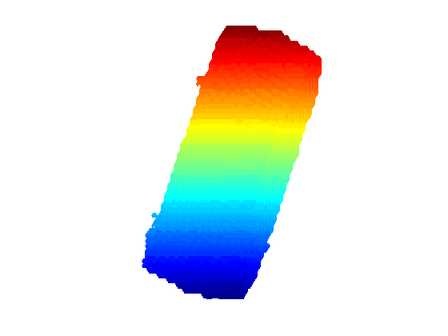} &
\includegraphics[height = 0.055\textheight, width = 0.195\textwidth, keepaspectratio = true]{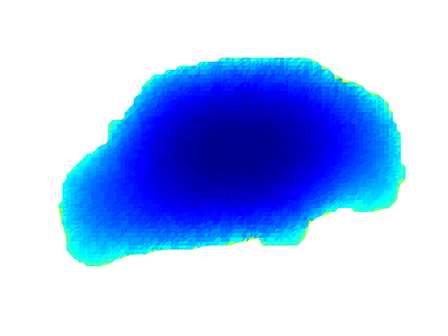} &
\includegraphics[height = 0.055\textheight, width = 0.195\textwidth, keepaspectratio = true]{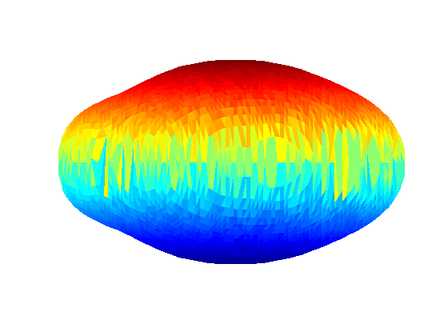} &
\includegraphics[height = 0.055\textheight, width = 0.195\textwidth, keepaspectratio = true]{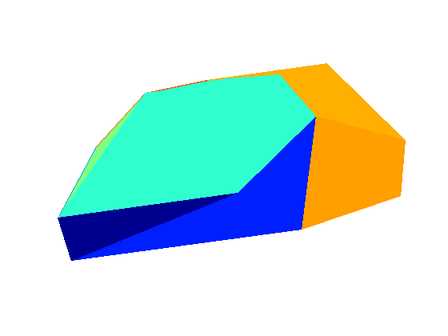} &
\includegraphics[height = 0.055\textheight, width = 0.195\textwidth, keepaspectratio = true]{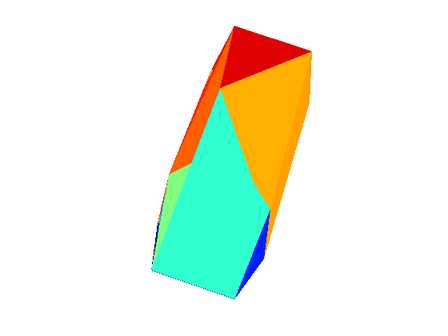}\\

\includegraphics[height = 0.055\textheight, width = 0.195\textwidth, keepaspectratio = true]{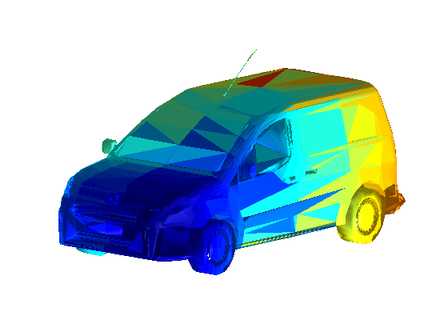} &
\includegraphics[height = 0.055\textheight, width = 0.195\textwidth, keepaspectratio = true]{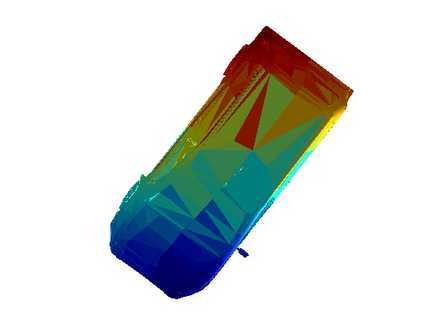} &
\includegraphics[height = 0.055\textheight, width = 0.195\textwidth, keepaspectratio = true]{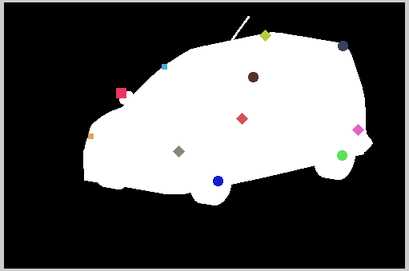} &
\includegraphics[height = 0.055\textheight, width = 0.195\textwidth, keepaspectratio = true]{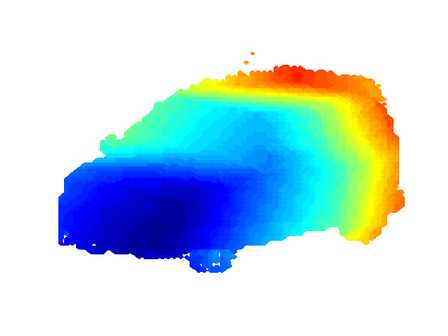} &
\includegraphics[height = 0.055\textheight, width = 0.195\textwidth, keepaspectratio = true]{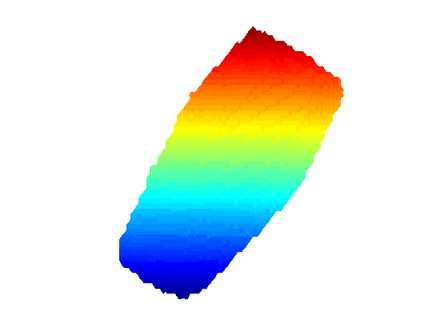} &
\includegraphics[height = 0.055\textheight, width = 0.195\textwidth, keepaspectratio = true]{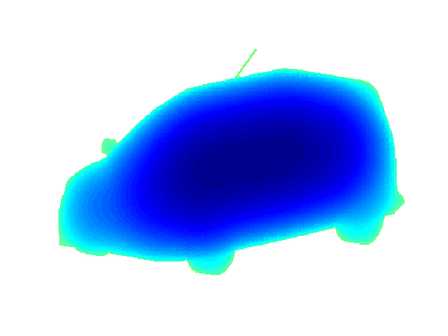} &
\includegraphics[height = 0.055\textheight, width = 0.195\textwidth, keepaspectratio = true]{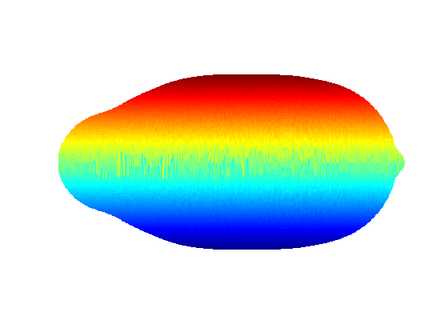} &
\includegraphics[height = 0.055\textheight, width = 0.195\textwidth, keepaspectratio = true]{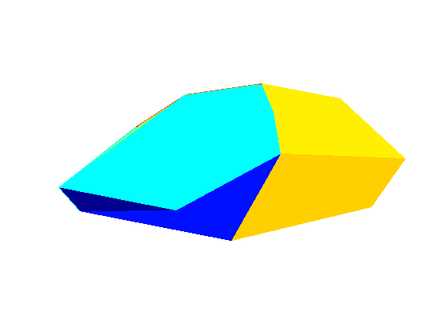} &
\includegraphics[height = 0.055\textheight, width = 0.195\textwidth, keepaspectratio = true]{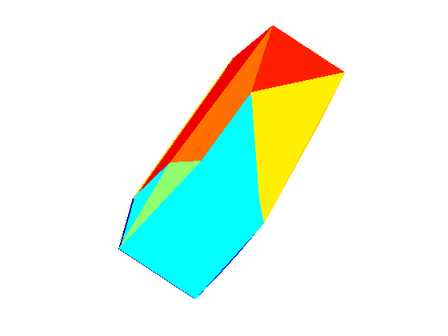}\\

\includegraphics[height = 0.055\textheight, width = 0.195\textwidth, keepaspectratio = true]{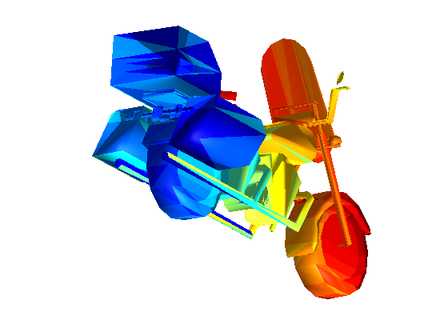} &
\includegraphics[height = 0.055\textheight, width = 0.195\textwidth, keepaspectratio = true]{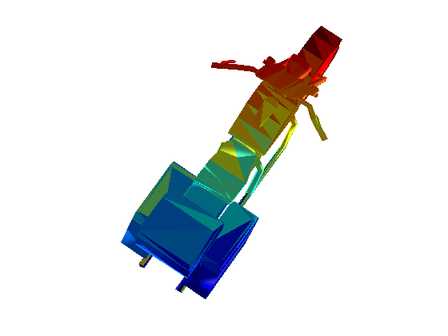} &
\includegraphics[height = 0.055\textheight, width = 0.195\textwidth, keepaspectratio = true]{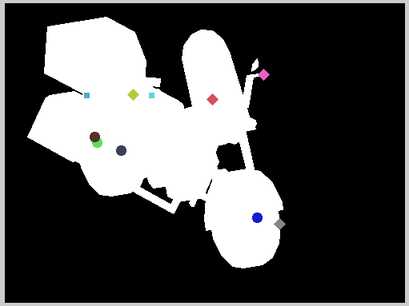} &
\includegraphics[height = 0.055\textheight, width = 0.195\textwidth, keepaspectratio = true]{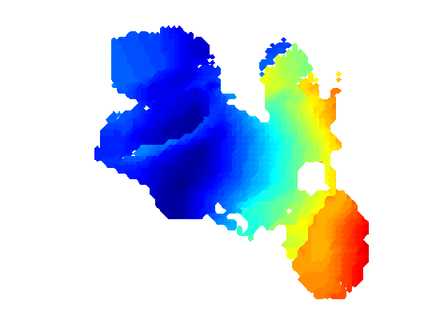} &
\includegraphics[height = 0.055\textheight, width = 0.195\textwidth, keepaspectratio = true]{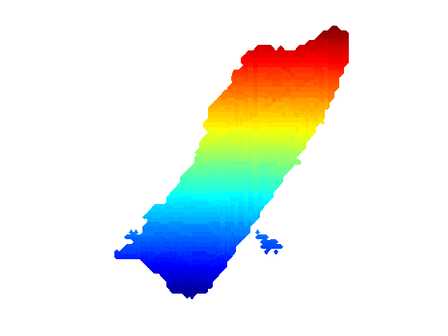} &
\includegraphics[height = 0.055\textheight, width = 0.195\textwidth, keepaspectratio = true]{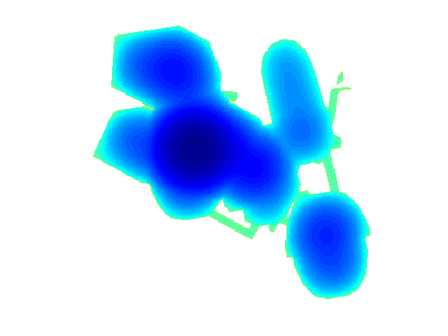} &
\includegraphics[height = 0.055\textheight, width = 0.195\textwidth, keepaspectratio = true]{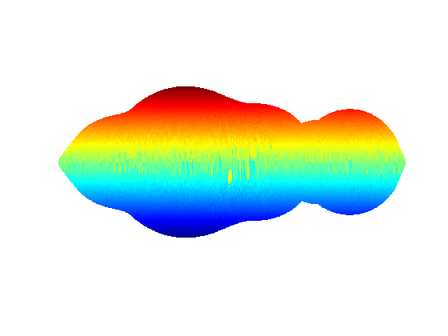} &
\includegraphics[height = 0.055\textheight, width = 0.195\textwidth, keepaspectratio = true]{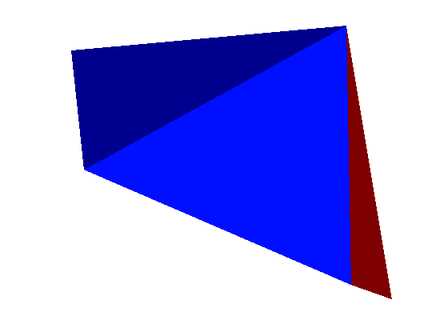} &
\includegraphics[height = 0.055\textheight, width = 0.195\textwidth, keepaspectratio = true]{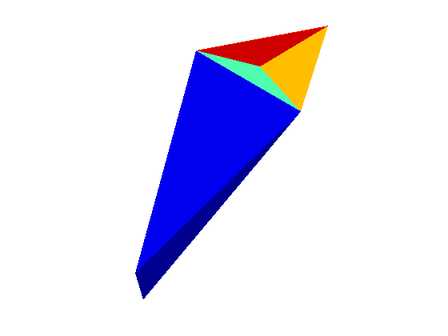}\\

\includegraphics[height = 0.055\textheight, width = 0.195\textwidth, keepaspectratio = true]{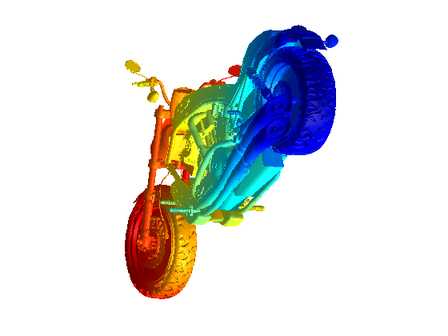} &
\includegraphics[height = 0.055\textheight, width = 0.195\textwidth, keepaspectratio = true]{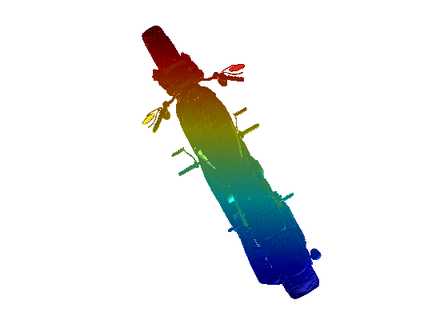} &
\includegraphics[height = 0.055\textheight, width = 0.195\textwidth, keepaspectratio = true]{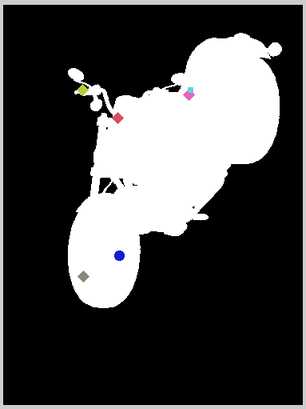} &
\includegraphics[height = 0.055\textheight, width = 0.195\textwidth, keepaspectratio = true]{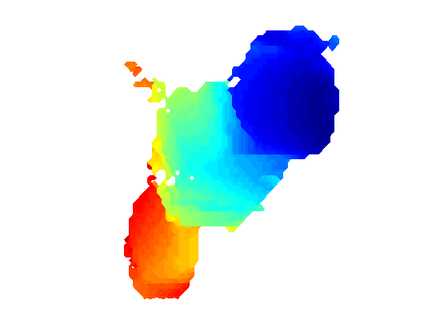} &
\includegraphics[height = 0.055\textheight, width = 0.195\textwidth, keepaspectratio = true]{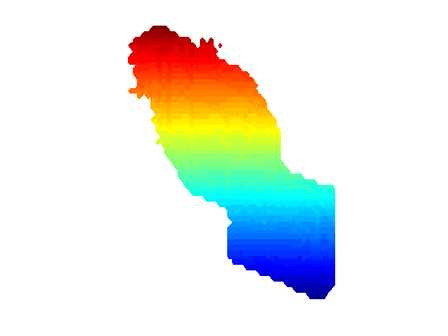} &
\includegraphics[height = 0.055\textheight, width = 0.195\textwidth, keepaspectratio = true]{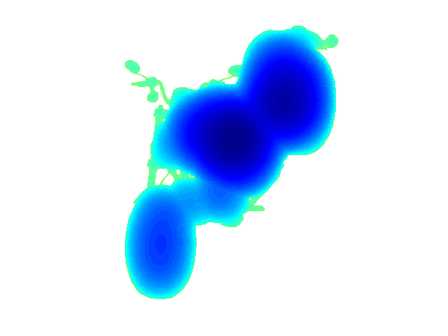} &
\includegraphics[height = 0.055\textheight, width = 0.195\textwidth, keepaspectratio = true]{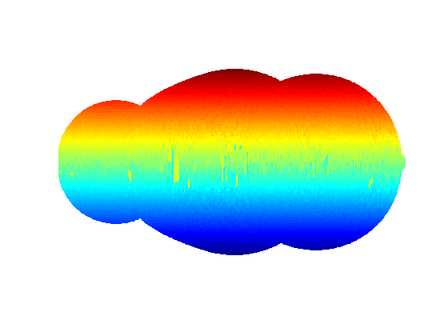} &
\includegraphics[height = 0.055\textheight, width = 0.195\textwidth, keepaspectratio = true]{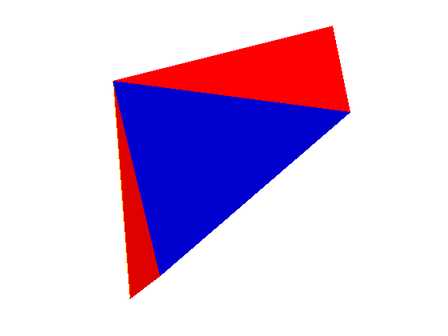} &
\includegraphics[height = 0.055\textheight, width = 0.195\textwidth, keepaspectratio = true]{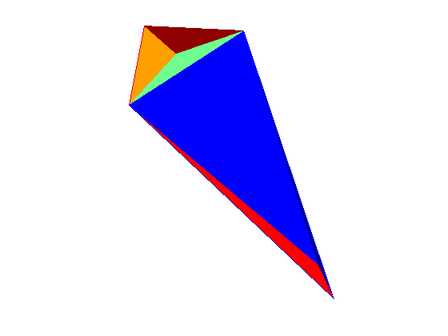}\\

\includegraphics[height = 0.055\textheight, width = 0.195\textwidth, keepaspectratio = true]{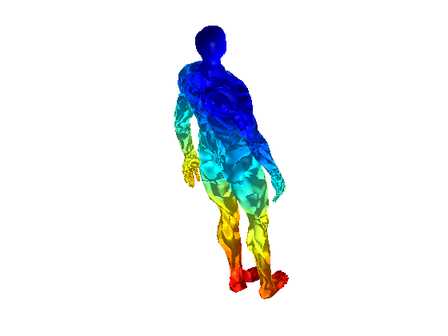} &
\includegraphics[height = 0.055\textheight, width = 0.195\textwidth, keepaspectratio = true]{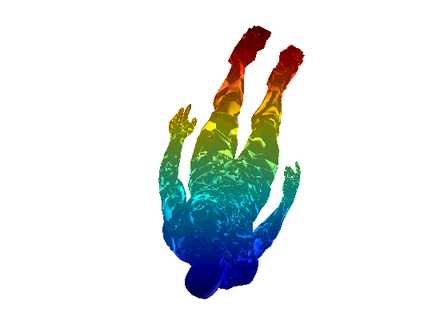} &
\includegraphics[height = 0.055\textheight, width = 0.195\textwidth, keepaspectratio = true]{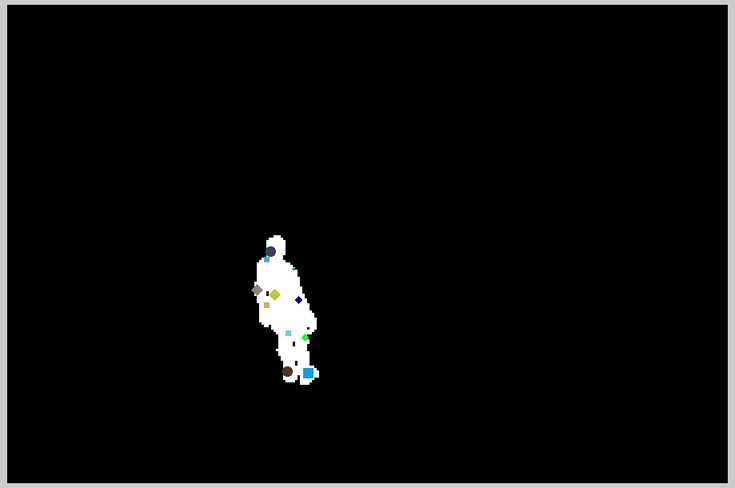} &
\includegraphics[height = 0.055\textheight, width = 0.195\textwidth, keepaspectratio = true]{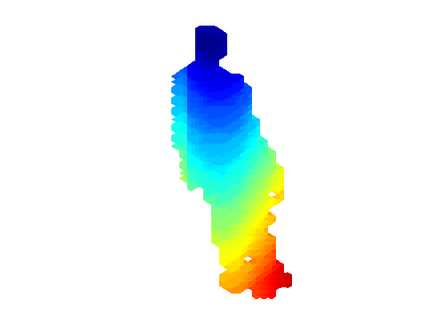} &
\includegraphics[height = 0.055\textheight, width = 0.195\textwidth, keepaspectratio = true]{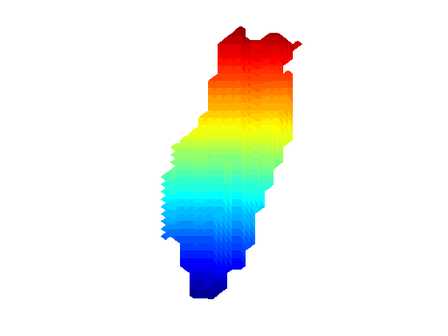} &
\includegraphics[height = 0.055\textheight, width = 0.195\textwidth, keepaspectratio = true]{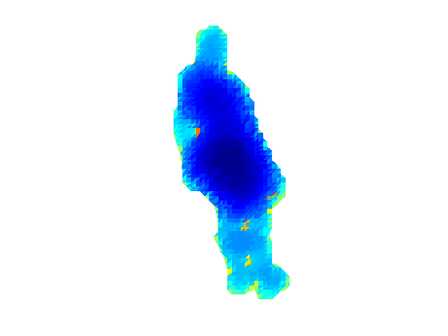} &
\includegraphics[height = 0.055\textheight, width = 0.195\textwidth, keepaspectratio = true]{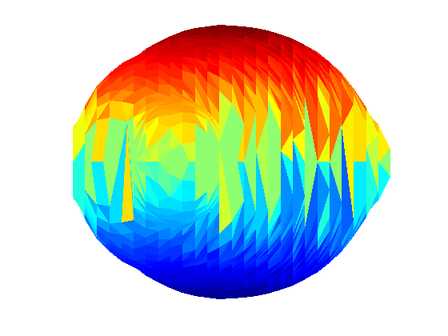} &
\includegraphics[height = 0.055\textheight, width = 0.195\textwidth, keepaspectratio = true]{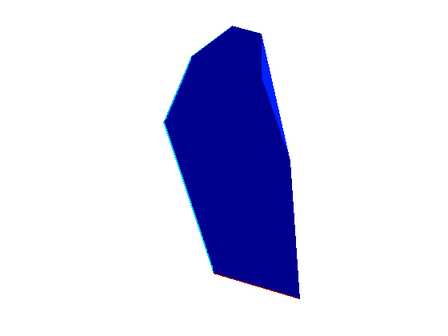} &
\includegraphics[height = 0.055\textheight, width = 0.195\textwidth, keepaspectratio = true]{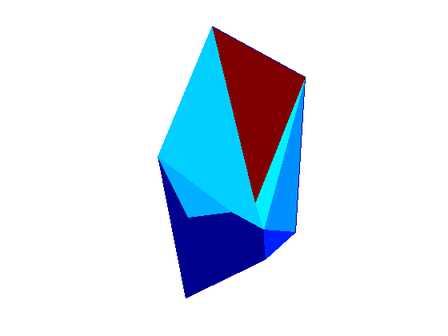}\\ 

\includegraphics[height = 0.055\textheight, width = 0.195\textwidth, keepaspectratio = true]{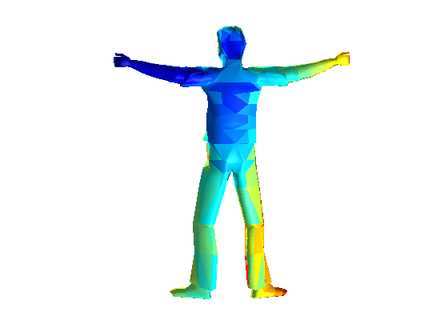} &
\includegraphics[height = 0.055\textheight, width = 0.195\textwidth, keepaspectratio = true]{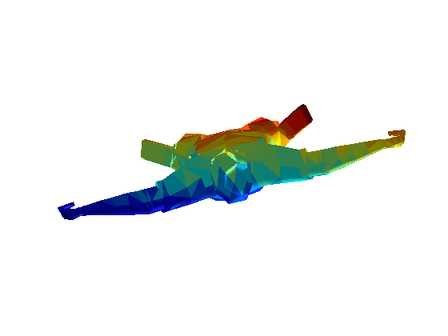} &
\includegraphics[height = 0.055\textheight, width = 0.195\textwidth, keepaspectratio = true]{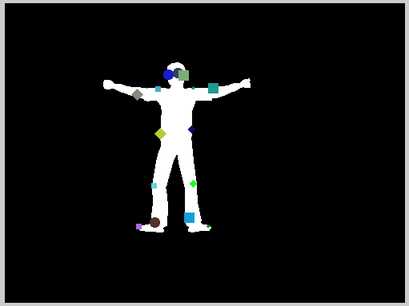} &
\includegraphics[height = 0.055\textheight, width = 0.195\textwidth, keepaspectratio = true]{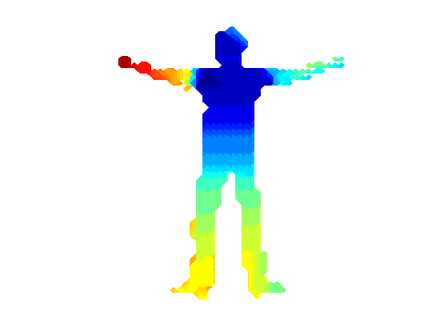} &
\includegraphics[height = 0.055\textheight, width = 0.195\textwidth, keepaspectratio = true]{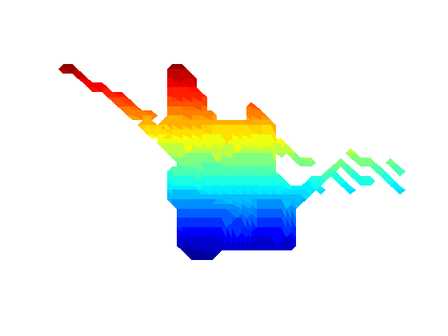} &
\includegraphics[height = 0.055\textheight, width = 0.195\textwidth, keepaspectratio = true]{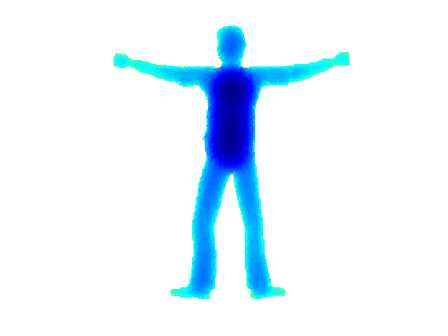} &
\includegraphics[height = 0.055\textheight, width = 0.195\textwidth, keepaspectratio = true]{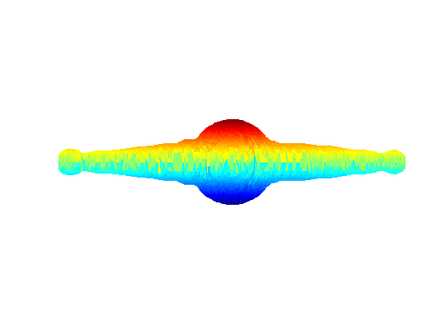} &
\includegraphics[height = 0.055\textheight, width = 0.195\textwidth, keepaspectratio = true]{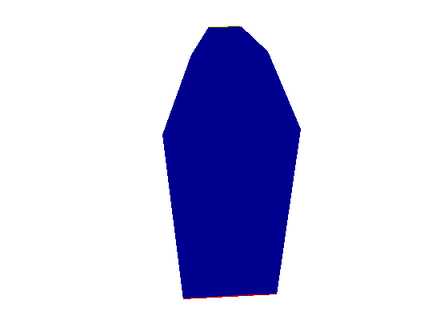} &
\includegraphics[height = 0.055\textheight, width = 0.195\textwidth, keepaspectratio = true]{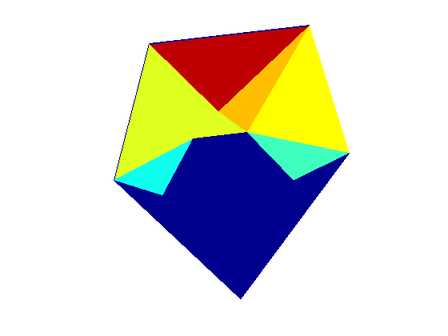}\\ 
\end{tabular}
\caption{\label{fig:synth_recons} Original and reconstructed shapes (synthetic dataset). Blue means closer to the camera, red means farther (best seen in color). The first two columns show the camera and top views of a CAD model from the synthetic dataset. The rendered mask and keypoints are shown in the third column and these are the sole inputs to our algorithm. The remaining columns show results of our method and two baselines, Puffball \cite{twarog2012playing} and a convex hull approach which generates a mesh directly from the output of our rigid structure-from-motion component.}
\end{figure*}

We normalize scale using the diagonal length of the bounding box of the ground truth 3D model, such that the error is a percentage of this length,
and report the average error over all the objects in each category. 
Table \ref{tab:synth_results} demonstrates the benefits of the different components of our proposed methodology. Since no other existing class reconstruction technique scales to such a large and diverse dataset using simple 2D annotations we compare to two simple baselines: an inflation technique originally proposed for silhouette based single-view reconstruction called \textit{Puffball} \cite{twarog2012playing} and a multiview baseline relying on our rigid SfM. Our method is significantly better for most classes, and a visual comparison of resulting reconstructions obtained is available in fig.~\ref{fig:synth_recons}, together with some of the CAD models in the dataset and their renderings.

\begin{figure*}[t]
    \centering
\includegraphics[width=0.32\linewidth]{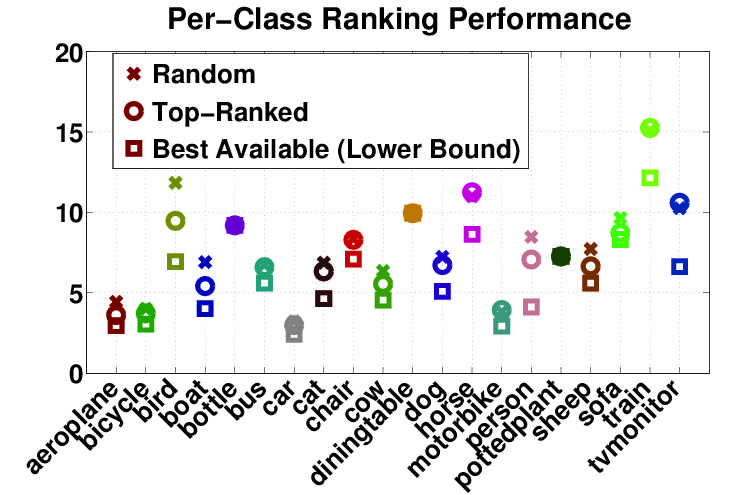}
\includegraphics[width=0.33\linewidth]{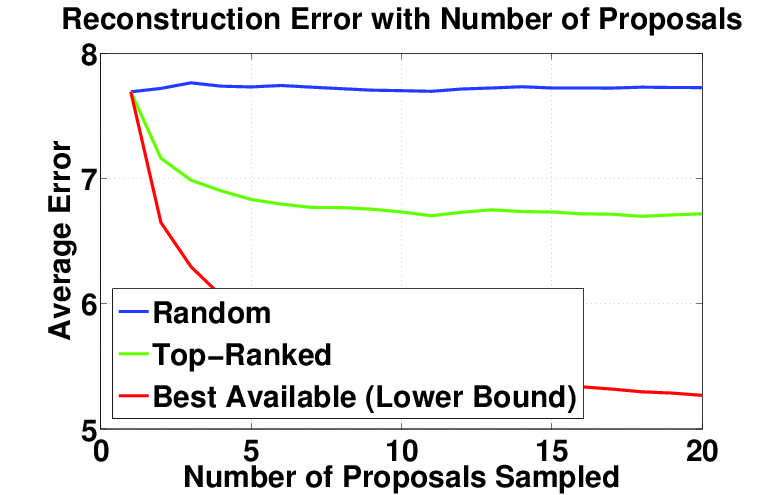}
\includegraphics[width=0.33\linewidth]{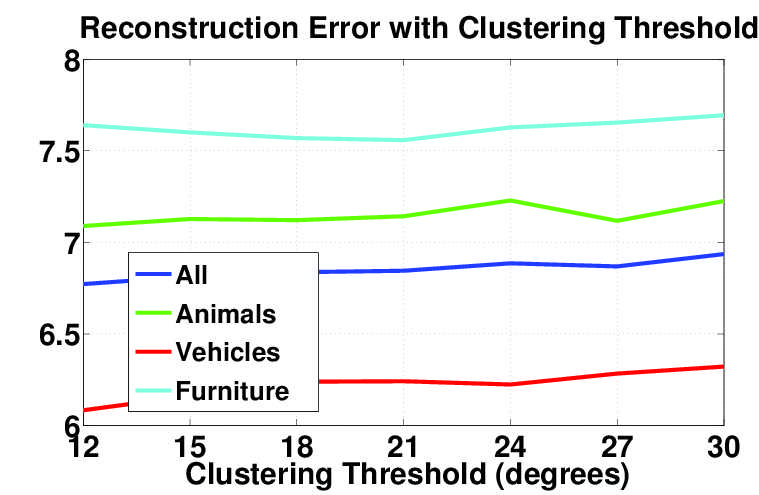}
\caption{Left: average per class symmetric RMS reconstruction error when considering the top ranked, randomly selected and best available reconstructions for each individual object. Middle: average symmetric RMS reconstruction error over all classes, as a function of the size of the pool of reconstruction proposals. ``Random'' and ``Best available'' represent, respectively, upper and lower bounds on ranking error. Right: error as function of the principal direction clustering threshold for different subsets of classes.\label{fig:ranking_best_rand_first}}
\end{figure*}

Fig.~\ref{fig:ranking_best_rand_first} suggests large gains of our simple ranking approach over random selection but also that there is much to improve with the addition of more advanced features. Fig.~\ref{fig:ranking_best_rand_first} also shows the effect of varying the principal direction clustering threshold, which we have set by default to 15º: reconstruction quality degrades slowly with looser thresholds. We have observed for example that cars tend to be more diamond-shaped if a tight frontal view is not available and instead views 30º or 40º away from the frontal view are used.


\section{Discussion}

While our results are encouraging for such a hard problem, there are several challenges that our approach does not address, such as modelling parts/articulation, occlusions and perspective effects. An additional limitation of our method is the use of a single ``average shape'' for the objects of a class. Although our experiments show that the camera viewpoint estimation step generally provides accurate results, this simplification may occasionally lead to incorrect camera pose estimates when the shape of the object instance differs significantly from the ``average shape''. Modelling subcategories would be a straightforward avenue for boosting the performance of all components - pose estimation, surrogate sampling and ranking. Two possible ways to obtain such subcategory information are: 1) to use image classifiers trained on a dataset with finer-grained category information and 2) to divide shapes into subcategories during the reconstruction process and iterate.

An additional potentially powerful direction  for future work is feature learning, in particular to improve the ranking of reconstructions, perhaps using one of the large collections of CAD models  available online \cite{shapenet}.

Finally, while our use of imprinting when computing visual hulls helps to mitigate the issues of using different object instances as surrogate shapes, this could be combined with other advanced visual hull techniques that explicitly deform the surrogate silhouettes to reduce inconsistencies between the silhouettes \cite{mitra2009shadow} or that enforce connectivity of the reconstruction \cite{Oswald-et-al-ECCV-2014}.

\section{Conclusion}
We have proposed a novel data-driven methodology for bootstrapping 3D reconstructions of objects in detection datasets, based on a small set of commonly available annotations, namely figure-ground segmentations and a small set of keypoints. Our approach is the first to target class-based 3D reconstruction on a challenging detection dataset, PASCAL VOC, and is demonstrated to achieve very promising performance. It produces convincing 3D shapes for most categories, handling widely different objects such as animals, vehicles and indoor furniture using the same integrated framework. We believe this paper contributes to the recently renewed interest in 3D modeling in recognition (eg. \cite{Hoiem:book,minsun}) and that it will promote progress in this direction since it provides the first semi-automatic solution to 3D model acquisition from detection data, which has been a difficult obstacle to research in joint object recognition and reconstruction.


%

\ifCLASSOPTIONcompsoc
  \section*{Acknowledgments}
\else
  \section*{Acknowledgment}
\fi

This work was supported by FCT grants PTDC/EEA-CRO/122812/2010 and SFRH/BPD/84194/2012, by the European Research Council under the ERC Starting Grant agreement 204871-HUMANIS. It was also also partly supported by the SecondHands project, funded from the European Unions Horizon 2020 Research and Innovation programme under grant agreement No 643950.

\ifCLASSOPTIONcaptionsoff
  \newpage
\fi



\bibliographystyle{IEEEtran}
\bibliography{all_refs}
%

%


\begin{IEEEbiography}[{\includegraphics[width=1\linewidth]{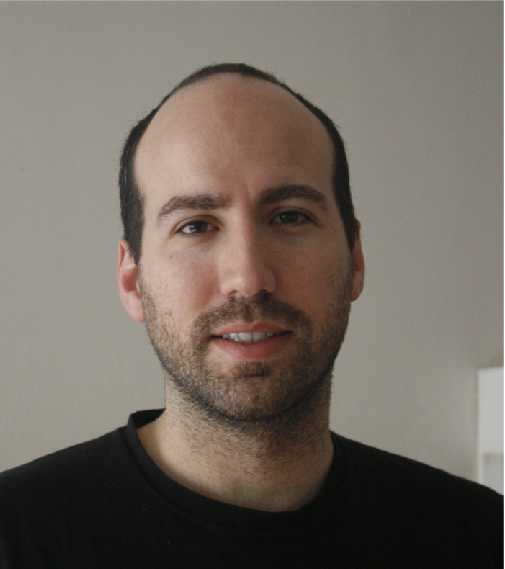}}]{Jo{\~a}o Carreira}
received his doctorate from the University of Bonn, Germany. His thesis focused on sampling class-independent object segmentation proposals using the CPMC algorithm, and on applying them in object recognition and localization. Systems authored by him and colleagues were winners of all four PASCAL VOC Segmentation challenges, 2009-2012. He did post-doctoral work at the Institute of Systems and Robotics in Coimbra, Portugal and is currently with the EECS department, at the University of California in Berkeley, USA. His research interests lie at the intersection of recognition, segmentation, pose estimation and shape reconstruction of objects from a single image.
\end{IEEEbiography}

\begin{IEEEbiography}[{\includegraphics[width=1\linewidth]{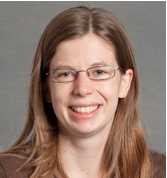}}]{Sara Vicente} received her PhD from University College London, United Kingdom. She was a postdoctoral researcher at Queen Mary, University of London and later at University College London. She currently works as a research scientist at Anthropics Technology.
Her research focuses on image segmentation and 3D
reconstruction of deformable objects from images.
\end{IEEEbiography}

\begin{IEEEbiography}[{\includegraphics[width=1\linewidth]{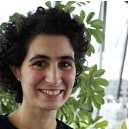}}]{Lourdes Agapito} received the BSc degree in
physics in 1991 and the PhD degree in 1996 from the Universidad Computense in Madrid, Spain. She was then a Marie Curie fellow at
Oxford’s Robotics Research Group. She is currently a reader in Vision and Imaging Science at University College London. In 2008, she was awarded an ERC Starting Grant. Her research focuses on the area of 3D
reconstruction of non-rigid structure from image sequences. She is a
member of the IEEE.
\end{IEEEbiography}

\begin{IEEEbiography}[{\includegraphics[width=1\linewidth]{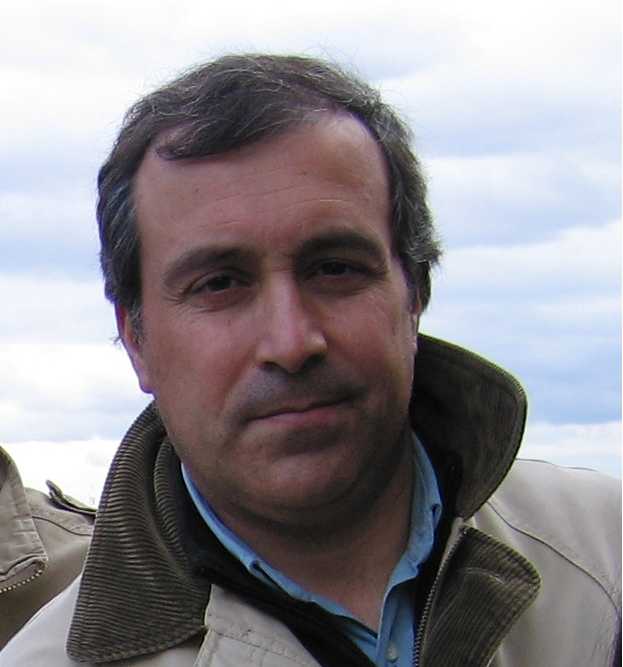}}]{Jorge Batista}
Prof. Jorge Batista received the M.Sc. and Ph.D. degree in Electrical Engineering from the University of Coimbra in 1992 and 1999, respectively. He joined the Department of Electrical Engineering and Computers, University of Coimbra, Coimbra, Portugal, in 1987 as a research assistant where he is currently an Associate Professor with tenure. He has been the Head of Department from 2011 to 2013. He is a founding member of the Institute of Systems and Robotics (ISR) in Coimbra, where he is a senior researcher and principal investigator of several research projects. His research interest focus on a wide range of computer vision and pattern analysis related issues, including real-time vision, video surveillance, video analysis, non-rigid modeling and facial analysis. 
\end{IEEEbiography}



\end{document}